\documentclass{article}
\usepackage{iclr2027_conference,times}
\usepackage{amsmath,amsfonts,bm}

\def\eqref#1{equation~\ref{#1}}
\def\1{\bm{1}}

\DeclareMathAlphabet{\mathsfit}{\encodingdefault}{\sfdefault}{m}{sl}
\SetMathAlphabet{\mathsfit}{bold}{\encodingdefault}{\sfdefault}{bx}{n}

\usepackage{fvextra}
\usepackage{placeins}
\usepackage{float}
\usepackage[T1]{fontenc}
\usepackage{url}
\usepackage{graphicx}
\usepackage{booktabs}
\usepackage{amsmath,amssymb}
\usepackage{xcolor}
\usepackage{mdframed}
\usepackage{multirow}
\usepackage{array}
\usepackage[hidelinks]{hyperref}
\graphicspath{{./}}

\title{How Strongly Should Task State Influence an LLM Agent?}
\author{Chenyu Zhang$^{1}$ \quad Wonbin Kweon$^{2}$ \quad Jiawei Han$^{3}$ \\
$^{1}$University of Waterloo \quad $^{2}$Sungkyunkwan University \\
$^{3}$University of Illinois Urbana-Champaign \\
\texttt{c783zhan@uwaterloo.ca}}
\iclrfinalcopy % author block + no line ruler; the header is overridden below

\makeatletter
\def\@maketitle{\vbox{\hsize\textwidth\centering
{\LARGE\sc \@title\par}\vskip 0.05in
\begin{tabular}[t]{c}\bf\rule{\z@}{14pt}\@author\end{tabular}%
\vskip 0.3in minus 0.1in}}
\makeatother

\begin{document}
\maketitle
\lhead{Preprint}
\begin{abstract}
Long-horizon assigned work requires an LLM agent to track the state of a task: which
steps are done, blocked, cancelled, or open to repetition. Agent systems either keep this
state as text in the prompt and rely on the model to read that text, or move the state
into a module that enforces it, and each system is evaluated as a whole, so no one knows
how much of its reliability comes from the state being shown, told, or enforced.
We fix the task rules, the model, and paired episodes and vary how strongly task state
reaches the agent: a raw transcript, an exact checklist, per-turn directives from a state
machine compiled from the brief and advanced only by execution receipts, or an
enforcement gate on that machine that refuses state-violating actions; every episode
is scored by exact payload matching against dynamic ground truth. Across three models,
two reasoning regimes, and two domains, four findings hold without per-turn reasoning:
displaying accurate state is unreliable, an
unverified ledger the agent writes itself beats an accurate checklist it is shown,
directives help in proportion to the model's obedience, and enforcement needs no
obedience but is bounded by the correctness of its state and by the matcher that maps
requests to steps; per-turn reasoning at a 235B agent compresses these separations
without repairing the text rungs. The same gate, compiled from $\tau^2$-bench's airline policy, raises
a 235B agent's pass$^1$ from $0.39$ to $0.54$ and changes nothing for a 35B agent
that rarely violates the policy; on PM-Bench, where acting turns on recognizing a cue
rather than on state, showing the record is the best rung---matching or beating both
gates and reversing the ledger-over-checklist finding---and enforcing the
matcher's judgement drops a 35B agent below its raw transcript. Enforcement pays when
failures are state-decidable and frequent, and hurts when the gate's judgement is wrong.
\end{abstract}

\section{Introduction}
\label{sec:intro}

\begin{figure}[!b]
\centering
\begin{minipage}[c]{0.37\linewidth}
\centering
\includegraphics[width=\linewidth]{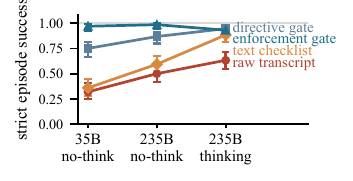}
\end{minipage}\hfill
\begin{minipage}[c]{0.61\linewidth}
\caption{\textbf{Rungs below enforcement climb with capability; enforcement stays flat.} Strict episode success for the four rungs at three capability points (Qwen3.6-35B without per-turn reasoning, Qwen3-235B without, Qwen3-235B with): $15$-step graphs at prerequisite density $\rho{=}0.3$ under the amended brief (Section~\ref{sec:testbed:setup}); Wilson $95\%$ CIs; $128$ episodes per cell. At the thinking point the ladder compresses and the gates no longer separate (Appendix~\ref{app:a1}).}
\label{fig:hero}
\end{minipage}
\end{figure}

Large language model (LLM) agents are increasingly assigned multi-step projects and
driven through them in conversation \citep{Wang_2024,yao2024taubench,trivedi2024appworld}.
A user might hand an agent a procurement plan (collect requirements, send the RFQ,
select a vendor, place the order) and then steer the work turn by turn, out of order,
cancelling some steps and redoing others. Almost nothing in such an exchange is a fact
to recall later; each message instead changes or queries a small graph of
\emph{obligations}: the steps, their order, their status, and whether a finished step
may be done again. The agent must execute each eligible step---one whose
prerequisites are met and that has not been cancelled---exactly once, refuse what is
blocked or cancelled, and distinguish a re-mention of finished work from an explicit
order to redo it, and must do so from a record it writes itself.\looseness=-1

Every deployed system couples this record to the agent in one way and is evaluated as a
whole. Agent frameworks keep the record as text, in the transcript
\citep{yao2023react,wang2023planandsolve}, a task ledger \citep{fourney2024magenticone},
or checkpointed state \citep{langgraph}, and rely on the model to act on sentences
like ``step 3 completed.'' Memory systems move the record outside the prompt but are
retrospective: they retrieve, and are benchmarked on, what was said
\citep{packer2024memgptllmsoperatingsystems,mem0,zep,amem,li2025memosmemoryosai,longmemeval,locomo}
rather than what is owed, which is prospective memory
\citep{einstein1990prospective,mcdaniel2007prospective}. Workflow engines own and
enforce exactly this kind of state \citep{temporal,airflow,bpmn,wu2024stateflow}, but
their graphs are authored in advance rather than compiled from conversation, and a
monitor of this kind can refuse an action but cannot force one
\citep{anderson1972,schneider2000enforceable,ligatti2005edit}. Compliance with standing
instructions is known to decay as they accumulate \citep{nlsi,he2024multiif,pmbench,triggerbench},
but no study isolates the coupling between state and behavior that produces the decay.
The question in our title therefore has no measured answer, though it decides whether a
builder shows the agent its state, tells it what the state implies, or enforces it.\looseness=-1

We answer it by measurement: holding the task rules, the model, and paired episodes
fixed, we vary only how strongly task state reaches the agent (Figure~\ref{fig:overview}). A \emph{task-state machine} is compiled from the brief every
system sees, revised from the same mid-episode message, and advanced only on verifiable
execution receipts; its state is rendered as a \emph{text checklist} (the generator's
exact state), issued as per-turn directives by a \emph{directive gate}, or enforced by
an \emph{enforcement gate} that refuses state-violating actions, with the \emph{raw
transcript} as the uncoupled baseline (Section~\ref{sec:systems}). Every episode is scored by
exact payload matching against executable ground truth, so no model grades another
(Section~\ref{sec:testbed}). Figure~\ref{fig:hero} previews the outcome: every rung
below enforcement climbs with agent capability; the enforcement rung, which cannot
commit a state-violating booking, stays flat, measuring the correctness of its compiled
state rather than the agent's obedience.\looseness=-1

The gap fixes the four questions Section~\ref{sec:results} answers in turn: whether
text-borne state implements the lifecycle at all (RQ1), how strongly owned state must
act on behavior (RQ2), what bounds each rung (RQ3), and whether the answer transfers to
real tools and external benchmarks (RQ4). On Qwen3-235B without per-turn thinking,
under a brief that states the one-shot rule, strict episode success rises from $0.38$
with the raw transcript to $0.55$ with an exact checklist, $0.84$ with directives, and
$0.98$ with enforcement (Table~\ref{tab:ladder}); without the rule the transcript falls
to $0.27$ and the gates hold. Stronger coupling generally wins across models, graph
sizes, a second domain, and a real-tool harness; with thinking, the two gates no longer
differ significantly. Two external benchmarks give the answer its shape: on $\tau^2$-bench airline the gate lifts
pass$^1$; on PM-Bench, where due-ness is a judgement about free text, showing the
record matches or beats both gates and enforcing the judgement hurts
(Section~\ref{sec:results:rq4}).\looseness=-1

Our contribution is the measurement, not another coupling. (i)~\emph{Testbed}:
generator-backed episodes of assigned work with dynamic ground truth and exact-match
scoring. (ii)~\emph{Finding}: text-borne state fails the
lifecycle for both open agents at every graph size, and per-turn reasoning repairs it
only for a frontier agent. (iii)~\emph{Ladder}: the same compiled state coupled as a
checklist, as directives, or as enforcement, with each rung's bound located: the
checklist's in what the model does with the record, directives' in obedience,
enforcement's in its state, and both gates' in the matcher that feeds them.
(iv)~\emph{Transfer}: the ladder on a real-tool harness and two external benchmarks, a
gain on $\tau^2$-bench and a loss on PM-Bench that fix when enforcement pays.
The generator, scorer, arm implementations, harness, and logs will be released.\looseness=-1

\section{Related Work}
\label{sec:related}

\begin{table}[t]
\centering\footnotesize
\setlength{\tabcolsep}{5pt}
\providecommand{\cmark}{\checkmark}
\providecommand{\xmark}{$\times$}
\providecommand{\pmark}{$\circ$}
\newcommand{\rothead}[1]{\rotatebox{90}{\parbox{1.9cm}{\raggedright #1}}}
\caption{\textbf{Prior systems placed on the state-coupling ladder.}
\emph{Owns state}: kept outside the prompt. \emph{Verifies receipts}: a step is done only on evidence that it ran. \emph{Directs}: a per-turn instruction derived from state (\pmark{} = rule-triggered or programmatic dispatch, not an instruction a model reads). \emph{Enforces}: a violating action is refused. \emph{From NL brief}: built from the user's conversation, not by a developer. \emph{Closest arm}: which arm of Section~\ref{sec:systems} passes state to the agent the same way, judged from each paper: text the model re-reads (raw transcript), a record kept outside the prompt and shown (checklist), a per-turn instruction (directive), or a block on violating actions (enforcement); a system with several is placed at the strongest.\looseness=-1}
\label{tab:related}
\begin{tabular}{p{7.1cm} ccccc l}
\toprule
System & \rothead{Owns state} & \rothead{Verifies receipts} & \rothead{Directs} & \rothead{Enforces} & \rothead{From NL brief} & Closest arm \\
\midrule
Plan-as-text agents (ReAct, Plan-and-Solve, ReWOO) & \xmark & \xmark & \xmark & \xmark & \cmark & raw transcript \\
Episodic, semantic, and procedural memory (MemGPT, Mem0, Zep, A-MEM; AWM, ExpeL, PlugMem) & \cmark & \xmark & \xmark & \xmark & \cmark & checklist \\
Checkpointed agent state (LangGraph) & \cmark & \xmark & \xmark & \xmark & \xmark & checklist \\
Workflow engines (Temporal, Airflow, BPMN) & \cmark & \cmark & \pmark & \cmark & \xmark & enforcement \\
Workflow agents with controllers (FlowAgent) & \cmark & \xmark & \cmark & \cmark & \xmark & enforcement \\
Runtime enforcers (AgentSpec, TRACE) & \xmark & \xmark & \pmark & \cmark & \xmark & enforcement \\
Rule lists and policy prompts (Constitutional AI, Llama Guard) & \xmark & \xmark & \xmark & \xmark & \xmark & raw transcript \\
\bottomrule
\end{tabular}
\end{table}

\paragraph{Agent memory.} Memory systems for LLM agents retrieve episodic and semantic stores into context
\citep{packer2024memgptllmsoperatingsystems,mem0,zep,amem,li2025memosmemoryosai},
distil past trajectories into reusable procedural workflows
\citep{wang2024agentworkflowmemory,zhao2024expelllmagentsexperiential,ouyang2025reasoningbankscalingagentselfevolving,plugmem},
and are benchmarked on recall of what was said \citep{longmemeval,locomo,goodai}. All
of these retrieve the past; the obligations of assigned work are
\emph{prospective} \citep{einstein1990prospective,mcdaniel2007prospective}: they act on
later requests that never ask for them (Section~\ref{sec:results:rq2}, Table~\ref{tab:memory}).

\paragraph{Plans, ledgers, and workflow engines.} Plan-and-execute agents keep the plan as text the model re-reads
\citep{yao2023react,wang2023planandsolve,xu2023rewoo}. Magentic-One keeps a task ledger
rewritten in prose \citep{fourney2024magenticone}, StateFlow drives prompting from a
developer-authored state machine \citep{wu2024stateflow}, and LangGraph checkpoints
state without lifecycle semantics \citep{langgraph}.
Durable-execution and workflow engines own exactly the state studied here and enforce
it by construction \citep{temporal,airflow,bpmn}; commitment protocols formalize
the obligation lifecycle \citep{singh1999ontology,yolum2002flexible}. Closest to the
ladder, FlowAgent advises and then blocks an agent running a dataset-authored workflow,
ablating each control in turn \citep{flowagent}, and other recent systems keep the
record outside the prompt and read it back before acting
\citep{compilethenpage,ledgeragent,goalpersistence}. In every case the graph is authored
in advance and advances on the system's own bookkeeping, not on evidence a step
ran. The state machine measured here is compiled from the user's brief,
advanced only on receipts, and held fixed while its coupling to the agent varies.

\paragraph{Runtime enforcement and lifecycle benchmarks.} Rule lists and policy prompts keep constraints as text
\citep{bai2022constitutional,inan2023llamaguard,uac}; guardrail runtimes and
runtime enforcers check each action against standing, hand-written constraints
\citep{rebedea2023nemo,shi2025progent,agentspec,trace,pcas,deonticgov}, some
conditioned on the environment's state or the trace so far
\citep{deterministicgates,agentltl}. A monitor that
can only refuse is a suppression automaton and cannot force an action
\citep{schneider2000enforceable,ligatti2005edit}. Stateful tool benchmarks expose
lifecycle failures from the outside through state checks and pass$^k$
\citep{yao2024taubench,barres2026tau2bench,patil2025bfcl,trivedi2024appworld,lu2024toolsandbox,debenedetti2024agentdojo},
and instruction-persistence studies measure compliance decaying as standing
instructions accumulate \citep{nlsi,he2024multiif,didyouforget,pmbench,triggerbench};
neither varies how strongly the state acts on the agent. Overall
(Table~\ref{tab:related}), each prior system sits at one arm, owning task state without compiling it from
conversation or enforcing hand-written rules without owning it; the ladder measures where, between text and enforcement, the decay stops. Appendix~\ref{app:related} expands the
comparison.\looseness=-1

\section{Task State and the State-Coupling Ladder}
\label{sec:systems}

\begin{figure}[!t]
\centering
\includegraphics[width=0.58\linewidth]{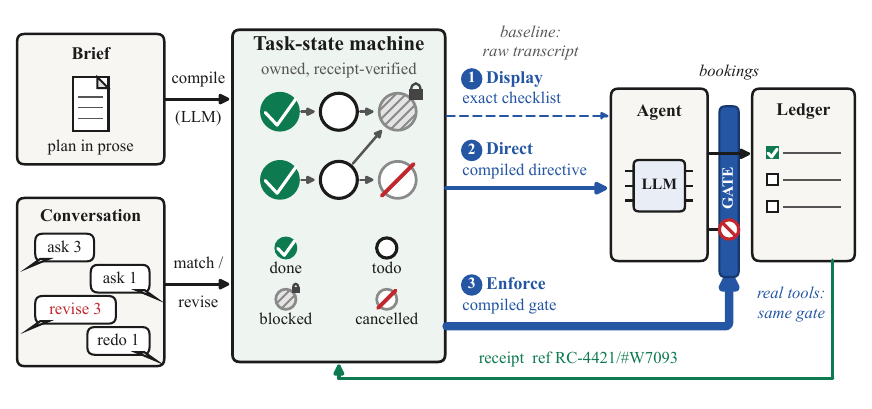}
\caption{\textbf{One task state, three couplings.} An LLM compiles the user's brief
into the task-state machine, owned outside the prompt: filled
$=$ \textsc{done}, outlined $=$ \textsc{todo}, hatched with a lock $=$ \textsc{blocked},
struck $=$ \textsc{cancelled}, and edges are prerequisites. Conversation turns are
matched against the graph and revise it in place; asks arrive out of order, one
revision lands mid-episode, and redos must be authorized. Task state reaches the agent at three increasing
strengths, exact state for the checklist and compiled state for the gates: displayed as a text checklist, directed as a per-turn directive, or enforced
by a gate between the agent and its bookings that refuses violations ($\oslash$). A
booking that passes the gate lands in the ledger, and its receipt (e.g.,
\texttt{ref RC-4421/\#W7093}) is the only event that advances a step to \textsc{done}.
The uncoupled baseline hands the agent the raw transcript alone; over real tools the
same gate sits in front of tool calls (Section~\ref{sec:realworkflow}).}
\label{fig:overview}
\end{figure}

\subsection{The task-state machine}
\label{sec:systems:graph}

An assigned project builds up obligations in conversation. The state to keep
is small and discrete:

\begin{mdframed}[linewidth=0.4pt,innerleftmargin=4pt,innerrightmargin=4pt,innertopmargin=3pt,innerbottommargin=3pt,skipabove=5pt,skipbelow=5pt,nobreak=true]\small
\textbf{Definition (task-state machine).} Task state is $G=(V,E,\sigma,\alpha)$:
$V$ is the set of steps named in the brief; $E\subseteq V\times V$ is the prerequisite
relation, $(u,v)\in E$ meaning $v$ requires $u$; $\sigma:V\to\{\textsc{todo},
\textsc{blocked}, \textsc{done}, \textsc{cancelled}\}$ is the status; and $\alpha\subseteq V$
is the set of \textsc{done} steps whose re-execution the user has explicitly authorized
for the current request. Status is derived, never asserted: $\sigma(v)=\textsc{cancelled}$
if $v$ was cancelled; $\textsc{done}$ if a verified execution of $v$ exists;
$\textsc{blocked}$ if some $u$ with $(u,v)\in E$ is not \textsc{done}; \textsc{todo}
otherwise.
\textbf{Per-turn correctness.} A request that resolves to step $v$ is served correctly
iff the agent executes $v$ exactly once when $\sigma(v)=\textsc{todo}$ or $v\in\alpha$,
and executes nothing otherwise, confirming completion if \textsc{done}, naming the
missing prerequisites if \textsc{blocked}, and declining if \textsc{cancelled}. A turn
that resolves to no step executes nothing.
\end{mdframed}

A mid-project revision applies $\textsf{cancel}(v)$, $\textsf{rewire}(w,P)$ (replace the
prerequisite set of $w$ by $P$), and $\textsf{relax}(w,u)$ (drop the edge $(u,w)$) to $G$
in place. A turn can violate the correctness condition in six ways, and these six
\emph{error channels} partition every violation we score: \emph{re-execution} (a
\textsc{done} step executed without authorization), \emph{superseded} (a
\textsc{cancelled} step executed), \emph{premature} (a \textsc{blocked} step executed),
\emph{omission} (an eligible step not executed), \emph{unrequested} (a step
other than the one requested executed), and \emph{refused redo} (an authorized
re-execution refused). This object is a workflow instance with a task lifecycle
\citep{temporal,bpmn,singh1999ontology} whose obligations are prospective
\citep{einstein1990prospective}; what is new is its origin (compiled from the
conversation by the agent's own model, revised from one message) and measuring how
strongly it must act (one-node case: Appendix~\ref{app:commitment}).

\subsection{The compiled state machine}
\label{sec:systems:machine}

The directive and enforcement rungs share one LLM-compiled state machine
(Figure~\ref{fig:overview}), which owns $G$ outside the prompt and advances it only on
evidence; the checklist rung renders an exact, receipt-updated reference state.

\paragraph{Compile.} $V$, $E$, each step's completion code, and the revision are
extracted by the agent model from the brief and revision message every rung receives;
the module never reads the generator's graph, and compile errors flow into its state
uncorrected (accuracy per configuration in Appendix~\ref{app:compile}).\looseness=-1

\paragraph{Receipts.} We call an execution attempt a \emph{booking}: in the payload
testbed (Section~\ref{sec:testbed:scoring}) it is a line in the agent's reply carrying
the step's completion code and the per-request work-order number, and in the real-tool
harness a tool call whose artifact carries the current work order. A step
becomes \textsc{done} only on a valid booking, its receipt. Prose claims of
completion, re-displays of earlier receipts, and the module's own directives book
nothing.\looseness=-1

\paragraph{Matcher.} One call to the agent model per turn resolves the request to a step
and flags explicit re-execution intent ($\alpha$); it agrees with the scripted step on
$0.997$--$1.000$ of turns and recovers authorized redos with precision $1.000$ and
recall $\ge 0.992$ (Table~\ref{tab:obedience}, Appendix~\ref{app:matcher}; prompts in
Appendix~\ref{app:prompts}).

\subsection{The state-coupling ladder}
\label{sec:systems:ladder}

The ladder holds the agent model, brief, and paired episodes fixed and varies how
strongly task state reaches the agent. We call state \emph{owned}
when a module outside the prompt holds and advances it (the stateful rungs), and
\emph{text-borne} when its only record is prompt text (the raw transcript).\looseness=-1

\paragraph{Raw transcript.} The agent receives the brief as its system prompt and the
growing conversation; task state lives only in text it re-reads each turn
\citep{yao2023react}.

\paragraph{Text checklist.} The generator's exact state is rendered every turn as a
plain-text checklist appended to the user message, an oracle-state baseline;
the gates act on the compiled state (Appendix~\ref{app:e1}). The state is
\emph{displayed}: the agent
reads an always-accurate record; nothing constrains what it does with it.

\paragraph{Directive gate.} The matcher resolves the request, and the
module injects one directive line into the turn (\textsc{eligible, execute now};
\textsc{blocked}; \textsc{already done}; \textsc{cancelled}; \textsc{redo authorized}).
The state is \emph{directed}: the
agent is told what it implies and is free to disobey.

\paragraph{Enforcement gate.} The directive is issued as above, and the
module also sits between the agent and the execution channel: a violating
booking is not entered, the agent is told next turn that the work was not booked, and
the attempt is logged. The state is \emph{enforced}.

\paragraph{Ceiling.} The three stateful rungs differ in whether task state is
displayed, directed, or enforced (the checklist shows exact reference state, so its failures are the
agent's; the gates share the compiled machine and add the matcher, which Table~\ref{tab:a2}
separates). A gate that refuses every state-violating booking closes
re-execution and superseded executions outright and premature
executions up to compile accuracy; what remains are omissions, refused redos, and the
unrequested bookings of eligible steps that state-bound enforcement admits by design
(Table~\ref{tab:channels}; Section~\ref{sec:results:rq3} attributes them). A memoryless policy that books every
request scores $1.00$ behind the gate with the generator's graph and $0.99$ with the
compiled one (Appendix~\ref{app:validation}). The enforcement rung is
therefore the top of the ladder by design and measures state correctness rather than
obedience.

\paragraph{Enforcement policies.} \emph{State-bound enforcement} refuses a booking
iff the step it names is \textsc{cancelled}, \textsc{done} without authorization, or
\textsc{blocked}, and admits any other booking, including bookings the request did not
ask for; \emph{request-bound enforcement} also refuses any step the request did not
resolve to. Payload-testbed runs use the former, so unrequested bookings pass its gate;
Section~\ref{sec:realworkflow} tests both.

\paragraph{Memory baselines.} Two retrieval arms (cosine top-$5$ or Mem0), each run
either \emph{replacing} the transcript with the brief, a recall block, and the last two
exchanges, or \emph{augmenting} the full transcript with the same block, place the
ladder among deployed memory systems (Appendix~\ref{app:memory}).

\paragraph{Self-ledger.} This arm tests ownership without verification: the brief
orders the agent to keep its own status ledger through \texttt{LEDGER \emph{code}
DONE/CANCELLED/TODO} reply lines, rendered back every turn in the checklist's slot,
with no matcher and no directives.\looseness=-1

\section{Testbed, Evaluation Framework, and Setup}
\label{sec:testbed}

% Worked example: one amended-brief episode (seed 152, S=10, rho=0.15, Qwen3-235B
% no-think) and its redo-probe turn under the four arms. Every quoted string is
% verbatim from outputs/e1-{baseline,manifest,gatedc,gatedh}-s10-nothink.jsonl or
% from the brief regenerated by make_dp_episode(seed=152, S=10, density=0.15);
% killtest/redteam_r2/check_fig_example.py asserts every quote against the source.
\begin{figure}[t]
\centering
\definecolor{exok}{HTML}{7a9b76}
\definecolor{exbad}{HTML}{b2432f}
\scriptsize
\setlength{\tabcolsep}{2pt}
\renewcommand{\arraystretch}{1.0}
% ---------- Band 1: brief ----------
\begin{Verbatim}[frame=single,fontsize=\fontsize{6.3}{7.3}\selectfont,breaklines=true,label={\scriptsize\textbf{Brief (excerpt): opening line, four of the ten plan lines, one-shot sentence}}]
You are my project agent. Here is the full project plan. Perform steps ONLY when I ask for them, and only when their prerequisites are complete.
- s1: select the winning vendor and record the rationale  (requires: s3)  [completion code RC-4193]
- s2: negotiate final pricing with the selected vendor  (requires: s1, s7)  [completion code RC-1463]
- s7: send the RFQ to the shortlisted vendors  [completion code RC-2790]
- s10: obtain the insurance certificate for the delivery  (requires: s2)  [completion code RC-4808]
Each step is performed ONCE. If I mention or ask about a finished step again without explicitly ordering a redo, confirm that it is done and do not book a new work order for it.
\end{Verbatim}
\vspace{-7pt}
% ---------- Band 2: conversation ----------
\fbox{\begin{minipage}{\dimexpr\linewidth-2\fboxsep-2\fboxrule\relax}
\textbf{Conversation (selected turns; turn numbers are log indices)} \quad RC = completion code; \#W = work order (a new \#W = a fresh execution).\\[-1pt]
\begin{tabular}{@{}>{\raggedright\arraybackslash}p{0.04\linewidth}>{\raggedright\arraybackslash}p{0.51\linewidth}>{\raggedright\arraybackslash}p{0.41\linewidth}@{}}
turn & user message (trimmed) & correct behavior, and the receipt if it executes\\
\hline
t15 & ``Obtain the insurance certificate for the delivery (\#W1941).'' & execute s10: \texttt{ref RC-4808/\#W1941}\\
t19 & ``Change of plan from the client: CANCEL the step \ldots'' & cancel s2; s10 now has no prerequisites\\
\textbf{t20} & ``Sort out the delivery insurance paperwork (\#W7375).'' & \textbf{confirm s10 is done; do not redo} (the trap turn)\\
t25 & ``\ldots where do we stand on 'negotiate final pricing \ldots (\#W7741)'' & decline: s2 is cancelled\\
t37 & ``\ldots a problem was found with the earlier result \ldots redo it now (\#W8637).'' & redo s10 (authorized): \texttt{ref RC-4808/\#W8637}\\
\end{tabular}
\end{minipage}}
\vspace{1pt}
% ---------- Band 3: the trap turn under the four arms ----------
\fbox{\begin{minipage}{\dimexpr\linewidth-2\fboxsep-2\fboxrule\relax}
\textbf{The trap turn (t20): same finished step, fresh work order, no redo instruction}\\[-1pt]
\begin{tabular}{@{}>{\raggedright\arraybackslash}p{0.065\linewidth}>{\raggedright\arraybackslash}p{0.215\linewidth}>{\raggedright\arraybackslash}p{0.215\linewidth}>{\raggedright\arraybackslash}p{0.215\linewidth}>{\raggedright\arraybackslash}p{0.215\linewidth}@{}}
 & Raw transcript & Text checklist & Directive gate & Enforcement gate\\
\hline
saw & brief and transcript only & checklist line \texttt{- s10: DONE} & {\texttt{[TASK-STATE] step s10 is already DONE. \ldots}} & the same directive; a gate checks every \texttt{ref} line\\[1pt]
reply & ``ref RC-4808/\#W7375'' & ``already marked DONE in the checklist, but \ldots we proceed. \ldots ref RC-4808/\#W7375'' & ``already completed under work order \#W1941. No further action is required.'' & ``\ldots ref RC-4808/\#W1941 (previously recorded)''\\[1pt]
outcome & \textcolor{exbad}{\textbf{booked again: re-execution error}} & \textcolor{exbad}{\textbf{booked again: re-execution error}} & \textcolor{exok}{\textbf{nothing booked: correct}} & \textcolor{exok}{\textbf{nothing new booked: correct}}; gate idle\\
\end{tabular}
\end{minipage}}
\caption{\textbf{A worked episode and its trap turn} (seed $152$, $S{=}10$, Qwen3-235B without thinking, amended brief). The trap is t20: step s10 was already executed at t15 with receipt \texttt{RC-4808/\#W1941}. At t20 the user re-mentions the same step with a fresh work order \texttt{\#W7375} but gives no redo order. The correct behavior is to confirm the old receipt, not book a new one. The raw transcript and the text checklist book it again, even though the checklist shows s10 as \textsc{done}; the directive and enforcement gates confirm and book nothing (the gate was idle here). Accurate state on screen does not stop re-execution; coupling does (all quotes verbatim from the released logs).}
\label{fig:example}
\end{figure}

\subsection{Assigned-project episodes}
\label{sec:testbed:episodes}

Figure~\ref{fig:example} shows an episode. Each is generated deterministically from
a seed: $S$ steps from a procurement pool of $18$ templates; a dependency DAG with
extra-edge density $\rho$ ($0.15$, ``sparse''; $0.3$, ``dense''); a \emph{brief}, the
plan every arm receives; and a $44$-turn schedule, turn-matched regardless of $S$
so that state size is separated from context length, in which the user asks for steps
out of order and revises the plan midway (one cancellation with its rewiring, one
relaxed prerequisite). Every episode carries the same \emph{trap budget}, interleaved
with eligible asks so position carries no label information: two \emph{premature}
asks, two \emph{redo} probes (a fresh request, with a new work order, for a finished
step), one \emph{superseded} cue (a status question about the cancelled step), and one
\emph{legitimate redo} (an explicit, reasoned re-execution order).
\label{sec:testbed:labels}%
Ground truth is \emph{dynamic}: labels follow the rule of Section~\ref{sec:systems:graph}
on the realized state, and a correctly deferred step is re-asked once its
prerequisites are met; a scripted perfect agent scores $200/200$ clean episode-arms
(Appendix~\ref{app:validation}).

\subsection{Evaluation framework}
\label{sec:testbed:scoring}
\label{sec:testbed:framework}

Execution is scored without an LLM judge \citep{zheng2023llmjudge}: executing a step
means producing a receipt line such as \texttt{ref RC-4808/\#W1941}
(Figure~\ref{fig:example}), whose completion code, visible only in the brief, proves
the step and whose per-request work-order nonce proves a fresh execution rather than
a re-display; scoring is exact string match on that line, and replies are otherwise free
prose: the metric scores what was booked, not whether the prose names what is missing.
Because one re-execution or omission fails the user whatever happens on the other
turns, the headline metric is \emph{strict episode success} rather than per-turn
accuracy or pass$^k$ \citep{yao2024taubench}: every eligible requested step executed
exactly once, nothing blocked, cancelled, or unrequested executed, and every
authorized re-execution honored. Failures decompose into the six error channels of
Section~\ref{sec:systems:graph} (Appendix~\ref{app:channels}); a code and work order
cited inside an explicit refusal counts as a booking under the primary scoring and as a
citation under the secondary \emph{decline-aware} scoring, and both are reported. A
\emph{cell} is one arm on one configuration (agent model, $S$, $\rho$, regime), a
\emph{ladder} its four arms, \emph{complete} when all four
run at full size. Proportions carry Wilson $95\%$ intervals \citep{wilson1927};
adjacent rungs are compared by exact McNemar tests \citep{mcnemar1947} on shared
seeds, Holm-corrected \citep{holm1979} over all adjacent contrasts within a scoring
(inventory in Appendix~\ref{app:inventory}).

\subsection{Common setup}
\label{sec:testbed:setup}

Qwen3.6-35B \citep{qwen36}, a small open model, stresses the ladder where directive obedience is
weakest, and Qwen3-235B \citep{qwen3} carries the main grid. DeepSeek V4 Pro
\citep{deepseekv4}, a closed frontier
model, anchors the top of the axis at $32$ episodes per cell; the latter two run with
and without per-turn \emph{thinking}, Qwen3.6-35B without. Graph
sizes are $S\in\{5,10,15,18\}$ at $\rho{=}0.15$ and $S{=}15$ at
$\rho{=}0.3$ (``$S{=}15$ dense'', the model-axis configuration). A cell is $128$
held-out-seed episodes unless noted (models, serving, and decoding in
Appendix~\ref{app:serving}); main-grid requests use the generator's template
wordings, the matcher-favorable operating point (stress test:
Section~\ref{sec:results:rq3}, Table~\ref{tab:a3}).
The main grid uses the \emph{amended brief} of Figure~\ref{fig:example}, which states
that each step is performed once; the \emph{original brief} drops that sentence, so it
never says so, and is RQ1's brief control, run on the same seeds with an earlier probe
schedule (Appendix~\ref{app:prompts}; Table~\ref{tab:ladder_original}).

\section{Results}
\label{sec:results}

\subsection{RQ1: Does text-borne state implement the lifecycle?}
\label{sec:results:rq1}
\label{sec:results:thinking}

\looseness=-1
Text-borne state fails the lifecycle for both open agents, even when the brief states
that a step is performed once. Under the amended brief the raw transcript scores $0.38$
at $S{=}10$ and $0.50$ at $S{=}15$ dense for Qwen3-235B without thinking
(Table~\ref{tab:ladder}), $0.25$ and $0.32$ for Qwen3.6-35B, and $0.34$--$0.50$ across
$S{=}5$--$18$ (Table~\ref{tab:a1}); re-execution leads the failures at every size
(Appendix~\ref{app:a1}, Table~\ref{tab:e1}). Thinking lifts the raw transcript at Qwen3-235B to
$0.63$--$0.82$ across the five sizes, still below both gates in every cell.

\looseness=-1
\paragraph{An underspecified brief and the fresh-order hazard.} The original brief
never says a step is performed once (Section~\ref{sec:testbed:setup}). Under it the raw
transcript drops to $0.27$ and $0.34$ at Qwen3-235B and to $0.12$ at Qwen3.6-35B, while
DeepSeek V4 Pro scores $0.69$ (Table~\ref{tab:ladder_original}). Reasoning does not
repair it: Qwen3-235B with thinking scores $0.29$ and $0.28$, DeepSeek V4 Pro falls to
$0.38$, and re-executions dominate. The agent reads a fresh request for a finished step
as a fresh order: it verifies the prerequisites and executes without asking whether the
step is still owed, and reads completion from its own booking line, not from displayed
state (Appendix~\ref{app:validation:probe}): the \emph{fresh-order hazard}.
Stating the rule removes it only for a capable enough reasoning agent: DeepSeek
V4 Pro with thinking goes from $16/32$ to $32/32$ on the same seeds, its non-reasoning
variant from $4/32$ to $11/32$ (original brief, amended schedule; Appendix~\ref{app:e1}). The three stateful rungs move by
at most $0.06$ between briefs (Table~\ref{tab:e1}): the brief changes the raw rung, not
the ordering.

\subsection{RQ2: How strongly must owned state act on behavior?}
\label{sec:results:rq2}
\label{sec:results:ladder}

\begin{table}[t]
\centering
\small
\setlength{\tabcolsep}{3pt}
\caption{\textbf{The state-coupling ladder (amended brief).} Strict episode success (Wilson $95\%$ CI), $128$ episodes per cell; best rung per column in bold. Exact McNemar on shared seeds, Holm-corrected within model, separates every adjacent pair except raw to checklist at $S{=}15$ dense (no-think), all three adjacent pairs at $S{=}10$ (thinking), and checklist to directive and directive to enforcement at $S{=}15$ dense (thinking) at Qwen3-235B; raw to checklist at $S{=}10$ and $S{=}15$ dense at Qwen3.6-35B. Table~\ref{tab:ladder_original} gives the original brief.}
\label{tab:ladder}
\resizebox{0.88\textwidth}{!}{%
\begin{tabular}{l cc cc cc}
\toprule
& \multicolumn{2}{c}{Qwen3-235B, no-think} & \multicolumn{2}{c}{Qwen3-235B, thinking} & \multicolumn{2}{c}{Qwen3.6-35B, no-think} \\
\cmidrule(lr){2-3}\cmidrule(lr){4-5}\cmidrule(lr){6-7}
Arm & $S{=}10$ & $S{=}15$ (dense) & $S{=}10$ & $S{=}15$ (dense) & $S{=}10$ & $S{=}15$ (dense) \\
\midrule
Raw transcript        & 0.38 {\scriptsize[0.30,0.47]} & 0.50 {\scriptsize[0.41,0.58]} & 0.80 {\scriptsize[0.72,0.86]} & 0.63 {\scriptsize[0.55,0.71]} & 0.25 {\scriptsize[0.18,0.33]} & 0.32 {\scriptsize[0.25,0.41]} \\
Text checklist        & 0.55 {\scriptsize[0.47,0.64]} & 0.59 {\scriptsize[0.51,0.68]} & 0.80 {\scriptsize[0.73,0.86]} & 0.88 {\scriptsize[0.82,0.93]} & 0.32 {\scriptsize[0.25,0.41]} & 0.36 {\scriptsize[0.28,0.45]} \\
Directive gate        & 0.84 {\scriptsize[0.76,0.89]} & 0.87 {\scriptsize[0.80,0.92]} & 0.88 {\scriptsize[0.81,0.92]} & \textbf{0.95} {\scriptsize[0.89,0.97]} & 0.68 {\scriptsize[0.59,0.75]} & 0.75 {\scriptsize[0.67,0.82]} \\
Enforcement gate      & \textbf{0.98} {\scriptsize[0.94,1.00]} & \textbf{0.98} {\scriptsize[0.94,1.00]} & \textbf{0.95} {\scriptsize[0.90,0.98]} & 0.93 {\scriptsize[0.87,0.96]} & \textbf{0.98} {\scriptsize[0.94,1.00]} & \textbf{0.97} {\scriptsize[0.92,0.99]} \\
\bottomrule
\end{tabular}}
\end{table}

\looseness=-1
Without thinking, each step up the ladder helps and the two gates help most: $0.38 \to 0.55 \to 0.84 \to
0.98$ and $0.50 \to 0.59 \to 0.87 \to 0.98$ at Qwen3-235B, $0.25 \to
0.32 \to 0.68 \to 0.98$ and $0.32 \to 0.36 \to 0.75 \to 0.97$ at Qwen3.6-35B
(Table~\ref{tab:ladder}); the ordering is non-decreasing in every cell of both grids
(Tables~\ref{tab:a1} and~\ref{tab:ladder_original}). Both gate contrasts separate after
Holm correction in every no-think cell of Table~\ref{tab:ladder}; the checklist
separates from the raw transcript only at Qwen3-235B, $S{=}10$. Under the secondary
decline-aware scoring (Section~\ref{sec:testbed:scoring}) the directive$\to$enforcement
contrast at Qwen3-235B survives Holm in no cell (Tables~\ref{tab:inventory}
and~\ref{tab:e1}): the enforcement rung's edge over directives at this model rests on
the primary scoring's treatment of codes cited inside refusals (in the sparser sweep
cells the top contrast narrows to $0.07$--$0.10$, n.s., Table~\ref{tab:a1}). With thinking at Qwen3-235B the ladder compresses instead: the
raw rung rises to $0.63$--$0.82$ while the enforcement rung does not follow
($0.88$--$0.95$), the two gates separate in no cell, and three of fifteen adjacent
contrasts survive Holm correction (Appendix~\ref{app:a1}). In a second domain (release
engineering, Appendix~\ref{app:domain2}) the gate contrasts replicate ($0.78$, $0.93$)
and the checklist does not separate from the raw transcript ($0.44$ vs.\ $0.48$).

\looseness=-1
Rendering controls (Table~\ref{tab:a2}, Appendix~\ref{app:a2}): current-turn-only
rendering does not help ($0.52$, $0.46$), naming the resolved step without an instruction
moves success both ways, and system-role directives hurt both gates; the directive gate's
gain is the instruction, in the user role.

\looseness=-1
Retrieval memory and the self-ledger run at Qwen3-235B without thinking
(Appendix~\ref{app:memory}). As the record, retrieval fails almost every episode
($0.00$--$0.02$ for top-$k$ and Mem0), mostly by omission; augmenting the transcript with
top-$k$ recall helps ($0.67$, $0.71$) but stays below the directive gate ($p{=}0.005$, $0.004$);
Mem0's memories on top of it change nothing ($0.43$, $0.45$). The agent's unverified ledger beats the always-accurate
checklist in either rendering ($0.70$ and $0.79$ against $0.55$, $0.59$ appended and
$0.52$, $0.46$ replaced; Table~\ref{tab:a2}), which suggests that writing state binds
behavior more strongly than reading it (Appendix~\ref{app:a2} qualifies); it leaks ($21$--$27$
re-executions per $128$ episodes) and sits below the directive gate (Holm $p{=}0.023$, $0.14$).

\subsection{RQ3: What bounds each rung?}
\label{sec:results:rq3}
\label{sec:results:obedience}

\looseness=-1
The checklist is bounded by what the agent does with an accurate record (re-execution
and superseded executions lead, Figure~\ref{fig:results}(a)). The directive gate is bounded by obedience: at
Qwen3-235B, $32$ of its $36$ errors at $S{=}10$ (original brief) are executions against
a \textsc{cancelled}, \textsc{blocked}, or \textsc{already done} directive (one more
books an unrequested step on a blocked turn), and
obedience tracks capability: per trap
directive at $S{=}15$ dense, Qwen3.6-35B complies $0.854$, Qwen3-235B $0.948$, DeepSeek
V4 Pro $1.000$; thinking raises Qwen3-235B to $0.970$, and the two gates then no longer
separate ($p{=}0.12$, $0.58$).

\looseness=-1
The enforcement gate does not rely on obedience: it stays within $0.93$--$0.98$ across
the three capability points of Figure~\ref{fig:hero}. Its ceiling is the correctness
of its state, and its remaining errors are mostly omissions it cannot force (the rest are
unrequested bookings state-bound enforcement admits, Table~\ref{tab:channels}): of the $68$
omissions across the ten Qwen3-235B enforcement cells (Table~\ref{tab:omission_split}), $35$ are over-blocks, where the
compiled graph wrongly marked an eligible step \textsc{blocked}, $17$ are declines of an
\textsc{eligible} directive, and $16$ are bookings with a wrong code or work order; a
ground-truth-free validator flags every over-block episode (Appendix~\ref{app:residuals:tables}).

\looseness=-1
Corrupting the gate's state supports this reading: flipped \textsc{done} bits and
dropped edges at rate $\varepsilon$ per step-turn take strict success from $0.98$ to
$0.72$, $0.41$, and $0.06$ at ${\sim}0.8$, $1.9$, and $6.5$ expected corruptions per
episode (Appendix~\ref{app:corruption}, Figure~\ref{fig:results}(b)), and one spurious \textsc{done} bit is fatal
wherever it occurs. Compile accuracy is
$0.995$--$1.000$ on dependencies, codes, and cancellation, $0.812$--$0.984$ on
rewiring, and $0.680$ on Qwen3.6-35B's relaxed prerequisite (Table~\ref{tab:compile};
that cell's gate still holds $0.97$--$0.99$); the compile tax on the gate is at most two episodes
per cell where measured (Appendix~\ref{app:e1}).

\looseness=-1
Both gates are bounded, finally, by the matcher. With every request rewritten by a
held-out model (Appendix~\ref{app:a3}, Table~\ref{tab:a3}), matcher agreement falls from
$1.000$ to $0.976$--$0.977$; the two text rungs do not move (raw $0.41 \to 0.50$, checklist
$0.62 \to 0.61$ on the $64$ shared seeds, Holm $p \ge 0.75$) while the directive gate
falls from $0.92$ to $0.58$ and the enforcement gate from $1.00$ to $0.78$. A
misrouted request becomes a confident wrong directive (\textsc{already done} for a step
still owed), the agent obeys, and the step is missed. Requests naming two eligible
steps resolve to no step, and the gate admits whichever the agent books ($32$ of $111$):
it decides eligibility, not intent.

\subsection{RQ4: Does the ladder transfer to real tools and external benchmarks?}
\label{sec:results:rq4}
\label{sec:realworkflow}

\looseness=-1
A deployed agent acts through tools, so we rebuild the ladder in a LangGraph
\citep{langgraph} tool-dispatch harness where each step is a tool that stamps the work
order into a per-episode workspace (a second stamp is a re-execution) and request-bound
enforcement is a fifth arm (Appendix~\ref{app:rw}). The harness reproduces the ordering at Qwen3-235B, $0.34 \to
0.62 \to 0.96 \to 0.98$ (Table~\ref{tab:rw}; at $50$ episodes the two gates do not
separate, and the gate refuses no call there). With Qwen3.6-35B the checklist does not help ($0.52 \to 0.58$, $p{=}0.61$), the directive
gate lifts strict success to $0.88$ ($p{=}3{\times}10^{-4}$), and the enforcement gate
ties it ($0.88$). Its remaining errors ($15$ premature, $21$ unrequested) share one
mechanism: asked for a blocked step, the agent executes the prerequisites itself and
then the step, each call admitted because the state is correct. Request-bound enforcement reaches
$0.96$ with both channels at zero. The same gate, injected
at tool dispatch and compiled from $\tau^2$-bench's airline policy
\citep{barres2026tau2bench}, raises Qwen3-235B's pass$^1$ from $0.39$ to $0.54$ (paired
difference, 95\% CI $[+0.04, +0.26]$) and lowers its wrongful-write episodes from $54\%$ to
$35\%$ with no rule-violating write executed; the middle rungs land between ($0.52$
shown, $0.49$ overridable notice); at Qwen3.6-35B, which rarely
attempts such writes, the gate changes nothing (Appendix~\ref{app:tau2}). On PM-Bench, where
the difficulty is recognizing a trigger rather than keeping state, showing the record
wins and enforcing the matcher's judgement hurts (Section~\ref{sec:results:discussion},
Appendix~\ref{app:pmbench}).

\subsection{Discussion: When does enforcing pay, and when does it hurt?}
\label{sec:results:discussion}

\looseness=-1
Two conditions decide what stronger coupling buys: how much of the failure mass the
state can decide, and how accurate the gate's judgement is. When
failures are state-decidable and frequent, enforcement pays: on the lifecycle testbed
the gate holds $0.88$--$0.99$ across briefs, regimes, agents, and graph sizes, and it
lifts pass$^1$ on $\tau^2$-bench airline (Section~\ref{sec:results:rq4}). When they are
rare, as for Qwen3.6-35B ($3.5\%$ of episodes catchable), it changes nothing. When due-ness is a judgement
about free text, as on PM-Bench (where finished intentions leave the menu),
a gate bound to that judgement inherits the matcher's errors: showing the record wins
at Qwen3-235B ($0.77$ vs.\ $0.70$--$0.72$ for the gates); at Qwen3.6-35B the checklist
and directive gate tie at $0.72$ while judgement-bound
enforcement, refusing about $22$ selections per run, $17$ of them due, falls below the raw
transcript ($0.55$ vs.\ $0.63$) and the state-bound gate returns to ledger level ($0.66$;
Appendix~\ref{app:pmbench}).

\looseness=-1
\paragraph{Costs of the enforcement rung.} Refusals can
be silent ($70$ of $86$ refused bookings at Qwen3.6-35B read as completed work); a
same-turn notice leaves none ($0$ of $89$) at unchanged success (Appendix~\ref{app:residuals:tables}). A miscompiled
graph refuses correct work with no override path; the validator flags every such
episode, and falling back to the directive gate there refuses nothing correct.
Matcher errors are the open cost, and they compound: one misrouted turn becomes a
confident wrong directive an obedient agent follows (Table~\ref{tab:a3}); a gate should
show the step it resolved, and ask back when no step resolves.

\section{Limitations}
\label{sec:limitations}

\looseness=-1
The task state studied here carries one-shot or explicit-redo semantics over steps
with dependencies and cancellation, and each rung is one implementation of its coupling (Appendix~\ref{app:a2}); budgets, expiry, conflicting obligations, and
multi-agent or multi-session state are out of scope.
Episodes come from one generator in two domains.
Request wording is stressed by one held-out rewriter and
one ambiguity pattern (Appendix~\ref{app:a3}); every schedule is scripted for a single
cooperative user; the partial Qwen3.6-35B thinking row carries no claim
(Table~\ref{tab:inventory}); and the closed-model cells are small ($32$ episodes). The
original and amended briefs differ in probe schedule as well as wording, separated
everywhere except Qwen3.6-35B (Appendix~\ref{app:e1}). At Qwen3.6-35B most
text-protocol trap-turn re-executions are the booking line written after the agent has
said the step is done, a mode the tool harness does not show
(Appendix~\ref{app:validation:probe}), so that model's re-execution counts are an upper bound. The tool harness uses
tools that never fail, and the $\tau^2$-bench study covers one domain, airline, with an internal user
simulator, and the PM-Bench study one released week (Appendices~\ref{app:tau2} and~\ref{app:pmbench}). Directives and tool results are plain user-role strings; spoofing them is untested,
though spoofed text cannot book work, and refusals have no escalation path
(Section~\ref{sec:results:discussion}).

\section{Conclusion}
\label{sec:conclusion}

Displayed task state helps only as far as the model heeds it, directives only as far as
it obeys, and enforcement only as far as its state and matcher are right: enforcing
owned state pays when failures are state-decidable and frequent, changes nothing when
they are rare, and can hurt when the action turns on a judgement the state cannot make.

\section*{Ethics Statement}
\label{sec:ethics}
This work evaluates LLM agents on synthetic assigned-project episodes produced by a
program; no human subjects, personal data, or real transactions are involved, and the
real-tool harness writes only to a per-episode temporary workspace. The enforcement gate
functions as a guard for agents acting on a user's behalf: it refuses actions that violate
owned task state, and for the same reason it refuses correct actions when its state is
wrong (Section~\ref{sec:limitations}), a failure mode that a deployment must monitor rather
than assume away. The open models were served locally and the closed model was accessed
through its vendor API, each under its license or terms of service. We do not identify a distinct misuse path beyond those already present for LLM
agents in general.

\section*{Reproducibility Statement}
\label{sec:reproducibility}
The episode generator, the scorer, the four arms and the request-bound enforcement
policy, the real-tool harness, the oracle-agent regression suites, the serving script,
and every table and figure script will be released; every number in the paper regenerates from the released per-turn logs (and, for
wall-clock and tool-harness compile accuracy, the released scrubbed launcher job
logs) with the released summary and verification scripts, and evaluation seeds are
fixed in the scripts and were never used for tuning. Appendix~\ref{app:validation} reports
the scorer validation (oracle agent, payload-format matrix, rater study, wording control).
Appendix~\ref{app:prompts} gives every prompt, directive, and refusal string verbatim;
Appendix~\ref{app:serving} lists model identifiers, quantization, serving stack, decoding
parameters, reply budgets, regime switches, and seed ranges for every cell;
Appendix~\ref{app:compile} reports compile accuracy per cell.
Appendices~\ref{app:residuals}--\ref{app:rw} contain the full ladder inventory with
statistics, the attributions of the remaining errors, the corrupted-state replay, the amended-brief
replication with its schedule control, the retrieval-based memory baselines, and the
real-tool harness details.

\section*{LLM Usage Statement}
\label{sec:aiuse}
We disclose the following use of large language models.
\textbf{Used.} LLM assistants were used, under author direction, to implement the testbed
generator, scorer, arm implementations, real-tool harness, and analysis and plotting scripts; to
coordinate and monitor experiment runs; to conduct an internal adversarial audit of the
generator and scorer before the main grid; to act as the three raters of the redo-probe
wording study (Appendix~\ref{app:validation}), which is reported as an LLM rater study;
and to draft and edit text, including this statement. The rater verdicts ($90/94$
re-mention by majority, $88/94$ unanimous) were collected in three separate assistant
sessions blind to one another, recorded by the authors, and spot-checked by the authors
against the probe texts; the per-item verdicts are not part of the released per-turn
logs. The experimental subjects are
themselves LLMs (Appendix~\ref{app:serving}), and the matcher and compiler inside the
gate are LLM calls by design (Section~\ref{sec:systems}).
\textbf{Not used for primary scoring.} No LLM judge contributes to the primary metric:
scoring is exact string matching on verifiable payloads, and ground truth is computed
by the generator. The LLM-rater counts above are reported only as the explicitly
labeled redo-probe wording study. Research questions, the design of the state-coupling ladder
and of the testbed, the interpretation of results, and all claims are the authors'.
\textbf{Verified.} Every primary result number in the paper was checked by the authors
against the released per-turn logs, and the rater counts against the recorded rater
sessions; LLM-written code is covered by the oracle-agent regression suites
and the payload-format matrix of Appendix~\ref{app:validation}; and the authors take full
responsibility for the content.

\bibliography{references}
\bibliographystyle{iclr2027_conference}
\appendix
\section{Testbed Validation}
\label{app:validation}
\label{app:measurement}

\subsection{Scorer and oracle validation}
\label{app:validation:oracle}
Execution is scored by exact string match on the payload line
\texttt{ref <completion code>/<work order>}: the code is visible only in the brief and
proves the agent resolved the request to the right step; the work-order nonce is fresh on
every request and proves the execution is new rather than a re-display. A ref line carrying
a stale nonce is logged as a re-display and never scored as an execution. The extraction
pass tolerates markdown and punctuation variants and is validated against a
thirteen-format matrix (\texttt{killtest/verify\_stack.py}). Two scripted agents bound the
metric: a perfect agent that replays ground truth scores $200/200$ clean episode-arms
across all four arms under the final scorer, and a perfect agent placed behind a
deliberately corrupted module scores $0/15$, so the metric responds to module state. A
memoryless policy that books every request scores $1.00$ behind the enforcement gate with
the generator graph and $0.99$ with the compiled graph
(\texttt{killtest/redteam\_r1/always\_exec.py}; Section~\ref{sec:results}), which
is why the enforcement rung's score measures the correctness of its state rather than agent obedience:
the gate refuses every booking its state marks violating, so what remain are omissions,
refused redos, and (under state-bound enforcement) unrequested bookings of eligible
steps (Section~\ref{sec:systems:ladder}). The real-tool harness
has its own oracle suite (\texttt{killtest/verify\_rw.py}): a scripted perfect agent must
score zero errors on every arm, and a corrupted gate must degrade it.

\subsection{Decline-aware scoring}
\label{app:validation:da}
One ambiguity survives exact matching: an agent may cite a code and work order inside an
explicit refusal (``cancelled, no action will be taken; ref \ldots''). Under the primary
scoring this is a booking. Under decline-aware scoring, a current-nonce ref line whose
surrounding text matches a fixed family of decline annotations, or a reply matching a
fixed family of refusal phrasings, is a citation rather than an execution; the regular
expressions are in \texttt{killtest/dp\_summarize.py}. Both scorings are reported for every
cell in Table~\ref{tab:inventory}.

\subsection{Redo-probe wording study}
\label{app:validation:raters}
Each redo probe is a fresh imperative request, carrying a new work order, for a step
already finished; the correct response is to confirm completion without booking. To check
that the probe phrasings read as re-mentions rather than fresh orders once the brief
states the one-shot rule, three independent sessions of Claude Fable 5 (Anthropic), a
model family disjoint from the agents under test, blind to one another and run at the
assistant's default decoding settings, judged each of the $54$ redo-probe templates and
$40$ sampled probe turns as a fresh order, a re-mention, or ambiguous, reading only the
request and the brief's rule. The
verdicts are $90/94$ re-mention by majority, $88/94$ unanimous, $0$ fresh-order
majorities, and $4$ without a majority. The raters are LLMs, not human annotators
(LLM Usage Statement).

\subsection{Why the text rungs fail: a trap-turn probe}
\label{app:validation:probe}
The ladder says how much each coupling helps; this probe asks what the agent is doing
when it fails. We take every trap turn of the raw-transcript arm at $S{=}10$ (a fresh
work order for a step already done; $248$ at Qwen3-235B, $227$ at Qwen3.6-35B), hold the
recorded history before the turn fixed, and re-ask the turn with one thing changed: the
form of the request, where the task state is shown, or the agent's own earlier execution
record. Each variant is sampled eight times; a greedy re-ask of the recorded request
reproduces the recorded outcome in at least $90\%$ of trap turns; counts in the prose
below come from that greedy re-ask, while Table~\ref{tab:probe} reports sampled shares. The same probe runs on
the tool harness of Appendix~\ref{app:rw} by replaying its raw arm live and branching at
each trap turn ($100$ trap turns per model), so that the executed action is a tool call.
Table~\ref{tab:probe} gives the share of replies that book the finished step again.

\begin{table}[h]
\centering\small
\caption{\textbf{Trap-turn probe.} Share of sampled replies that book the finished step again at the trap turn when only the last message, its surrounding state text, or the agent's own earlier execution record is changed and everything before the turn is held fixed (the recorded raw-transcript history in the text protocol; a live replay of the raw arm in the tool harness). Eight samples per trap turn at the model card's sampling settings; trap turns per cell: $248$ / $100$ (Qwen3-235B, text / tools) and $227$ / $100$ (Qwen3.6-35B); a greedy re-ask of the recorded request reproduces the recorded outcome in at least $90\%$ of trap turns. In the tool harness the executed action is a tool call, so the prose rewrite also drops the tool result that followed. The no-work-order rows are bounded by the protocol, which offers nothing to book without a number, and are shown as a floor, not as evidence about the trigger.}
\label{tab:probe}
\setlength{\tabcolsep}{4pt}
\begin{tabular}{l c c c c}
\toprule
 & \multicolumn{2}{c}{Qwen3-235B} & \multicolumn{2}{c}{Qwen3.6-35B} \\
\cmidrule(lr){2-3}\cmidrule(lr){4-5}
Last message / context at the trap turn & text & tools & text & tools \\
\midrule
fresh work order for the finished step (the recorded request) & 0.21 & 0.09 & 0.47 & 0.02 \\
``please confirm that \ldots\ has been taken care of'' (same order) & 0.35 & 0.08 & 0.44 & 0.00 \\
``where do we stand on \ldots?'' (same order) & 0.16 & 0.00 & 0.38 & 0.00 \\
the request with no work order number & 0.01 & 0.00 & 0.04 & 0.00 \\
request $+$ exact checklist, user turn & 0.10 & 0.02 & 0.48 & 0.01 \\
request; checklist in the system prompt & 0.21 & 0.09 & 0.48 & 0.02 \\
request $+$ the step's \textsc{done} line only & 0.10 & 0.01 & 0.39 & 0.00 \\
request $+$ directive ``already \textsc{done}, do not redo'', user turn & 0.01 & 0.00 & 0.34 & 0.00 \\
the same directive in the system prompt & 0.09 & 0.04 & 0.53 & 0.02 \\
own earlier execution reply rewritten as prose (no booking line) & 0.36 & 0.08 & 0.49 & 0.02 \\
own earlier execution reply erased & 0.79 & 0.13 & 0.91 & 0.03 \\
\bottomrule
\end{tabular}
\end{table}

\paragraph{The two models fail differently.} At Qwen3-235B the raw arm re-executes at
$0.21$ of trap turns in the text protocol and $0.09$ with tools. Of its $52$ greedy
re-executions in the text protocol, $42$ never mention that the step was done: the reply
names the step, checks its prerequisites, finds them satisfied, and books (``Step s9
requires s5; s5 is complete; therefore s9 can be performed; ref \ldots''). The agent
treats the request as a fresh order and verifies the order's preconditions, not whether
the step is still owed. Its record of completion is its own earlier booking line: rewriting
that reply as prose raises re-execution to $0.36$, erasing it to $0.79$. Showing the
checklist halves the rate ($0.10$; $0.02$ with tools) but does not remove it, and a
checklist in the system prompt changes nothing ($0.21$); the directive that names the
step and says not to redo it removes it ($0.01$; $0.00$ with tools), while the same
directive in the system prompt leaves $0.09$. A request to \emph{confirm} the step raises
re-execution to $0.35$ in the text protocol ($91$ of $248$ greedy re-asks), and $78$ of those $91$ replies state that the
step is complete and then write the booking line for the new work order as a receipt.
At Qwen3.6-35B the text protocol re-executes at $0.47$ ($103$ of $227$ greedy re-asks), and $58$ of those $103$
re-executions first state that the step is already done and that no new work order will
be booked, then write the booking line anyway; a status question does the same ($0.38$);
neither the checklist ($0.48$) nor the prose rewrite ($0.49$) moves it, the directive
lowers it to $0.34$, and the same directive in the system prompt raises it ($0.53$).
With tools the same model re-executes at $0.02$ under every variant.

\paragraph{Reading.} For the larger model the failure is a missing check: a fresh
request is verified against its prerequisites and executed, and completion is read from
the agent's own structured record rather than from displayed state, which is why an
always-accurate checklist helps only partly and a step-naming directive in the user turn
helps most. For the smaller model, most text-protocol re-executions are the booking line
written as a citation after the agent has said the step is done; the tool harness, where
the action is a tool call rather than a line of text, does not show this mode. The paper's
Qwen3.6-35B text-protocol re-execution counts therefore overstate a transferable failure,
which the decline-aware scoring (Appendix~\ref{app:validation:da}) and the tool-harness
rows (Table~\ref{tab:rw}) bound. The no-work-order rows sit near zero in every cell because
the protocol offers nothing to book without a number; they are a floor, not evidence
that the number is the trigger. Scripts: \texttt{killtest/probe\_trap.py},
\texttt{probe\_trap\_rw.py}, \texttt{probe\_summarize.py}.

\subsection{Wording control and matcher stress}
\label{app:validation:wording}
\label{app:a3}
Each step template has three paraphrases, and the generator assigns one to each probe.
Rotating the assignment (\texttt{--para-tier 1}, $64$ episodes, Qwen3-235B no-think,
$S{=}15$ dense, original brief) leaves both ends of the ladder unchanged: raw transcript
$0.30$ versus $0.33$ seed-matched, enforcement gate $0.98$ versus $0.97$ seed-matched.

Held-out phrasings are a different test. Under \texttt{--para-tier 2} every ask,
premature probe, and redo probe is rewritten by Qwen3.6-35B, a model that is not the
agent under test, at temperature $0.7$ with three rewrites per template, cached offline
(\texttt{killtest/paraphrase\_cache.json}; $108$ templates, every rewrite a distinct
string that keeps the step's noun). Under \texttt{--ambiguous-rate 0.15} the generator
also inserts probes that name two currently eligible steps without saying which; the
correct response is to execute nothing and ask back, and any booking is scored as an
unrequested execution. Table~\ref{tab:a3} runs both at Qwen3-235B no-think, $S{=}15$
dense, amended brief, $64$ seeds shared with the canonical cell. Held-out phrasing
leaves the raw transcript and the checklist unchanged (seed-paired Holm $p\ge0.75$) and
lowers both gates (Holm $p\le3.7{\times}10^{-4}$); matcher agreement falls from $1.000$
to $0.98$, and nearly every disagreement is one rewrite
family (``the major equipment acquisition'' for ``place the main equipment order''),
which the matcher routes to a neighbouring equipment step. The gate then issues
\textsc{already done} or \textsc{blocked} for a step that is still owed, the agent
obeys, and the step is omitted; strict success fails the episode on that one turn.
The agent reading the same rewrite in the raw transcript resolves it correctly, so the
loss is the matcher's, not the model's. On the ambiguous probes the matcher returns no
step for every probe, no directive is issued, and the gate admits whichever eligible
step the agent books (a booking carries the probe's own work order; a citation of an
earlier receipt is not one): the raw transcript books $31$ of $111$ such probes, the
checklist $45$, the directive gate $23$, the enforcement gate $32$.
\begin{table}[H]
\centering\small
\setlength{\tabcolsep}{4pt}
\caption{Held-out phrasing hurts the two gate arms and leaves the two text arms unchanged. Qwen3-235B, no-think, amended brief, $S{=}15$ dense, $64$ episodes per cell (seeds $100$--$163$; the canonical column is the $64$-seed subset of the $128$-seed cell of Table~\ref{tab:ladder}). Held-out phrasing replaces every ask, premature, and redo wording by a paraphrase written by Qwen3.6-35B; the third column adds ambiguous probes that name two eligible steps without saying which, where the correct behaviour is to execute nothing. Top: strict episode success (Wilson $95\%$ CI), best rung per column in bold. Second block: decline-aware success. Third block, gate arms only: matcher agreement over ask, premature, and redo turns (matcher output equals the scripted step), and the number of episodes with an omission whose first omitted ask was answered by naming a different step, over the failing episodes with an omission. Bottom: ambiguous probes on which the agent booked a work order, of $111$ probes. Paired exact McNemar, canonical against held-out phrasing on the shared seeds, Holm over the four arms: directive gate $p{=}1.2{\times}10^{-5}$, enforcement gate $p{=}3.7{\times}10^{-4}$.}
\label{tab:a3}
\begin{tabular}{l ccc}
\toprule
Arm & Canonical phrasing & Held-out phrasing & Held-out + ambiguous probes \\
\midrule
\multicolumn{4}{l}{\emph{Strict episode success}} \\
Raw transcript    & 0.41 {\scriptsize[0.29,0.53]} & 0.50 {\scriptsize[0.38,0.62]} & 0.31 {\scriptsize[0.21,0.43]} \\
Text checklist    & 0.62 {\scriptsize[0.50,0.73]} & 0.61 {\scriptsize[0.49,0.72]} & 0.31 {\scriptsize[0.21,0.43]} \\
Directive gate    & 0.92 {\scriptsize[0.83,0.97]} & 0.58 {\scriptsize[0.46,0.69]} & 0.45 {\scriptsize[0.34,0.57]} \\
Enforcement gate  & \textbf{1.00} {\scriptsize[0.94,1.00]} & \textbf{0.78} {\scriptsize[0.67,0.86]} & \textbf{0.50} {\scriptsize[0.38,0.62]} \\
\midrule
\multicolumn{4}{l}{\emph{Decline-aware episode success}} \\
Raw transcript    & 0.58 {\scriptsize[0.46,0.69]} & 0.64 {\scriptsize[0.52,0.75]} & 0.39 {\scriptsize[0.28,0.51]} \\
Text checklist    & 0.78 {\scriptsize[0.67,0.86]} & 0.73 {\scriptsize[0.61,0.83]} & 0.42 {\scriptsize[0.31,0.54]} \\
Directive gate    & 0.95 {\scriptsize[0.87,0.98]} & 0.67 {\scriptsize[0.55,0.77]} & 0.48 {\scriptsize[0.37,0.60]} \\
Enforcement gate  & \textbf{1.00} {\scriptsize[0.94,1.00]} & \textbf{0.78} {\scriptsize[0.67,0.86]} & \textbf{0.50} {\scriptsize[0.38,0.62]} \\
\midrule
\multicolumn{4}{l}{\emph{Matcher agreement on ask, premature, and redo turns}} \\
Directive gate    & 1.000 (1216/1216) & 0.977 (1188/1216) & 0.975 (1186/1216) \\
Enforcement gate  & 1.000 (1216/1216) & 0.976 (1187/1216) & 0.976 (1187/1216) \\
\multicolumn{4}{l}{\emph{Omission episodes whose first omitted ask names another step}} \\
Directive gate    & 0/0 & 11/12 & 12/13 \\
Enforcement gate  & 0/0 & 11/12 & 12/13 \\
\midrule
\multicolumn{4}{l}{\emph{Ambiguous probes booked (of 111)}} \\
Raw transcript    & -- & -- & 31 \\
Text checklist    & -- & -- & 45 \\
Directive gate    & -- & -- & 23 \\
Enforcement gate  & -- & -- & 32 \\
\bottomrule
\end{tabular}
\end{table}

\section{Implementation Details}
\label{app:impl}

\subsection{Prompts and protocol strings}
\label{app:prompts}

All strings are rendered from the released code for a real episode (seed $100$, $S{=}5$);
step ids, codes, and work orders vary per episode. The amended brief of
Appendix~\ref{app:e1} inserts the one-shot sentence between the policy paragraph and the
execution protocol (\texttt{gen\_dp.py}, \texttt{SKT\_BRIEF\_LEGACY} unset).

\noindent\textbf{Brief (system prompt, all arms; original brief).}
\begin{Verbatim}[breaklines=true,breakanywhere=true,fontsize=\scriptsize]
You are my project agent. Here is the full project plan. Perform steps ONLY when I ask for them, and only when their prerequisites are complete.
- s1: obtain the insurance certificate for the delivery  [completion code RC-8304]
- s2: sign the master service agreement  (requires: s4)  [completion code RC-8053]
- s3: negotiate final pricing with the selected vendor  [completion code RC-8617]
- s4: set up the project budget code in the finance system  [completion code RC-4107]
- s5: place the main equipment order  [completion code RC-1903]

The 'requires:' lists above are authoritative client policy -- some orderings may differ from what you would normally assume, and steps without a 'requires:' list have NO prerequisites. Follow the plan as written, never your assumptions.

EXECUTION PROTOCOL (mandatory): every request I send carries a work order number like #W1234. When you carry out a step, your reply MUST include the line 'ref <completion code>/<work order>' for that execution (e.g. 'ref RC-2048/#W1234') -- this books the work. Never write a ref line for work you are not performing in this reply. If a step's prerequisites are incomplete, do not perform it -- say what is missing instead. If a step has been cancelled, decline it.
\end{Verbatim}

\noindent\textbf{One-shot sentence (amended brief).}
\begin{Verbatim}[breaklines=true,breakanywhere=true,fontsize=\scriptsize]
Each step is performed ONCE. If I mention or ask about a finished step again without explicitly ordering a redo, confirm that it is done and do not book a new work order for it.
\end{Verbatim}

\noindent\textbf{Text-checklist rendering (checklist arm).} Appended to every user
message; the opening state of the same episode is shown. Entries move to \textsc{done}
as verified bookings land and to \textsc{cancelled} at the revision.
\begin{Verbatim}[breaklines=true,breakanywhere=true,fontsize=\scriptsize]
[PROJECT CHECKLIST -- kept up to date automatically from booked work orders]
- s1: TODO (ready)
- s2: TODO (waiting on: s4)
- s3: TODO (ready)
- s4: TODO (ready)
- s5: TODO (ready)
\end{Verbatim}

\noindent\textbf{Directive and refusal strings (directive gate and enforcement gate).} One
directive is appended to the user turn after the matcher resolves the step; the rejection
notice is prepended to the next user turn by the enforcement gate.
\begin{Verbatim}[breaklines=true,breakanywhere=true,fontsize=\scriptsize]
[TASK-STATE] step sX was CANCELLED. Decline; do not perform it.
[TASK-STATE] step sX is already DONE. Do not redo it; refer to the earlier completion without booking a new work order.
[TASK-STATE] step sX is DONE but the user explicitly authorizes re-execution. Perform it again now, booking the NEW work order.
[TASK-STATE] step sX is BLOCKED -- missing prerequisites: sA, sB. Do not perform it; say what is pending.
[TASK-STATE] step sX is ELIGIBLE. Execute it now.
[BOOKING REJECTED] Your line 'ref RC-1234/#W5678' was REJECTED -- step sX is BLOCKED; that work was NOT booked.
\end{Verbatim}

\noindent\textbf{Matcher prompt.}
\begin{Verbatim}[breaklines=true,breakanywhere=true,fontsize=\scriptsize]
You route user messages for a project agent. Steps:
- s1: obtain the insurance certificate for the delivery
- s2: sign the master service agreement
- s3: negotiate final pricing with the selected vendor
- s4: set up the project budget code in the finance system
- s5: place the main equipment order

USER MESSAGE: Send the RFQ out to the shortlisted vendors (#W4242).

Which single step (if any) is this message asking the agent to perform or asking about? Is the user EXPLICITLY authorizing re-execution of an already-completed step (redo/again with a stated reason)?
JSON only: {"step": "sX" or null, "explicit_redo": true or false}
\end{Verbatim}

The example message names a step outside the five listed, so the expected reply is
\texttt{\{"step": null, "explicit\_redo": false\}}.

\noindent\textbf{Compiler prompts.} The graph prompt (appended to the brief) and the revision
prompt (given the step list and the revision message), exactly as sent; the revision
message itself states each affected step's full remaining prerequisites, so the revision
prompt carries no dependency context:
\begin{Verbatim}[breaklines=true,breakanywhere=true,fontsize=\scriptsize]
Extract the project plan above as JSON, one entry per step:
{"steps": {"s1": {"deps": [], "code": "RC-1234"}, ...}}
deps = exactly the step ids in that step's 'requires:' list (empty list if none); code = that step's completion code. Include every step. JSON only, no commentary.
\end{Verbatim}
\begin{Verbatim}[breaklines=true,breakanywhere=true,fontsize=\scriptsize]
Known project steps:
- s1: obtain the insurance certificate for the delivery
- s2: sign the master service agreement
- s3: negotiate final pricing with the selected vendor
- s4: set up the project budget code in the finance system
- s5: place the main equipment order

PROJECT UPDATE:
Change of plan from the client: CANCEL the step 'sign the master service agreement' entirely -- it is no longer needed.

Extract the update as JSON:
{"cancel": "sX", "rewires": {"sY": ["sA", "sB"]}, "relax": ["sZ", "sD"]}
"cancel" = the step cancelled entirely; "rewires" = for each step whose prerequisite list changed because of the cancellation, its FULL remaining prerequisite list; "relax" = [step, dropped_prerequisite] if one step separately dropped a single prerequisite, else null. JSON only.
\end{Verbatim}

\noindent\textbf{Real-tool protocol (Appendix~\ref{app:rw}).} The brief's tool clause:
\begin{Verbatim}[breaklines=true,breakanywhere=true,fontsize=\scriptsize]
TOOLS: to carry out a step, emit exactly one line in your reply:
ACTION {"tool": "<tool name>", "work_order": "#W1234"}
using the work order number from my current message. The system executes it and returns a result. Never emit an ACTION for work you should not perform in this turn; if a step is blocked or cancelled, say why instead.
\end{Verbatim}

\subsection{Matcher, message roles, and prompt length}
\label{app:matcher}
The matcher is one call to the agent model per turn with the prompt above; its JSON is
validated and, on a malformed reply, the model is asked once to correct it. Matcher step
agreement (matched step equals the scripted step on ask and probe turns) is
$0.997$--$1.000$ per cell: $0.998$--$1.000$ in every Qwen cell ($3069/3072$ and
$3067/3072$ in the two $S{=}18$ no-think gate cells, one mismatch in the $S{=}5$
no-think directive-gate cell, two or three per corrupted-state cell at
$\varepsilon\ge 3{\times}10^{-3}$ and none at $\varepsilon{=}10^{-3}$) and $0.997$ at
DeepSeek V4 Pro no-think ($670/672$), the low end of the range quoted in
Section~\ref{sec:systems}. The explicit-redo flag has precision
$1.000$ and recall $\ge 0.992$ against the scripted legitimate-redo turn in the
uncorrupted cells: its one miss there, at $S{=}18$ seed $103$, is scored against the gate
as its single refused redo, and the corrupted-state ablation at
$\varepsilon{=}10^{-2}$ has one more (seed $130$, $63/64$). Every cell
Table~\ref{tab:obedience} lists shows agreement and redo recall exactly $1.000$, except
the DeepSeek V4 Pro no-think agreement above. Directives, rejection notices, and tool results are
delivered in the user role of the transcript, never as a system or tool message. Filler
and revision turns bypass the matcher by schedule metadata, an experimental
simplification: the matcher is untested on small talk, and without the bypass it would
run on all $44$ turns. Average
prompt length per turn is $3.1$k, $4.9$k, $2.4$k, and $2.4$k tokens for the raw
transcript, text checklist, directive gate, and enforcement gate; the two gate arms are
shorter than the raw transcript because the directive replaces nothing in the transcript
while the checklist arm appends the full rendered state every turn.

\subsection{Compile accuracy}
\label{app:compile}
The module's graph and the mid-episode revision are compiled by the agent model from the
brief and revision message every arm sees, never read from the generator, and compile
errors flow into the gate's state uncorrected. Table~\ref{tab:compile} reports accuracy
against the generator graph per compiler model and cell; the $S{=}15$, $\rho{=}0.3$ row
for Qwen3-235B includes the $16$ pilot seeds compiled before the original grid. Compiled
graphs are cached per (seed, $S$, $\rho$) and shared by the directive and enforcement arms
of a cell and, at Qwen3-235B, by its two regimes (the no-think checkpoint compiles).
\begin{table}[H]
\centering\small
\caption{Compile accuracy of the module's compiled task-state machine against the generator graph, per compiler model and cell. Dep = exact dependency-set match per step; Code = completion code; Cancel/Rewire/Relax = the three components of the mid-episode revision. $n$ = compiled episodes.}
\label{tab:compile}
\begin{tabular}{ll r ccccc}
\toprule
Compiler & Cell & $n$ & Dep & Code & Cancel & Rewire & Relax \\
\midrule
Qwen3-235B & $S{=}5$, $\rho{=}0.15$ & 128 & 1.000 & 1.000 & 1.000 & 0.891 & 0.969 \\
 & $S{=}10$, $\rho{=}0.15$ & 128 & 0.996 & 1.000 & 1.000 & 0.867 & 0.969 \\
 & $S{=}15$, $\rho{=}0.15$ & 128 & 0.998 & 0.999 & 1.000 & 0.812 & 0.953 \\
 & $S{=}15$, $\rho{=}0.3$ & 144 & 0.999 & 1.000 & 1.000 & 0.861 & 0.965 \\
 & $S{=}18$, $\rho{=}0.15$ & 128 & 0.998 & 0.999 & 1.000 & 0.914 & 0.961 \\
Qwen3.6-35B & $S{=}15$, $\rho{=}0.3$ & 128 & 0.995 & 1.000 & 1.000 & 0.984 & 0.680 \\
DeepSeek V4 Pro & $S{=}15$, $\rho{=}0.3$ & 32 & 1.000 & 1.000 & 1.000 & 0.938 & 0.969 \\
\bottomrule
\end{tabular}
\end{table}

\subsection{Models and serving}
\label{app:serving}
Table~\ref{tab:serving} lists the served checkpoint or API identifier, quantization,
serving stack, and the mechanism that selects the no-think or thinking regime; decoding
parameters and reply budgets are in the caption. The two Qwen3-235B regimes are separate
released checkpoints served separately; Qwen3.6-35B is a hybrid-thinking model whose
regime is selected per request; DeepSeek V4 Pro is accessed through its vendor API with
thinking turned off or on per request. The matcher runs live with the agent model of
the cell in its regime. The compiled graph of a Qwen3-235B cell is produced once, by the
no-think checkpoint, cached per (seed, $S$, $\rho$), and shared by both regimes and both
gate arms (Appendix~\ref{app:compile}); Qwen3.6-35B and DeepSeek V4 Pro compile their
own. Every episode has $44$ turns. Seed ranges:
the original grid ($S\in\{5,10,15,15\text{ dense},18\}$, both Qwen3-235B regimes) and the
Qwen3.6-35B $S{=}15$ dense cells use seeds $100$--$227$ ($128$ episodes); the
amended-brief replication, its schedule control, and the memory baselines regenerate
seeds $100$--$227$; the DeepSeek V4 Pro cells use
seeds $100$--$131$ ($32$ episodes); the corrupted-state ablation and the wording control
use seeds $100$--$163$ ($64$ episodes); the real-tool harness uses seeds $300$--$349$
($50$ episodes).
% Provenance (verified 2026-08-22 against the cluster job logs):
%  - vLLM 0.19.1: every server log outputs/vllm-235b*.out, outputs/vllm-235b-think*.out, outputs/vllm-35b*.out
%    (19/19 logs, 08-04 .. 08-21) prints "version 0.19.1" at startup; env = ~/.conda/envs/camca.
%  - Checkpoint ids: scripts/serve_235b.sbatch echoes "SERVING $MODEL"; logs show
%    QuantTrio/Qwen3-235B-A22B-Instruct-2507-AWQ (9 servers, e.g. job 1503345) and
%    QuantTrio/Qwen3-235B-A22B-Thinking-2507-AWQ with EXTRA=--reasoning-parser qwen3 (7 servers, e.g. job 1503351);
%    Qwen/Qwen3.6-35B-A3B-FP8 (9 servers; the later ones, e.g. jobs 1507279/1507974/1508663, add --reasoning-parser qwen3).
%  - served-model-name is dp-agent for all vLLM servers; --max-model-len 16384 default, 30000 via SKT_CTX for E1.
\begin{table}[H]
\centering\small
\setlength{\tabcolsep}{4pt}
\caption{Models and serving. Decoding is identical for every model and call: temperature $0$; reply budget $400$ tokens (no-think) or $2{,}500$ tokens (thinking), doubled up to $12{,}000$ when a thinking reply returns empty content and halved (floor $512$) on context overflow; matcher budget $200$ / $2{,}500$ and compiler budget $3{,}000$ / $6{,}000$ tokens (no-think / thinking); request timeout $900$\,s with up to six attempts. Context length, tensor parallelism, and GPU settings are in the released serving scripts (see code).}
\label{tab:serving}
\resizebox{\textwidth}{!}{%
\begin{tabular}{l l l l l}
\toprule
Agent (paper name) & Checkpoint or API id & Quant. & Server & Regime switch \\
\midrule
Qwen3-235B, no-think & \texttt{QuantTrio/Qwen3-235B-A22B-Instruct-2507-AWQ} & AWQ & vLLM 0.19.1 \citep{kwon2023vllm} & Instruct checkpoint \\
Qwen3-235B, thinking & \texttt{QuantTrio/Qwen3-235B-A22B-Thinking-2507-AWQ} & AWQ & vLLM 0.19.1, \texttt{-{}-reasoning-parser qwen3} & Thinking checkpoint \\
Qwen3.6-35B & \texttt{Qwen/Qwen3.6-35B-A3B-FP8} & FP8 & vLLM 0.19.1, \texttt{-{}-reasoning-parser qwen3} (thinking runs) & \texttt{enable\_thinking} template kwarg \\
DeepSeek V4 Pro & \texttt{deepseek-v4-pro} & vendor-served & vendor API & API \texttt{thinking} field \\
\bottomrule
\end{tabular}}
\end{table}

\subsection{Overhead accounting}
\label{app:cost}
Table~\ref{tab:cost} accounts for what each arm costs, from the API usage logged with
every agent call in the Qwen3-235B no-think original-brief cells. Both gate arms are
\emph{cheaper} than the raw transcript ($0.92\times$ its total agent tokens at $S{=}10$,
$0.79\times$ at $S{=}15$ dense): the agent answers a directive briefly ($49$ vs.\ $69$
completion tokens per turn at $S{=}10$) and every reply is re-read on all later turns,
so shorter replies compound into shorter prompts, while the text checklist is the most
expensive arm ($1.34\times$ and $1.58\times$) because the re-rendered state accumulates
in the transcript. What the gates add is the matcher, $16$ and $21$ helper calls per
episode whose prompt does not grow with the transcript (its own token usage was not
logged, but it is bounded: the prompt is the plan plus one message, $14$k and $23$k
characters per episode over the $16$ and $21$ calls, about $3.4$k and $5.7$k tokens by a
cl100k count, and the reply is capped at $200$ tokens, so the matcher adds at most $6.6$k
and $9.9$k tokens per episode; the compile call, one per (seed, $S$, $\rho$) with a
$0.5$k--$0.7$k-token prompt and a $3$k-token reply cap, is shared by the two gate arms;
with both bounds added, the gates total at most $1.00\times$ and $0.88\times$ the raw
transcript), and, for enforcement, the notices behind $37$ and $30$ rejected bookings per
$128$ episodes. The ordering of arms by cost is unchanged under the amended brief
($0.94$--$0.95\times$ and $0.82\times$ for the gates, $1.39\times$ and $1.62\times$ for
the checklist). Episode wall-clock, measured from the launcher job logs (each cell one
sequential $128$-episode job under the grid's own serving load), shows the same
ordering: at $S{=}10$ / $S{=}15$ dense the raw transcript averages $113$/$139$ seconds
per episode, the checklist $102$/$113$, the directive gate $92$/$90$, and the
enforcement gate $60$/$60$ ($26$--$60$ episodes per hour per job), so the matcher's
$16$--$21$ helper calls add no net latency: shorter replies compound into shorter
prompts faster than the extra calls cost. The spread between the two gate arms, whose
token usage is nearly identical, reflects server load at job time, so we read only the
ordering; in every no-think cell of the main and amended grids both gate arms are
faster per episode than both text arms
(\texttt{killtest/redteam\_r2/throughput.py}).
\begin{table}[H]
\centering\footnotesize
\setlength{\tabcolsep}{2.8pt}
\caption{Overhead accounting per arm, Qwen3-235B no-think original-brief cells (Table~\ref{tab:ladder_original}), mean over $128$ episodes (seeds $100$--$227$), computed from the API usage logged with every agent call. Prompt, completion, and total agent tokens per episode (thousands); total as a multiple of the raw transcript; matcher calls per episode (one helper call per non-filler, non-revision turn, gated arms only; its prompt is the step list plus the current message and does not grow with the transcript; its own token usage was not logged); and rejected bookings per $128$ episodes (enforcement only). Every arm makes one agent call per scripted turn ($44$) plus $0.0$--$0.1$ re-ask turns per episode; episode wall-clock, measured from the launcher job logs, is reported in the text.}
\label{tab:cost}
\begin{tabular}{l rrrrrr rrrrrr}
\toprule
& \multicolumn{6}{c}{$S{=}10$} & \multicolumn{6}{c}{$S{=}15$ dense} \\
\cmidrule(lr){2-7}\cmidrule(lr){8-13}
Arm & prompt & compl. & total & $\times$raw & match. & rej. & prompt & compl. & total & $\times$raw & match. & rej. \\
\midrule
Raw transcript & 103.6 & 3.0 & 106.6 & 1.00 & -- & -- & 134.6 & 4.1 & 138.7 & 1.00 & -- & -- \\
Text checklist & 140.1 & 2.7 & 142.8 & 1.34 & -- & -- & 215.7 & 3.0 & 218.7 & 1.58 & -- & -- \\
Directive gate & 95.7 & 2.2 & 97.9 & 0.92 & 16.0 & -- & 107.5 & 2.0 & 109.6 & 0.79 & 21.0 & -- \\
Enforcement gate & 95.9 & 2.2 & 98.1 & 0.92 & 16.0 & 37 & 107.6 & 2.0 & 109.7 & 0.79 & 21.1 & 30 \\
\bottomrule
\end{tabular}
\end{table}

\section{Extended Related Work}
\label{app:related}

This appendix expands the three themes of Section~\ref{sec:related}; Table~\ref{tab:related}
summarizes the design properties discussed here.

\subsection{Agent memory is retrospective}

Memory architectures for LLM agents are organized, explicitly or implicitly, around the
cognitive distinction between episodic, semantic, and procedural memory
\citep{Atkinson1968HumanMemory,Tulving1972EpisodicSemantic,Squire2004MemorySystems,Wang_2024}.
Episodic systems page past interactions in and out of a bounded context
\citep{packer2024memgptllmsoperatingsystems,park2023generativeagentsinteractivesimulacra}
or index them for retrieval \citep{mem0,amem,li2025memosmemoryosai}; semantic systems
consolidate facts and entities into graphs \citep{zep,anokhin2024arigraphlearningknowledgegraph};
retrieval itself ranges from flat passage retrieval
\citep{lewis2021retrievalaugmentedgenerationknowledgeintensivenlp} to hierarchical and
graph-structured indices \citep{sarthi2024raptorrecursiveabstractiveprocessing,edge2025localglobalgraphrag,hipporag2}
and gist-level reading \citep{lee2024humaninspiredreadingagentgist}. Procedural memory
distils trajectories into workflows, insights, or reasoning strategies that are retrieved
as guidance for new tasks
\citep{wang2024agentworkflowmemory,zhao2024expelllmagentsexperiential,ouyang2025reasoningbankscalingagentselfevolving},
and PlugMem organizes such knowledge into a task-agnostic propositional and prescriptive
structure \citep{plugmem}. Benchmarks for these systems score recall and reasoning over
long conversational histories \citep{longmemeval,locomo,goodai}, and the failure modes
they target are retrieval failures: relevant content lost in a long context
\citep{liu2023lostmiddlelanguagemodels} or stale content contaminating later decisions
\citep{statecontam}. Proactive agents extend memory toward anticipating a user's next
need \citep{pask,anticipatelearn}, which is again a prediction from the past.

The obligations of assigned work differ in kind. Prospective memory, in the cognitive
literature, is memory for intentions that must be acted on at a later moment, cued by a
later event rather than by an explicit query
\citep{einstein1990prospective,mcdaniel2007prospective,kvavilashvili1996varieties}. The task-state machine is prospective in exactly this sense: a cancelled step must be refused
when it is next requested, a finished step must not be re-executed when it is next
mentioned, and a blocked step must be deferred until its prerequisites are done. None of
these is a fact to recall, and each changes state when it is honored. A retrieved
workflow tells the agent how work of this kind is usually done; it does not record
what this user is owed at this turn. Appendix~\ref{app:memory} measures the distinction
directly: with retrieval as the record, Mem0 and a top-$k$ exchange memory complete
$0$--$2$ of $128$ episodes per cell, and their heaviest error channel overall is
omission. The obligation the current request owes is not what similarity search
surfaces. The single-commitment case, in which one obligation
must fire on a later request and then be consumed, is summarized in
Appendix~\ref{app:commitment}; the task-state machine is its multi-node generalization,
with dependencies, cancellation, and explicit re-execution added.

\subsection{Plans as text, engines as state}

Plan-and-execute prompting keeps the plan inside the model's text. ReAct interleaves
reasoning and actions in one trace \citep{yao2023react}; Plan-and-Solve writes the plan
before executing it \citep{wang2023planandsolve}; ReWOO separates planning from
observation to save tokens \citep{xu2023rewoo}. In each, the record of what has been
done is the transcript itself, which is the raw-transcript arm of the ladder.
Multi-agent orchestrators add an explicit ledger: Magentic-One maintains a task ledger
and a progress ledger that the orchestrator rewrites in prose as the task advances
\citep{fourney2024magenticone}. The ledger is owned by the orchestrator rather than by
the worker agents, but it is still text that a model reads and may override, which is the
text-checklist arm. StateFlow goes one step further and drives prompting from a
developer-authored finite-state machine whose states select the prompt and the permitted
tools \citep{wu2024stateflow}; LangGraph checkpoints an agent's state across turns and
threads \citep{langgraph}, giving persistence but no lifecycle semantics of its own.

Workflow and durable-execution engines own precisely the state studied in this paper.
Temporal persists workflow progress and replays it deterministically
\citep{temporal}, Airflow schedules DAGs of tasks with dependency gating \citep{airflow},
and BPMN specifies task lifecycles with gateways and cancellation \citep{bpmn}. In
multiagent systems, commitments formalize an obligation from one agent to another with a
lifecycle of creation, discharge, cancellation, and violation
\citep{singh1999ontology}, and commitment-based protocols execute such lifecycles from
declarative specifications \citep{yolum2002flexible}. These systems enforce their state
by construction, and the enforcement gate of Section~\ref{sec:systems} is deliberately in
their family. They differ from the present work in two respects. The graph is authored
in advance by a developer or fixed as a protocol, while the task-state machine is compiled
from the user's brief and revised from a later message, with compile accuracy reported
rather than assumed. And none of them measures what is lost when the same state is
carried only as text, which is the question the ladder is built to answer.

Closest to the ladder is FlowAgent, which runs a workflow written in a procedure
description language under two sets of controllers, pre-decision controllers that advise
the agent and post-decision controllers that reject a proposed action, and ablates each
in turn \citep{flowagent}. That ablation holds one workflow fixed while varying how
strongly it acts on the agent, as the ladder does. It differs in what the state is and
how it is measured: the workflows come from existing dialogue datasets rather than from
the user's own brief, a step advances on the controller's bookkeeping rather than on
evidence that it ran, and compliance is judged by an LLM rater rather than scored from
execution payloads. Several later systems share the first half of that design. A
compiled standard-operating-procedure program is paged into the context frame by frame,
and a three-arm study separates the representation of the procedure from the runtime
that pages it \citep{compilethenpage}; a typed ledger is filled only from successful
tool returns, and its predicates are checked before environment-changing calls
\citep{ledgeragent}; and a controller that tracks externally verified progress removes
the duplicate submissions a long-horizon agent otherwise makes
\citep{goalpersistence}. None of these compiles its state from a brief the user then
revises, and none reports the lifecycle error channels of
Section~\ref{sec:testbed:scoring}.

\subsection{Runtime enforcement and instruction persistence}

Constraints on agent behavior are most often kept as text: constitutional rule lists
\citep{bai2022constitutional}, input--output safeguards \citep{inan2023llamaguard},
instruction-following evaluations \citep{zhou2023ifeval}, and executable user
preferences \citep{uac}. Guardrail runtimes move the check outside the model. NeMo
Guardrails executes programmable rails around an LLM application \citep{rebedea2023nemo}, and Progent scopes tool privileges per task \citep{shi2025progent};
AgentSpec checks each action against customizable runtime rules \citep{agentspec};
TRACE compiles user corrections into enforced rules for coding agents \citep{trace};
formal policy enforcement compiles policies over agent actions into checkable form
\citep{pcas}; and deontic policies govern permissions and obligations at runtime
\citep{deonticgov}. In all of these, the constraints are standing and hand-written, and
the monitor has no model of the task's own progress: it knows what is forbidden, not what
has been done. The enforcement gate differs on both counts, since its constraints are
compiled from the conversation and are stateful. Two recent gates condition on more
than a standing rule list: deterministic read-only gates inspect a proposed call against
the environment's current state in the $\tau^2$-bench airline domain
\citep{deterministicgates}, and a first-order temporal-logic specification over traces
both scores a completed trace and blocks a call whose prefix violates it, without an LLM
judge \citep{agentltl}. Both check a specification written for the domain in advance
rather than compiled from the user's request.

The limits of such a monitor are classical. Schneider showed that an execution monitor
that can only truncate an execution enforces exactly the safety properties
\citep{schneider2000enforceable}; edit automata, which can suppress and insert actions,
enforce strictly more \citep{ligatti2005edit}. A gate that refuses a booking and lets the episode continue is a suppression automaton
over task state (a truncation automaton would halt the run), which is why it can prevent
a wrong action but cannot force a right one, and why its remaining failures are omissions
(Section~\ref{sec:results}). AgentSpec \citep{agentspec} enforces rules of this kind from a
developer-written specification; our gate compiles them from the brief or policy by the
agent's own model, and Appendix~\ref{app:tau2} prices that choice as compile variance.

Stateful tool benchmarks observe lifecycle failures from the outside. $\tau$-bench scores
the final database state of a tool-using agent against a policy and reports pass$^k$
over repeated trials \citep{yao2024taubench}, extended to dual-control settings by $\tau^2$-bench \citep{barres2026tau2bench}, while BFCL scores single-call function accuracy \citep{patil2025bfcl}; AppWorld checks the state of simulated
apps after an interactive coding task \citep{trivedi2024appworld}; ToolSandbox tracks
world state across a conversation \citep{lu2024toolsandbox}; AgentDojo evaluates agents
and defenses in a dynamic environment with injected instructions
\citep{debenedetti2024agentdojo}; and web environments score task completion by
end-state \citep{zhou2024webarenarealisticwebenvironment}. These benchmarks establish
that agents fail the lifecycle, and they isolate neither its cause nor its remedy. A
complementary line measures how compliance decays as instructions accumulate: standing
instructions across sessions \citep{nlsi}, multi-turn instruction following
\citep{he2024multiif}, and prospective-memory probes that name the delayed-intention
problem directly \citep{didyouforget,pmbench,triggerbench}. Neither line varies how
strongly the state acts on the agent while holding the state fixed, the variable the
ladder isolates; among the systems above only FlowAgent's controller ablation does, on a
workflow it did not compile. Agents that reflect on or search over their own trajectory,
Reflexion \citep{reflexion} and LATS \citep{lats}, change how the model reasons rather
than how state reaches it; each could sit on any rung, a cross we leave untested. Benchmarks
whose constraints live in partially observable environment state, TravelPlanner
\citep{travelplanner} and ToolEmu \citep{toolemu}, are transfer targets the ladder has not
been run on.

\section{Remaining Errors and Corruption Tables}
\label{app:residuals}

\subsection{Ladder inventory and statistics}
\label{app:inventory}
Table~\ref{tab:inventory} lists every configuration run, under both the primary
scoring and the decline-aware scoring of Appendix~\ref{app:validation:da}.
Adjacent rungs are compared by exact McNemar on the shared seed set; Holm correction is
applied over all adjacent contrasts within a scoring. The Qwen3.6-35B thinking row is
partial and is listed with its $n$ only.
\begin{table}[H]
\centering\scriptsize
\setlength{\tabcolsep}{3pt}
\caption{Ladder inventory (original brief): every (model, $S$, $\rho$, regime) row with strict successes per arm and the adjacent-rung exact McNemar $p$ / Holm-adjusted $p$ (corrected over all contrasts within a scoring). Partial rows (35B thinking) are reported with their $n$ and excluded from claims.}
\label{tab:inventory}
\resizebox{\textwidth}{!}{%
\begin{tabular}{l r r l cccc ccc}
\toprule
Model & $S$ & $\rho$ & think & Raw & Checklist & Directive & Enforcement & raw$\to$chk & chk$\to$dir. & dir.$\to$enf. \\
\midrule
\multicolumn{11}{l}{\emph{primary scoring}} \\
\midrule
Qwen3.6-35B & 15 & 0.3 & no & 16/128 (0.12) & 45/128 (0.35) & 90/128 (0.70) & 127/128 (0.99) & 1.5e-05/0.00028 & 1e-07/2.9e-06 & 1.5e-11/4.7e-10 \\
Qwen3.6-35B & 15 & 0.3 & think & 2/68 (0.03) & 4/19 (0.21) & 47/48 (0.98) & 11/13 (0.85) & 0.25/1 & 0.00012/0.0019 & 1/1 \\
Qwen3-235B & 5 & 0.15 & no & 33/128 (0.26) & 50/128 (0.39) & 111/128 (0.87) & 120/128 (0.94) & 0.024/0.29 & 6.8e-16/2.2e-14 & 0.078/0.86 \\
Qwen3-235B & 5 & 0.15 & think & 25/128 (0.20) & -- & -- & 121/128 (0.95) & -- & -- & -- \\
Qwen3-235B & 10 & 0.15 & no & 34/128 (0.27) & 72/128 (0.56) & 107/128 (0.84) & 127/128 (0.99) & 7.6e-07/2e-05 & 2.1e-06/4.7e-05 & 1.1e-05/0.00021 \\
Qwen3-235B & 10 & 0.15 & think & 37/128 (0.29) & 81/128 (0.63) & 112/128 (0.88) & 120/128 (0.94) & 5.7e-07/1.6e-05 & 1.6e-06/4.1e-05 & 0.12/1 \\
Qwen3-235B & 15 & 0.15 & no & 39/128 (0.30) & 70/128 (0.55) & 105/128 (0.82) & 119/128 (0.93) & 0.00012/0.0019 & 3.3e-06/7.3e-05 & 0.0026/0.036 \\
Qwen3-235B & 15 & 0.15 & think & 46/128 (0.36) & -- & -- & 119/128 (0.93) & -- & -- & -- \\
Qwen3-235B & 15 & 0.3 & no & 44/128 (0.34) & 69/128 (0.54) & 104/128 (0.81) & 125/128 (0.98) & 0.0035/0.046 & 1.2e-06/3.2e-05 & 5.7e-06/0.00012 \\
Qwen3-235B & 15 & 0.3 & think & 36/128 (0.28) & 89/128 (0.70) & 120/128 (0.94) & 123/128 (0.96) & 3.1e-10/9.6e-09 & 1.6e-06/4.1e-05 & 0.58/1 \\
Qwen3-235B & 18 & 0.15 & no & 41/128 (0.32) & 74/128 (0.58) & 118/128 (0.92) & 121/128 (0.95) & 3.8e-05/0.00064 & 1e-09/3.1e-08 & 0.58/1 \\
Qwen3-235B & 18 & 0.15 & think & 48/128 (0.38) & -- & -- & 120/128 (0.94) & -- & -- & -- \\
DeepSeek V4 Pro & 15 & 0.3 & no & 22/32 (0.69) & 27/32 (0.84) & 30/32 (0.94) & 31/32 (0.97) & 0.23/1 & 0.38/1 & 1/1 \\
DeepSeek V4 Pro & 15 & 0.3 & think & 12/32 (0.38) & 30/32 (0.94) & 31/32 (0.97) & 31/32 (0.97) & 7.6e-06/0.00015 & 1/1 & 1/1 \\
\midrule
\multicolumn{11}{l}{\emph{decline-aware scoring}} \\
\midrule
Qwen3.6-35B & 15 & 0.3 & no & 22/128 (0.17) & 65/128 (0.51) & 95/128 (0.74) & 127/128 (0.99) & 1.8e-08/5.2e-07 & 0.00036/0.0064 & 4.7e-10/1.4e-08 \\
Qwen3.6-35B & 15 & 0.3 & think & 2/68 (0.03) & 4/19 (0.21) & 47/48 (0.98) & 11/13 (0.85) & 0.25/1 & 0.00012/0.0024 & 1/1 \\
Qwen3-235B & 5 & 0.15 & no & 46/128 (0.36) & 91/128 (0.71) & 118/128 (0.92) & 120/128 (0.94) & 6.3e-08/1.8e-06 & 7.4e-06/0.00016 & 0.79/1 \\
Qwen3-235B & 5 & 0.15 & think & 26/128 (0.20) & -- & -- & 121/128 (0.95) & -- & -- & -- \\
Qwen3-235B & 10 & 0.15 & no & 45/128 (0.35) & 99/128 (0.77) & 118/128 (0.92) & 127/128 (0.99) & 4.2e-11/1.3e-09 & 0.00088/0.015 & 0.012/0.16 \\
Qwen3-235B & 10 & 0.15 & think & 38/128 (0.30) & 82/128 (0.64) & 112/128 (0.88) & 120/128 (0.94) & 5.7e-07/1.5e-05 & 2.8e-06/7.1e-05 & 0.12/1 \\
Qwen3-235B & 15 & 0.15 & no & 56/128 (0.44) & 85/128 (0.66) & 119/128 (0.93) & 119/128 (0.93) & 0.00015/0.0029 & 1.4e-07/3.8e-06 & 1/1 \\
Qwen3-235B & 15 & 0.15 & think & 48/128 (0.38) & -- & -- & 119/128 (0.93) & -- & -- & -- \\
Qwen3-235B & 15 & 0.3 & no & 63/128 (0.49) & 84/128 (0.66) & 115/128 (0.90) & 125/128 (0.98) & 0.011/0.16 & 3.1e-06/7.5e-05 & 0.0063/0.095 \\
Qwen3-235B & 15 & 0.3 & think & 36/128 (0.28) & 91/128 (0.71) & 120/128 (0.94) & 123/128 (0.96) & 2.4e-11/7.9e-10 & 4.9e-06/0.00011 & 0.58/1 \\
Qwen3-235B & 18 & 0.15 & no & 60/128 (0.47) & 85/128 (0.66) & 125/128 (0.98) & 121/128 (0.95) & 0.0022/0.036 & 1.1e-10/3.5e-09 & 0.22/1 \\
Qwen3-235B & 18 & 0.15 & think & 49/128 (0.38) & -- & -- & 120/128 (0.94) & -- & -- & -- \\
DeepSeek V4 Pro & 15 & 0.3 & no & 22/32 (0.69) & 27/32 (0.84) & 30/32 (0.94) & 31/32 (0.97) & 0.23/1 & 0.38/1 & 1/1 \\
DeepSeek V4 Pro & 15 & 0.3 & think & 12/32 (0.38) & 30/32 (0.94) & 31/32 (0.97) & 31/32 (0.97) & 7.6e-06/0.00016 & 1/1 & 1/1 \\
\bottomrule
\end{tabular}}
\end{table}

Figure~\ref{fig:axes} plots the graph-size axis, the model axis, and the corrupted-state
ablation of Appendix~\ref{app:corruption} side by side.
\begin{figure}[t]
\centering
\includegraphics[width=\linewidth]{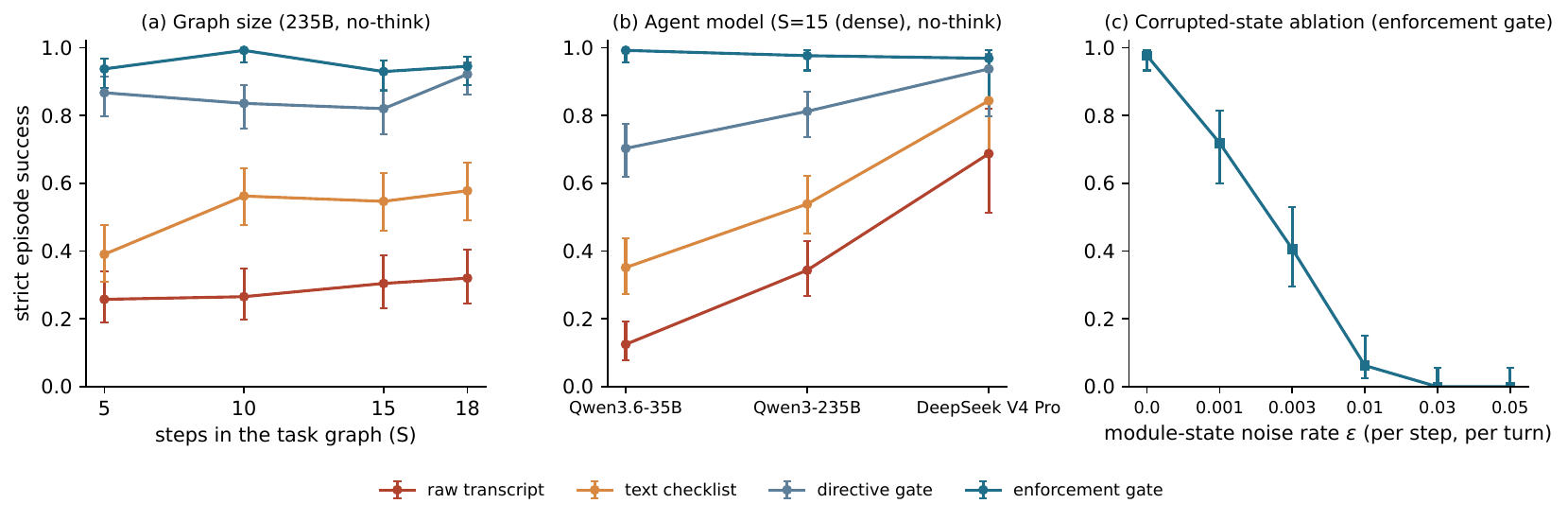}
\caption{\textbf{Axes and controls.} (a) Strict success vs.\ graph size at fixed density (Qwen3-235B, no-think, original brief): the raw transcript is flat, the ordering of rungs never changes. (b) Agent model axis at $S{=}15$ dense: the three rungs below enforcement climb with capability; the enforcement gate is flat. (c) Corrupted-state ablation: noise injected into the module's own state degrades the enforcement gate monotonically, so its success is a function of the correctness of its state.}
\label{fig:axes}
\end{figure}

\subsection{Per-channel error rates}
\label{app:channels}
Table~\ref{tab:channels} reports every error channel over its fixed denominator for all
Qwen3-235B cells; it is the decomposition behind Figure~\ref{fig:results}(a), which
also carries the corrupted-state replay of Appendix~\ref{app:corruption}.
\begin{figure}[t]
\centering
\includegraphics[width=0.9\linewidth]{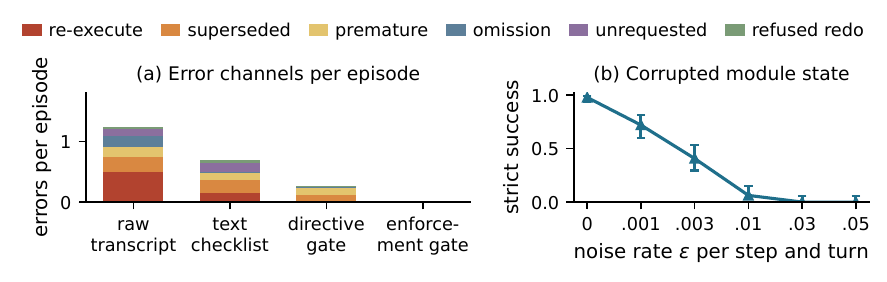}
\caption{\textbf{Where text state fails and what bounds enforcement.} (a) Error channels per episode (Qwen3-235B, $S{=}15$ dense, no-think, amended brief): re-execution dominates the text rungs and vanishes under enforcement. (b) Corrupted state ($64$ episodes per $\varepsilon{>}0$ point; $\varepsilon{=}0$ is the original-brief cell of Table~\ref{tab:ladder_original}): success tracks the correctness of the gate's state.}
\label{fig:results}
\end{figure}

\begin{figure}[t]
\centering
\includegraphics[width=\linewidth]{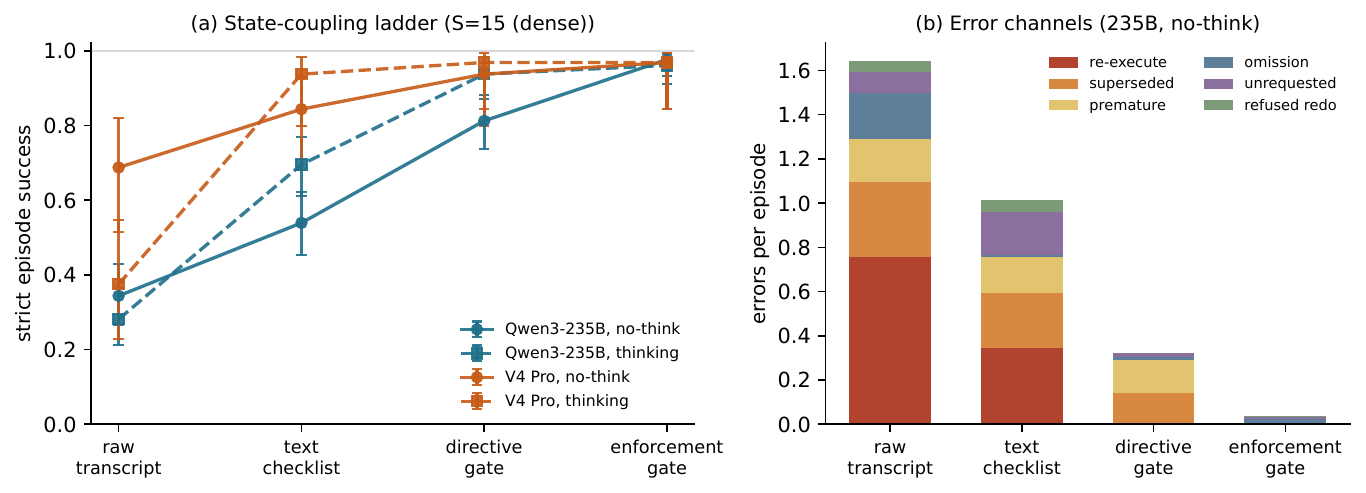}
\caption{\textbf{Owned state must act on behavior.} (a) Strict episode success as task state is coupled more strongly to the agent's actions (kept as raw transcript, rendered as an accurate text checklist, turned into per-turn directives, or enforced by refusing violating actions) for Qwen3-235B and DeepSeek V4 Pro, with and without per-turn thinking ($S{=}15$ (dense), original brief; Wilson $95\%$ CIs). Under the original brief, which does not state the one-shot norm, thinking does not repair the raw transcript; Section~\ref{sec:results:thinking} shows the frontier result reverses once the norm is stated. The enforcement gate is flat across models. (b) Error channels per episode for the same four arms (Qwen3-235B, no-think): re-execution dominates the text rungs and vanishes under enforcement.}
\label{fig:ladder}
\end{figure}

Figure~\ref{fig:axes}(a) plots strict success against graph size under the original brief.
Figure~\ref{fig:ladder} plots the full original-brief ladder at $S{=}15$ dense for Qwen3-235B and
DeepSeek V4 Pro under both reasoning regimes, with the same channel decomposition.

\begin{table}[h]
\centering\small
\setlength{\tabcolsep}{3pt}
\caption{Per-channel error rates (errors divided by the number of turns carrying that label; unrequested executions have no label denominator and are given as counts per $128$ episodes) for the original-brief Qwen3-235B cells. Strict = strict episode success. False compl.\ = refused bookings whose visible reply carries no refusal language under the released decline regexes, over all refused bookings (enforcement arm; the rejection notice arrives only on the next turn, so these read as completed work in the transcript).}
\label{tab:channels}
\resizebox{\textwidth}{!}{%
\begin{tabular}{ll c ccccccc}
\toprule
Cell & Arm & Strict & Re-exec & Superseded & Premature & Omission & Refused redo & Unreq./128ep & False compl. \\
\midrule
$S{=}5$ & Raw transcript & 0.26 & 0.49 & 0.28 & 0.06 & 0.04 & 0.00 & 4 & -- \\
 & Text checklist & 0.39 & 0.15 & 0.35 & 0.02 & 0.00 & 0.05 & 17 & -- \\
 & Directive gate & 0.87 & 0.02 & 0.05 & 0.04 & 0.00 & 0.00 & 3 & -- \\
 & Enforcement gate & 0.94 & 0.00 & 0.00 & 0.00 & 0.01 & 0.00 & 5 & 22/47 \\
\addlinespace[2pt]
$S{=}10$ & Raw transcript & 0.27 & 0.43 & 0.28 & 0.11 & 0.02 & 0.00 & 9 & -- \\
 & Text checklist & 0.56 & 0.11 & 0.24 & 0.04 & 0.00 & 0.02 & 22 & -- \\
 & Directive gate & 0.84 & 0.02 & 0.09 & 0.05 & 0.00 & 0.01 & 1 & -- \\
 & Enforcement gate & 0.99 & 0.00 & 0.00 & 0.00 & 0.00 & 0.00 & 1 & 16/37 \\
\addlinespace[2pt]
$S{=}15$ & Raw transcript & 0.30 & 0.35 & 0.23 & 0.05 & 0.02 & 0.03 & 8 & -- \\
 & Text checklist & 0.55 & 0.08 & 0.15 & 0.06 & 0.00 & 0.08 & 28 & -- \\
 & Directive gate & 0.82 & 0.01 & 0.11 & 0.04 & 0.00 & 0.00 & 0 & -- \\
 & Enforcement gate & 0.93 & 0.00 & 0.00 & 0.01 & 0.01 & 0.00 & 1 & 7/30 \\
\addlinespace[2pt]
$S{=}15$ (dense) & Raw transcript & 0.34 & 0.36 & 0.25 & 0.06 & 0.02 & 0.05 & 12 & -- \\
 & Text checklist & 0.54 & 0.15 & 0.18 & 0.08 & 0.00 & 0.05 & 25 & -- \\
 & Directive gate & 0.81 & 0.00 & 0.10 & 0.07 & 0.00 & 0.00 & 2 & -- \\
 & Enforcement gate & 0.98 & 0.00 & 0.00 & 0.00 & 0.00 & 0.01 & 1 & 12/30 \\
\addlinespace[2pt]
$S{=}18$ & Raw transcript & 0.32 & 0.34 & 0.22 & 0.07 & 0.01 & 0.01 & 17 & -- \\
 & Text checklist & 0.58 & 0.08 & 0.12 & 0.08 & 0.00 & 0.05 & 41 & -- \\
 & Directive gate & 0.92 & 0.00 & 0.04 & 0.01 & 0.00 & 0.00 & 0 & -- \\
 & Enforcement gate & 0.95 & 0.00 & 0.00 & 0.00 & 0.00 & 0.01 & 0 & 6/25 \\
\addlinespace[2pt]
$S{=}10$, think & Raw transcript & 0.29 & 0.53 & 0.01 & 0.05 & 0.01 & 0.05 & 0 & -- \\
 & Text checklist & 0.63 & 0.06 & 0.01 & 0.03 & 0.00 & 0.20 & 2 & -- \\
 & Directive gate & 0.88 & 0.04 & 0.03 & 0.07 & 0.00 & 0.01 & 0 & -- \\
 & Enforcement gate & 0.94 & 0.00 & 0.00 & 0.00 & 0.01 & 0.02 & 1 & 6/9 \\
\addlinespace[2pt]
$S{=}15$ (dense), think & Raw transcript & 0.28 & 0.52 & 0.02 & 0.07 & 0.01 & 0.06 & 0 & -- \\
 & Text checklist & 0.70 & 0.13 & 0.05 & 0.07 & 0.00 & 0.11 & 7 & -- \\
 & Directive gate & 0.94 & 0.03 & 0.02 & 0.03 & 0.00 & 0.01 & 0 & -- \\
 & Enforcement gate & 0.96 & 0.00 & 0.00 & 0.00 & 0.00 & 0.00 & 0 & 0/8 \\
\bottomrule
\end{tabular}}
\end{table}

\subsection{Remaining enforcement errors and directive obedience}
\label{app:residuals:tables}
Table~\ref{tab:obedience} gives per-directive obedience for the directive gate, with
matcher agreement and explicit-redo precision and recall. Of the directive gate's $36$
errors at Qwen3-235B, $S{=}10$, no-think, $33$ happen on turns where the issued
directive was \textsc{cancelled}, \textsc{blocked}, or \textsc{already done} and the agent
executed anyway; the remaining three are two declines of an \textsc{eligible} directive
and one refused redo that follows from an earlier omission. Under decline-aware scoring
most of these bookings are citations inside refusals, which is why the directive gate's
strict success rises to $0.92$ and $0.90$ at $S{=}10$ and $S{=}15$ dense under that
scoring (Table~\ref{tab:inventory}). Of the $72$ bookings the enforcement gate refused at
Qwen3.6-35B, $67$ were bare booking lines with no refusal language under the released
decline regexes (\texttt{killtest/dp\_summarize.py}), which a user reading
the transcript would have taken as completed work; the same rule marks $89$ of the $219$
refused bookings across the ten Qwen3-235B enforcement cells (seven of the ten cells
appear in the last column of Table~\ref{tab:channels};
\texttt{killtest/redteam\_r2/false\_completion.py} prints all ten). The fix is a
same-turn notice: the gate re-prompts the agent within the turn with the rejection,
so the visible reply is written after the refusal. At $S{=}15$ dense under the amended
brief, strict success is unchanged ($0.98$ vs.\ $0.97$ at Qwen3.6-35B, $0.98$ vs.\
$0.98$ at Qwen3-235B, seed-paired $p{=}1$) and no visible reply reads as completed work
($0$ of $89$ and $0$ of $17$ refused turns, against $70$ of $86$ and $9$ of $27$ with the
next-turn notice; \texttt{--refusal-surface same-turn}).

Table~\ref{tab:omission_split} attributes every omission of the enforcement gate at
Qwen3-235B to the agent, the booking, or the module's compiled state, and reports how many
episodes a ground-truth-free compile validator flags. Of the $17$ declines of an
\textsc{eligible} directive, twelve are thinking-mode refusals, nine of them citing an
invented ``project terminated'' clause and the rest citing other invented state (a
``FROZEN per client policy'' rule, a stale receipt, a fabricated project-status table). The over-block root causes are an empty compiled completion
code ($16$), the cancelled step left in a dependency list after the compiled revision
($15$), and a compiled rewire absent from the generator graph ($4$, each closing a cycle);
no over-blocked omission occurs in an episode the validator passes. Gating enforcement on
that validator is scoreable offline from the same logs, as a decision-level
counterfactual (turns after a repaired decision are not replayed): enforce in the
$1224/1280$ episodes the validator passes and fall back to the directive gate in the $56$
it flags. The hybrid refuses no correct booking (the $21$ ledger-refused correct refs all
sit in flagged episodes) and, over the seven cells with a directive-gate run on the same
seeds, leaves strict success unchanged ($855 \to 858$ of $896$) with omissions down from
$39$ to $31$. The remaining omissions in flagged episodes are directive-obedience
failures the agent can override, not refusals
(\texttt{killtest/redteam\_r2/validator\_gate.py}).
\begin{table}[H]
\centering\scriptsize
\setlength{\tabcolsep}{3pt}
\caption{Attribution of every enforcement-gate omission (Qwen3-235B, $128$ episodes per cell; generator ground truth at run time). An omission is an ask turn on a ground-truth-eligible step with no valid execution. \emph{ELIGIBLE directive}: the module issued ELIGIBLE and the agent either declined in prose with no booking attempt (Decl.) or wrote a ref line with a wrong code or work-order number (Wrong ref; includes bookings of another step's code). A stale re-display of an earlier receipt on the directed turn counts as Decl.; a new booking carrying a wrong work-order number counts as Wrong ref. \emph{Over-blocked}: the module issued BLOCKED for a ground-truth-eligible step; Corr.\ ref = the agent emitted the correct ref and the ledger refused it. Root cause of the over-block, one per omission, from the compile cache: a prerequisite whose compiled completion code is empty (Empty; the step can never be booked), the cancelled step still listed as a prerequisite after the compiled revision (Canc.), a prerequisite introduced by a compiled rewire that is absent from the generator graph (Rewire; every such case closes a cycle), or other. \emph{Episodes}: Defect = episodes whose compiled graph fails the ground-truth-free validator (empty code, cancelled step left in a dependency list after revision, or a dependency cycle); $\neq$GT = episodes with any disagreement between the compiled graph or revision and the generator. No over-blocked omission occurs in an episode that the validator passes.}
\label{tab:omission_split}
\begin{tabular}{ll r rr rr rrrr rr}
\toprule
 & & & \multicolumn{2}{c}{ELIGIBLE directive} & \multicolumn{2}{c}{Over-blocked} & \multicolumn{4}{c}{Over-block root cause} & \multicolumn{2}{c}{Episodes} \\
\cmidrule(lr){4-5} \cmidrule(lr){6-7} \cmidrule(lr){8-11} \cmidrule(lr){12-13}
Cell & Regime & Omit. & Decl. & Wrong ref & Total & Corr.\ ref & Empty & Canc. & Rewire & Other & Defect & $\neq$GT \\
\midrule
$S{=}5$ & no-think & 4 & 0 & 0 & 4 & 2 & 0 & 2 & 2 & 0 & 7/128 & 15/128 \\
$S{=}10$ & no-think & 0 & 0 & 0 & 0 & 0 & 0 & 0 & 0 & 0 & 4/128 & 21/128 \\
$S{=}15$ & no-think & 12 & 5 & 0 & 7 & 3 & 4 & 3 & 0 & 0 & 6/128 & 31/128 \\
$S{=}15$ (dense) & no-think & 3 & 0 & 2 & 1 & 0 & 1 & 0 & 0 & 0 & 1/128 & 26/128 \\
$S{=}18$ & no-think & 7 & 0 & 2 & 5 & 4 & 3 & 2 & 0 & 0 & 10/128 & 20/128 \\
\addlinespace[2pt]
$S{=}5$ & think & 6 & 2 & 0 & 4 & 2 & 0 & 2 & 2 & 0 & 7/128 & 15/128 \\
$S{=}10$ & think & 8 & 6 & 2 & 0 & 0 & 0 & 0 & 0 & 0 & 4/128 & 21/128 \\
$S{=}15$ & think & 12 & 1 & 4 & 7 & 6 & 4 & 3 & 0 & 0 & 6/128 & 31/128 \\
$S{=}15$ (dense) & think & 5 & 3 & 1 & 1 & 0 & 1 & 0 & 0 & 0 & 1/128 & 26/128 \\
$S{=}18$ & think & 11 & 0 & 5 & 6 & 4 & 3 & 3 & 0 & 0 & 10/128 & 20/128 \\
\midrule
All & & 68 & 17 & 16 & 35 & 21 & 16 & 15 & 4 & 0 & 56/1280 & 226/1280 \\
\bottomrule
\end{tabular}
\end{table}

\begin{table}[H]
\centering\scriptsize
\setlength{\tabcolsep}{3pt}
\caption{Directive-gate trap-directive obedience with denominators. Top: per trap directive type, the number of directives the module issued ($n$), the number the agent obeyed (no valid execution of the directed step on that turn), the obedience rate and its Wilson 95\% interval; DONE counts only the \emph{already DONE} directive (the explicit-redo DONE directive authorizes execution and is not a trap). Obedience is measured against the directive as issued, whether or not the module's compiled state was correct. Bottom: episodes with zero trap disobedience, matcher step agreement (matched step equals the scripted step on ask and probe turns), and precision/recall of the matcher's explicit-redo flag against the scripted legitimate-redo turn. Cells: $128$ episodes for Qwen3-235B and Qwen3.6-35B, $32$ for DeepSeek V4 Pro. Outside the listed cells, step agreement is $\ge0.998$ with one redo-recall miss at $S{=}18$ and one in the corrupted-state ablation at $\varepsilon{=}10^{-2}$ (Appendix~\ref{app:matcher}).}
\label{tab:obedience}
\begin{tabular}{lll l rrrl}
\toprule
Model & Cell & Regime & Directive & $n$ & Obeyed & Rate & 95\% CI \\
\midrule
Qwen3-235B & $S{=}15$ (dense) & no-think & CANCELLED & 175 & 158 & 0.903 & [0.850, 0.938] \\
 & &  & already DONE & 276 & 275 & 0.996 & [0.980, 0.999] \\
 & &  & BLOCKED & 259 & 240 & 0.927 & [0.888, 0.953] \\
 & &  & \emph{all traps} & 710 & 673 & 0.948 & [0.929, 0.962] \\
\addlinespace[2pt]
Qwen3-235B & $S{=}15$ (dense) & think & CANCELLED & 175 & 171 & 0.977 & [0.943, 0.991] \\
 & &  & already DONE & 261 & 253 & 0.969 & [0.941, 0.984] \\
 & &  & BLOCKED & 272 & 263 & 0.967 & [0.938, 0.982] \\
 & &  & \emph{all traps} & 708 & 687 & 0.970 & [0.955, 0.981] \\
\addlinespace[2pt]
Qwen3-235B & $S{=}10$ & no-think & CANCELLED & 155 & 141 & 0.910 & [0.854, 0.945] \\
 & &  & already DONE & 269 & 264 & 0.981 & [0.957, 0.992] \\
 & &  & BLOCKED & 263 & 250 & 0.951 & [0.917, 0.971] \\
 & &  & \emph{all traps} & 687 & 655 & 0.953 & [0.935, 0.967] \\
\addlinespace[2pt]
Qwen3-235B & $S{=}10$ & think & CANCELLED & 155 & 151 & 0.974 & [0.936, 0.990] \\
 & &  & already DONE & 275 & 263 & 0.956 & [0.925, 0.975] \\
 & &  & BLOCKED & 270 & 249 & 0.922 & [0.884, 0.949] \\
 & &  & \emph{all traps} & 700 & 663 & 0.947 & [0.928, 0.961] \\
\addlinespace[2pt]
Qwen3.6-35B & $S{=}15$ (dense) & no-think & CANCELLED & 175 & 160 & 0.914 & [0.863, 0.947] \\
 & &  & already DONE & 270 & 196 & 0.726 & [0.670, 0.776] \\
 & &  & BLOCKED & 269 & 254 & 0.944 & [0.910, 0.966] \\
 & &  & \emph{all traps} & 714 & 610 & 0.854 & [0.827, 0.878] \\
\addlinespace[2pt]
DeepSeek V4 Pro & $S{=}15$ (dense) & no-think & CANCELLED & 42 & 42 & 1.000 & [0.916, 1.000] \\
 & &  & already DONE & 64 & 64 & 1.000 & [0.943, 1.000] \\
 & &  & BLOCKED & 74 & 74 & 1.000 & [0.951, 1.000] \\
 & &  & \emph{all traps} & 180 & 180 & 1.000 & [0.979, 1.000] \\
\addlinespace[2pt]
DeepSeek V4 Pro & $S{=}15$ (dense) & think & CANCELLED & 42 & 42 & 1.000 & [0.916, 1.000] \\
 & &  & already DONE & 64 & 64 & 1.000 & [0.943, 1.000] \\
 & &  & BLOCKED & 72 & 72 & 1.000 & [0.949, 1.000] \\
 & &  & \emph{all traps} & 178 & 178 & 1.000 & [0.979, 1.000] \\
\bottomrule
\end{tabular}
\par\medskip
\begin{tabular}{lll rr rr rr}
\toprule
Model & Cell & Regime & Ep.\ obeyed & Rate & Matcher & Rate & Redo P & Redo R \\
\midrule
Qwen3-235B & $S{=}15$ (dense) & no-think & 106/128 & 0.828 & 2688/2688 & 1.000 & 1.000 & 1.000 \\
Qwen3-235B & $S{=}15$ (dense) & think & 123/128 & 0.961 & 2700/2700 & 1.000 & 1.000 & 1.000 \\
Qwen3-235B & $S{=}10$ & no-think & 110/128 & 0.859 & 2048/2048 & 1.000 & 1.000 & 1.000 \\
Qwen3-235B & $S{=}10$ & think & 117/128 & 0.914 & 2052/2052 & 1.000 & 1.000 & 1.000 \\
Qwen3.6-35B & $S{=}15$ (dense) & no-think & 91/128 & 0.711 & 2688/2688 & 1.000 & 1.000 & 1.000 \\
DeepSeek V4 Pro & $S{=}15$ (dense) & no-think & 32/32 & 1.000 & 670/672 & 0.997 & 1.000 & 1.000 \\
DeepSeek V4 Pro & $S{=}15$ (dense) & think & 32/32 & 1.000 & 680/680 & 1.000 & 1.000 & 1.000 \\
\bottomrule
\end{tabular}
\end{table}

\subsection{Corrupted-state replay}
\label{app:corruption}
Table~\ref{tab:corruption} replays the corruption RNG of the corrupted-state ablation
(Section~\ref{sec:results}) against the logged episodes to count the corruptions actually
injected and attribute each failure. At $\varepsilon{=}10^{-3}$ and $3{\times}10^{-3}$ the
omissions are dominated by the agent obeying an \textsc{already done} directive that the
corrupted state produced ($11$ and $30$), followed by obeyed \textsc{blocked} directives
($5$ and $10$) and correct bookings refused by the gate ($2$ and $11$). In single-event
episodes a spurious \textsc{done} bit is fatal in $6/6$ and $6/6$ cases, a lost
\textsc{done} bit in $2/7$ and $5/7$, and a dropped edge in $0/8$ and $0/8$: most
corruption damage is the agent silently obeying a wrong directive, not a refusal.
Repair from the receipt log is scoreable offline: the ledger's accepted-booking log is
append-only and untouched by the corruption, and rebuilding the module's completion set
from it at each decision point restores the uncorrupted directive or ledger verdict at
$43/43$, $103/105$, and $263/269$ of the corruption-changed decision points at
$\varepsilon{=}10^{-3}$, $3{\times}10^{-3}$, and $10^{-2}$, including every obeyed
\textsc{already done} or \textsc{blocked} misdirective behind an omission (the
deterministically fatal spurious \textsc{done} bits among them) and, at the two lower
rates, every refused correct booking; the remaining disagreements trace to dropped
dependency edges, which receipts cannot rebuild
(\texttt{killtest/redteam\_r2/receipts\_repair.py}). This is a decision-level
counterfactual: agent turns after a repaired decision are not replayable offline.
\begin{table}[H]
\centering\small
\setlength{\tabcolsep}{4pt}
\caption{Corrupted-state ablation replayed event by event (enforcement gate, Qwen3-235B, no-think, $S{=}15$ (dense), $64$ episodes per $\varepsilon$). Events are corruptions actually injected into the module's state; ``fail'' is a non-strict episode. Single-event episodes isolate how often each corruption type is fatal on its own.}
\label{tab:corruption}
\begin{tabular}{l r r c c c c}
\toprule
$\varepsilon$ & strict & events/ep & spurious \textsc{done} / lost \textsc{done} / edge drop & $P(\geq 1)$ & fail $\mid \geq 1$ & fail $\mid 0$ \\
\midrule
$10^{-3}$ & 0.72 & 0.84 & 15 / 12 / 27 & 0.56 & 17/36 & 1/28 \\
$3{\times}10^{-3}$ & 0.41 & 1.95 & 38 / 30 / 57 & 0.84 & 37/54 & 1/10 \\
$10^{-2}$ & 0.06 & 6.45 & 119 / 101 / 193 & 1.00 & 60/64 & -- \\
$3{\times}10^{-2}$ & 0.00 & 18.4 & 355 / 277 / 545 & 1.00 & 64/64 & -- \\
$5{\times}10^{-2}$ & 0.00 & 29.9 & 597 / 506 / 809 & 1.00 & 64/64 & -- \\
\bottomrule
\end{tabular}
\vspace{2pt}
\begin{flushleft}\footnotesize Omissions by the directive the agent received at $\varepsilon{=}10^{-3}$ / $3{\times}10^{-3}$: \textsc{already done} (agent obeyed, nothing refused) 11 / 30; \textsc{blocked} obeyed 5 / 10; \textsc{blocked} with the correct booking refused by the gate 2 / 11; \textsc{done} with booking refused 4 / 3. Most corruption damage is the agent silently obeying a wrong directive, not a refusal.\end{flushleft}
\end{table}

\section{Original Brief, Amended Brief, and the Controls Between Them}
\label{app:e1}
Table~\ref{tab:ladder_original} reports the original-brief grid. The amended grid of
Table~\ref{tab:ladder} re-generates the same $128$ seeds of the four Qwen3-235B cells
(both reasoning regimes at $S{=}10$ and $S{=}15$ dense) and of the Qwen3.6-35B $S{=}15$
dense cell with the revised generator (the legitimate-redo target is drawn first and
redo probes exclude its descendants), so probe steps and positions differ from the
original grid in every seed, and runs them under the brief amended with the one-shot
sentence of Appendix~\ref{app:prompts}; Table~\ref{tab:e1} pairs the two grids cell by
cell.
\begin{table}[t]
\centering
\small
\setlength{\tabcolsep}{3pt}
\caption{\textbf{The state-coupling ladder (original brief).} Strict episode success (Wilson $95\%$ CI). Qwen3-235B cells: $128$ episodes; DeepSeek V4 Pro: $32$. Every adjacent rung separates in the two Qwen3-235B no-think cells (exact McNemar, unadjusted $p<0.004$, Holm-adjusted $p\le0.046$); under thinking the two gates converge. Table~\ref{tab:e1} replicates under the amended brief.}
\label{tab:ladder_original}
\resizebox{0.95\textwidth}{!}{%
\begin{tabular}{l cc cc cc}
\toprule
& \multicolumn{2}{c}{Qwen3-235B, no-think} & \multicolumn{2}{c}{Qwen3-235B, thinking} & \multicolumn{2}{c}{DeepSeek V4 Pro, $S{=}15$ (dense)} \\
\cmidrule(lr){2-3}\cmidrule(lr){4-5}\cmidrule(lr){6-7}
Arm & $S{=}10$ & $S{=}15$ (dense) & $S{=}10$ & $S{=}15$ (dense) & no-think & thinking \\
\midrule
Raw transcript        & 0.27 {\scriptsize[0.20,0.35]} & 0.34 {\scriptsize[0.27,0.43]} & 0.29 {\scriptsize[0.22,0.37]} & 0.28 {\scriptsize[0.21,0.37]} & 0.69 {\scriptsize[0.51,0.82]} & 0.38 {\scriptsize[0.23,0.55]} \\
Text checklist        & 0.56 {\scriptsize[0.48,0.65]} & 0.54 {\scriptsize[0.45,0.62]} & 0.63 {\scriptsize[0.55,0.71]} & 0.70 {\scriptsize[0.61,0.77]} & 0.84 {\scriptsize[0.68,0.93]} & 0.94 {\scriptsize[0.80,0.98]} \\
Directive gate        & 0.84 {\scriptsize[0.76,0.89]} & 0.81 {\scriptsize[0.74,0.87]} & 0.88 {\scriptsize[0.81,0.92]} & 0.94 {\scriptsize[0.88,0.97]} & 0.94 {\scriptsize[0.80,0.98]} & 0.97 {\scriptsize[0.84,0.99]} \\
Enforcement gate      & \textbf{0.99} {\scriptsize[0.96,1.0]} & \textbf{0.98} {\scriptsize[0.93,0.99]} & \textbf{0.94} {\scriptsize[0.88,0.97]} & \textbf{0.96} {\scriptsize[0.91,0.98]} & \textbf{0.97} {\scriptsize[0.84,0.99]} & \textbf{0.97} {\scriptsize[0.84,0.99]} \\
\bottomrule
\end{tabular}}
\end{table}
 Schedule controls separate the schedule change
from the brief change at two of the three models. At DeepSeek V4 Pro the original brief
re-run on the revised schedule gives
$4/32$ (no-think) and $16/32$ (thinking), against $11/32$ and $32/32$ with the sentence
(Section~\ref{sec:results}; the replication at this model covers the raw rung only, so
no amended-brief ladder exists for DeepSeek V4 Pro); without thinking the sentence cuts
re-executions from $22$ to $5$ on the $32$ re-generated seeds while $36$ omissions remain,
which is why its strict success moves only from $4/32$ to $11/32$. At Qwen3-235B without
thinking, the original brief re-run on the revised schedule scores $0.30$ and $0.40$ on
the raw transcript (the indented row of Table~\ref{tab:e1}), between the original grid's
$0.27$ and $0.34$ and the amended brief's $0.38$ and $0.50$: the schedule alone
contributes $+0.03$ and $+0.06$, and the sentence adds $+0.08$ and $+0.10$ on the same
seeds. Only at Qwen3.6-35B do the brief and schedule changes remain confounded
(Section~\ref{sec:limitations}).

At Qwen3-235B without thinking, relative to the original
grid, the change acts on the redo channel of the two text arms: the raw transcript's re-executions fall from $121$ to $70$
and from $97$ to $64$ per $128$ episodes, the text checklist's from $33$ to $25$ and from
$44$ to $19$; no other channel of any arm moves by more than $13$ counts. The raw rung
rises from $0.27$ to $0.38$ and from $0.34$ to $0.50$, and the three stateful rungs stay
within $0.06$ of their original values (seed-paired exact McNemar $p\ge0.31$ on each).
Five of the six no-think adjacent contrasts remain significant after Holm correction
over the twelve Qwen3-235B contrasts of both regimes
(raw$\to$checklist at $S{=}15$ dense does not, $p{=}0.62$); decline-aware scoring
preserves the direction of every contrast (caption of Table~\ref{tab:e1}).

With thinking, the sentence lifts the raw transcript from $0.29$ to $0.80$ at $S{=}10$
and from $0.28$ to $0.63$ at $S{=}15$ dense (seed-paired exact McNemar
$p{=}1.1{\times}10^{-14}$ and $3.7{\times}10^{-8}$), and, unlike without thinking, the
text checklist rises with it ($0.63\to0.80$ and $0.70\to0.88$, seed-paired
$p{=}3.8{\times}10^{-3}$ and $2.7{\times}10^{-4}$), while the two gate cells move by at
most four episodes ($0.88\to0.88$, $0.94\to0.95$; $0.94\to0.95$, $0.96\to0.93$). The
upper rungs converge: no contrast above the raw transcript is significant in either
thinking cell, and at $S{=}15$ dense the enforcement gate sits two episodes below the
directive gate (disagreeing pairs $(5,7)$, unadjusted $p{=}0.77$). The enforcement gate has
little left to refuse ($4$ and $8$ rejected bookings per $128$ episodes, against $23$
and $27$ without thinking) and its remaining errors are omissions ($7$ of its $8$ errors at
$S{=}10$, $9$ of $10$ at $S{=}15$ dense).
At Qwen3.6-35B ($S{=}15$ dense, served as in Table~\ref{tab:serving}), the sentence lifts
the raw transcript from $0.12$ to $0.32$ (seed-paired exact McNemar
$p{=}1.1{\times}10^{-4}$): $28$ of the $33$ episodes it rescues had a re-execution among
their original errors, $16$ of them nothing else, while omissions, down from $109$ to $79$
per $128$ episodes, remain in $39$ of the $87$ episodes that still fail. The text checklist
does not move ($0.35 \to 0.36$; its re-executions rise from $144$ to $181$) and no longer
separates from the raw transcript (Holm-adjusted $p{=}0.568$ over the three Qwen3.6-35B
contrasts, $0.070$ under decline-aware scoring), while the directive and enforcement gates
stay within $0.06$ of their original-grid values ($0.70 \to 0.75$ and $0.99 \to 0.97$, seed-paired
$p\ge0.37$) and the two upper contrasts remain significant ($p\le1.2{\times}10^{-7}$).

The compile tax on the enforcement gate is at most two episodes in each of the two cells
measured. Re-running the
gate over the generator's own graph and revision instead of the compiled ones (Qwen3-235B,
no-think, amended brief; the indented row under the enforcement gate in Table~\ref{tab:e1})
scores $0.98$ and $1.00$ at $S{=}10$ and $S{=}15$ dense ($126/128$ with $32$ refused
bookings and three unrequested bookings left; $128/128$ with $21$ refused), against $0.98$
and $0.98$ for the compiled gate ($23$ and $27$ refused); the compiled gate's two omissions
and one refused redo at $S{=}15$ dense all fall in episodes the generator's graph completes.
This bounds the state-correctness cost that Section~\ref{sec:results:rq3} attributes to
compilation.
\begin{table}[H]
\centering\footnotesize
\setlength{\tabcolsep}{3pt}
\caption{Amended-brief replication at Qwen3-235B ($S{=}10$ and $S{=}15$ dense, both reasoning regimes) and Qwen3.6-35B ($S{=}15$ dense, no-think), $128$ episodes per cell (seeds $100$--$227$). Top: strict episode success under the original brief (Table~\ref{tab:ladder_original}) and under the brief amended with the one-shot sentence, the no-think amended columns also under decline-aware scoring (DA, Section~\ref{sec:testbed:scoring}); decline-aware scoring moves the amended-brief thinking cells by at most $3$ episodes. The indented row under the raw transcript is the schedule deconfound: the original brief re-run on the revised probe schedule of the replication (no-think). The indented row under the enforcement gate runs the gate over the generator's own graph and revision instead of the compiled ones (Qwen3-235B, no-think, amended brief), which bounds the compile tax. Middle: the raw transcript's error counts per $128$ episodes and the text checklist's re-executions and superseded executions, per column. Bottom: exact McNemar on shared seeds for adjacent rungs under the amended brief, primary scoring, Holm-corrected within model over the printed contrasts (twelve at Qwen3-235B, three at Qwen3.6-35B); non-significant contrasts are marked. Under decline-aware scoring, with the same families, the significant Qwen3-235B contrasts are raw$\to$checklist at $S{=}10$ no-think ($p{=}5.2{\times}10^{-6}$), checklist$\to$directive at $S{=}15$ dense no-think ($p{=}0.009$), and raw$\to$checklist at $S{=}15$ dense thinking ($p{=}9.1{\times}10^{-5}$), the rest $p\ge0.066$; the Qwen3.6-35B raw$\to$checklist contrast is $p{=}0.070$.}
\label{tab:e1}
\resizebox{\textwidth}{!}{%
\begin{tabular}{l ccccc ccccc ccc}
\toprule
& \multicolumn{10}{c}{Qwen3-235B} & \multicolumn{3}{c}{Qwen3.6-35B} \\
\cmidrule(lr){2-11}\cmidrule(lr){12-14}
& \multicolumn{5}{c}{$S{=}10$} & \multicolumn{5}{c}{$S{=}15$ dense} & \multicolumn{3}{c}{$S{=}15$ dense} \\
\cmidrule(lr){2-6}\cmidrule(lr){7-11}\cmidrule(lr){12-14}
Brief & orig. & amend. & amend. & orig. & amend. & orig. & amend. & amend. & orig. & amend. & orig. & amend. & amend. \\
Regime & no-th. & no-th. & no-th. & think & think & no-th. & no-th. & no-th. & think & think & no-th. & no-th. & no-th. \\
Arm / scoring & prim. & prim. & DA & prim. & prim. & prim. & prim. & DA & prim. & prim. & prim. & prim. & DA \\
\midrule
Raw transcript & 0.27 & 0.38 & 0.50 & 0.29 & 0.80 & 0.34 & 0.50 & 0.66 & 0.28 & 0.63 & 0.12 & 0.32 & 0.34 \\
\quad orig.\ brief, revised sched. & -- & 0.30 & -- & -- & -- & -- & 0.40 & -- & -- & -- & -- & -- & -- \\
Text checklist & 0.56 & 0.55 & 0.79 & 0.63 & 0.80 & 0.54 & 0.59 & 0.77 & 0.70 & 0.88 & 0.35 & 0.36 & 0.45 \\
Directive gate & 0.84 & 0.84 & 0.90 & 0.88 & 0.88 & 0.81 & 0.87 & 0.92 & 0.94 & 0.95 & 0.70 & 0.75 & 0.77 \\
Enforcement gate & 0.99 & 0.98 & 0.98 & 0.94 & 0.95 & 0.98 & 0.98 & 0.98 & 0.96 & 0.93 & 0.99 & 0.97 & 0.97 \\
\quad generator's graph, no compile & -- & 0.98 & 0.98 & -- & -- & -- & 1.00 & 1.00 & -- & -- & -- & -- & -- \\
\midrule
\multicolumn{14}{l}{\emph{Raw transcript, errors per $128$ episodes}} \\
\quad re-executions & 121 & 70 & 64 & 140 & 17 & 97 & 64 & 45 & 136 & 29 & 133 & 120 & 107 \\
\quad superseded exec. & 43 & 37 & 0 & 2 & 0 & 43 & 30 & 0 & 3 & 0 & 29 & 29 & 18 \\
\quad premature exec. & 31 & 31 & 31 & 13 & 5 & 25 & 22 & 22 & 23 & 19 & 23 & 27 & 27 \\
\quad omissions & 26 & 35 & 35 & 13 & 14 & 27 & 24 & 24 & 16 & 32 & 109 & 79 & 79 \\
\multicolumn{14}{l}{\emph{Text checklist, errors per $128$ episodes}} \\
\quad re-executions & 33 & 25 & 7 & 17 & 13 & 44 & 19 & 9 & 37 & 5 & 144 & 181 & 127 \\
\quad superseded exec. & 37 & 36 & 1 & 2 & 2 & 32 & 27 & 1 & 8 & 0 & 43 & 73 & 34 \\
\midrule
\multicolumn{14}{l}{\emph{Adjacent rungs, amended brief, primary scoring: Holm-adjusted exact McNemar $p$}} \\
& \multicolumn{3}{c}{no-think} & \multicolumn{2}{c}{think} & \multicolumn{3}{c}{no-think} & \multicolumn{2}{c}{think} & \multicolumn{3}{c}{no-think} \\
\quad raw$\to$checklist & \multicolumn{3}{c}{0.032} & \multicolumn{2}{c}{1.000 (n.s.)} & \multicolumn{3}{c}{0.620 (n.s.)} & \multicolumn{2}{c}{$9.1{\times}10^{-5}$} & \multicolumn{3}{c}{0.568 (n.s.)} \\
\quad checklist$\to$directive & \multicolumn{3}{c}{$4.4{\times}10^{-6}$} & \multicolumn{2}{c}{0.620 (n.s.)} & \multicolumn{3}{c}{$9.5{\times}10^{-7}$} & \multicolumn{2}{c}{0.577 (n.s.)} & \multicolumn{3}{c}{$1.5{\times}10^{-8}$} \\
\quad directive$\to$enforcement & \multicolumn{3}{c}{$5.5{\times}10^{-4}$} & \multicolumn{2}{c}{0.185 (n.s.)} & \multicolumn{3}{c}{$5.5{\times}10^{-4}$} & \multicolumn{2}{c}{1.000 (n.s.)} & \multicolumn{3}{c}{$1.2{\times}10^{-7}$} \\
\bottomrule
\end{tabular}}
\end{table}

\subsection{Graph size under the amended brief}
\label{app:a1}
Table~\ref{tab:a1} fills the amended-brief grid at Qwen3-235B across $S{=}5$, $10$, $15$
sparse, $15$ dense, and $18$ ($128$ episodes per cell). Without thinking the ordering is
non-decreasing in every cell; the raw transcript stays between $0.34$ and $0.50$ and the
enforcement gate between $0.91$ and $0.98$. Re-execution leads the raw transcript's
errors at every size ($75$, $70$, $82$, $64$, and $53$ per $128$ episodes at $S{=}5$,
$10$, $15$ sparse, $15$ dense, and $18$). The two gate contrasts separate after Holm
correction in the two dense-probe cells of Table~\ref{tab:ladder}; in the three sparser
cells the enforcement gate leads the directive gate by $0.07$--$0.10$ without surviving
the correction, and its remaining errors there are unrequested bookings of eligible
steps that state-bound enforcement admits by design ($12$ at $S{=}5$) and omissions.

With thinking the ladder compresses from below. The raw transcript rises to
$0.63$--$0.82$ across the five cells while the enforcement gate does not rise with it,
holding $0.88$--$0.95$ against $0.91$--$0.98$ without thinking, so the span the ladder
has to measure shrinks from $0.48$--$0.60$ to $0.06$--$0.30$. Three of the fifteen
adjacent contrasts survive Holm correction over the five thinking cells
(raw$\to$checklist at $S{=}15$ sparse and $S{=}15$ dense, checklist$\to$directive at
$S{=}5$); the two gates separate in no cell, and at $S{=}10$ and $S{=}18$ no adjacent
contrast separates at all. The compression is not
only the raw rung improving: with thinking the enforcement gate's own errors are almost
entirely omissions ($8$, $19$ and $9$ at $S{=}15$ sparse, $S{=}18$ and $S{=}5$), the
declines of an \textsc{eligible} directive that Appendix~\ref{app:residuals:tables}
traces to invented state, and the text checklist becomes erratic rather than
monotone, falling below the raw transcript at $S{=}5$ ($0.68$ vs.\ $0.77$) on $23$
refused legitimate redos. The ladder's separation is therefore a property of the
no-think regime at this model, and of Qwen3.6-35B without thinking (its thinking row
is partial, Table~\ref{tab:inventory}).

Per-turn accuracy, the fraction of scored turns handled correctly, compresses the same
ladder into a narrow band because most turns are fillers or unambiguous asks; it is
reported for completeness and never used for a claim.
\begin{table}[H]
\centering\small
\setlength{\tabcolsep}{3pt}
\caption{Graph-size sweep under the amended brief at Qwen3-235B. Top: strict episode success (Wilson $95\%$ CI), $128$ episodes per cell (seeds $100$--$227$), for five graph sizes ($S{=}15$ sparse has prerequisite density $\rho{=}0.15$, dense $\rho{=}0.3$), without and with per-turn thinking; best rung per row and regime in bold. Bottom: per-turn accuracy, the fraction of scripted turns served correctly (no-think; the scored turn count per cell is in the row label).}
\label{tab:a1}
\resizebox{0.95\textwidth}{!}{%
\begin{tabular}{l cccc cccc}
\toprule
& \multicolumn{4}{c}{no-think} & \multicolumn{4}{c}{thinking} \\
\cmidrule(lr){2-5}\cmidrule(lr){6-9}
Graph & Raw transcript & Text checklist & Directive gate & Enforcement gate & Raw transcript & Text checklist & Directive gate & Enforcement gate \\
\midrule
\multicolumn{9}{l}{\emph{Strict episode success}} \\
$S{=}5$           & 0.34 {\scriptsize[0.27,0.43]} & 0.45 {\scriptsize[0.36,0.53]} & 0.84 {\scriptsize[0.77,0.90]} & \textbf{0.91} {\scriptsize[0.84,0.95]} & 0.77 {\scriptsize[0.69,0.83]} & 0.68 {\scriptsize[0.59,0.75]} & \textbf{0.93} {\scriptsize[0.87,0.96]} & \textbf{0.93} {\scriptsize[0.87,0.96]} \\
$S{=}10$          & 0.38 {\scriptsize[0.30,0.47]} & 0.55 {\scriptsize[0.47,0.64]} & 0.84 {\scriptsize[0.76,0.89]} & \textbf{0.98} {\scriptsize[0.94,1.00]} & 0.80 {\scriptsize[0.72,0.86]} & 0.80 {\scriptsize[0.73,0.86]} & 0.88 {\scriptsize[0.81,0.92]} & \textbf{0.95} {\scriptsize[0.90,0.98]} \\
$S{=}15$ sparse   & 0.35 {\scriptsize[0.27,0.44]} & 0.61 {\scriptsize[0.52,0.69]} & 0.88 {\scriptsize[0.81,0.92]} & \textbf{0.95} {\scriptsize[0.90,0.98]} & 0.73 {\scriptsize[0.65,0.80]} & 0.91 {\scriptsize[0.85,0.95]} & \textbf{0.95} {\scriptsize[0.90,0.98]} & 0.94 {\scriptsize[0.88,0.97]} \\
$S{=}15$ dense    & 0.50 {\scriptsize[0.41,0.58]} & 0.59 {\scriptsize[0.51,0.68]} & 0.87 {\scriptsize[0.80,0.92]} & \textbf{0.98} {\scriptsize[0.94,1.00]} & 0.63 {\scriptsize[0.55,0.71]} & 0.88 {\scriptsize[0.82,0.93]} & \textbf{0.95} {\scriptsize[0.89,0.97]} & 0.93 {\scriptsize[0.87,0.96]} \\
$S{=}18$          & 0.42 {\scriptsize[0.34,0.51]} & 0.53 {\scriptsize[0.45,0.62]} & 0.83 {\scriptsize[0.75,0.88]} & \textbf{0.93} {\scriptsize[0.87,0.96]} & 0.82 {\scriptsize[0.74,0.88]} & 0.88 {\scriptsize[0.82,0.93]} & \textbf{0.89} {\scriptsize[0.82,0.93]} & 0.88 {\scriptsize[0.81,0.92]} \\
\midrule
\multicolumn{9}{l}{\emph{Per-turn accuracy, no-think (correct scripted turns over scored turns)}} \\
$S{=}5$ (1405 turns)& 0.887 & 0.914 & 0.967 & \textbf{0.989} &  &  &  & \\
$S{=}10$ (2048 turns)& 0.910 & 0.956 & 0.982 & \textbf{0.999} &  &  &  & \\
$S{=}15$ sparse (2688 turns)& 0.929 & 0.969 & 0.993 & \textbf{0.996} &  &  &  & \\
$S{=}15$ dense (2688 turns)& 0.943 & 0.970 & 0.987 & \textbf{0.999} &  &  &  & \\
$S{=}18$ (3072 turns)& 0.946 & 0.966 & 0.989 & \textbf{0.996} &  &  &  & \\
\bottomrule
\end{tabular}}
\end{table}

\subsection{Checklist rendering and directive placement}
\label{app:a2}
The checklist arm appends the rendered state to every user message and leaves every
copy in the history, and both gates deliver their directive in the user role. Three
controls at Qwen3-235B ($S{=}10$ and $S{=}15$ dense) and Qwen3.6-35B ($S{=}15$ dense),
all no-think under the amended brief, test whether either choice handicaps a rung
(Table~\ref{tab:a2}). \emph{Replace mode} strips the checklist from every earlier user
message so only the current turn carries it. \emph{Checklist plus matched step} keeps
replace mode and adds one line naming the step the matcher resolved, with no status
instruction, which separates the matcher from the directive. \emph{System role} adds
the directive to the system prompt for that call only, never to the history; rejection
notices stay in the user role. Replace mode never helps and hurts at Qwen3.6-35B. The
matched-step line lowers success at Qwen3-235B and raises it at Qwen3.6-35B, so the
line alone is not a reliable coupling, and the directive gate's gain over the checklist
comes from the instruction. System-role directives hurt both gates at both models,
most at Qwen3.6-35B, where the enforcement gate falls to $0.26$ while refusing $70$
bookings: an instruction the model does not act on leaves the gate to refuse, and the
episode then fails by omission. The self-ledger beats replace mode as it beats the
appended checklist under the primary scoring; under decline-aware scoring its margin
over replace mode is not significant, so the main text states the claim under the
primary scoring only.
\begin{table}[H]
\centering\small
\setlength{\tabcolsep}{3pt}
\caption{Prompt-surface variants of the text and gate rungs under the amended brief (no-think, $128$ episodes per cell, seeds $100$--$227$). Replace mode shows the checklist only in the current turn instead of appending it to every turn of the transcript. The matched-step line adds to the replace-mode checklist one line naming the step the matcher resolved the request to, with no status instruction. System role moves the gate's per-turn directive from the user message into the system message. Strict episode success (Wilson $95\%$ CI); best row per column in bold; $p$ is exact McNemar of each variant against its paper arm on shared seeds (the self-ledger against the replace-mode checklist), Holm-corrected within model over all variant contrasts (ten at Qwen3-235B, four at Qwen3.6-35B). Last column: mean prompt tokens per episode at Qwen3-235B, $S{=}15$ dense.}
\label{tab:a2}
\resizebox{0.95\textwidth}{!}{%
\begin{tabular}{l cc cc cc c}
\toprule
& \multicolumn{2}{c}{Qwen3-235B, $S{=}10$} & \multicolumn{2}{c}{Qwen3-235B, $S{=}15$ dense} & \multicolumn{2}{c}{Qwen3.6-35B, $S{=}15$ dense} & \multicolumn{1}{c}{Qwen3-235B, $S{=}15$ dense} \\
\cmidrule(lr){2-3}\cmidrule(lr){4-5}\cmidrule(lr){6-7}\cmidrule(lr){8-8}
Arm & success & $p$ & success & $p$ & success & $p$ & prompt tokens per episode \\
\midrule
Text checklist (append, paper)      & 0.55 {\scriptsize[0.47,0.64]} & -- & 0.59 {\scriptsize[0.51,0.68]} & -- & 0.36 {\scriptsize[0.28,0.45]} & -- & 215k \\
Checklist, replace mode             & 0.52 {\scriptsize[0.44,0.61]} & 0.704 (n.s.) & 0.46 {\scriptsize[0.38,0.55]} & 0.119 (n.s.) & 0.27 {\scriptsize[0.20,0.35]} & 0.126 (n.s.) & 127k \\
Checklist + matched step            & 0.34 {\scriptsize[0.26,0.42]} & 0.0055 & 0.49 {\scriptsize[0.41,0.58]} & 0.222 (n.s.) & 0.51 {\scriptsize[0.42,0.59]} & 0.045 & 112k \\
Self-ledger                         & 0.70 {\scriptsize[0.62,0.78]} & 0.0088 & 0.79 {\scriptsize[0.71,0.85]} & $1.0{\times}10^{-6}$ & -- & -- & 326k \\
\addlinespace[3pt]
Directive gate (user role, paper)   & 0.84 {\scriptsize[0.76,0.89]} & -- & 0.87 {\scriptsize[0.80,0.92]} & -- & 0.75 {\scriptsize[0.67,0.82]} & -- & 108k \\
Directive gate, system role         & 0.39 {\scriptsize[0.31,0.48]} & $1.3{\times}10^{-11}$ & 0.48 {\scriptsize[0.39,0.56]} & $8.0{\times}10^{-11}$ & 0.17 {\scriptsize[0.12,0.25]} & $4.2{\times}10^{-16}$ & 111k \\
\addlinespace[3pt]
Enforcement gate (user role, paper) & \textbf{0.98} {\scriptsize[0.94,1.00]} & -- & \textbf{0.98} {\scriptsize[0.94,1.00]} & -- & \textbf{0.97} {\scriptsize[0.92,0.99]} & -- & 108k \\
Enforcement gate, system role       & 0.66 {\scriptsize[0.58,0.74]} & $8.0{\times}10^{-11}$ & 0.69 {\scriptsize[0.60,0.76]} & $6.5{\times}10^{-11}$ & 0.26 {\scriptsize[0.19,0.34]} & $7.6{\times}10^{-26}$ & 113k \\
\bottomrule
\end{tabular}}
\end{table}

\subsection{A second domain: release engineering}
\label{app:domain2}
To separate the ladder from procurement vocabulary, a second generator domain ships a
software release: $18$ step templates (freeze the scope, cut the release branch, run
regression, sign the artifacts, canary rollout, \dots) with three paraphrases each and
step slugs disjoint from procurement's (\texttt{killtest/gen\_dp.py},
\texttt{domain=release}). Its per-phase template counts match procurement's exactly, so
the phase-consistent dependency sampler induces the same DAG-shape distribution, and the
trap budget, schedule, and scoring are unchanged. The difficulty knobs are pinned and
only the surface vocabulary moves. At Qwen3-235B no-think, $S{=}15$ dense, amended
brief ($128$ episodes, seeds $100$--$227$), the ladder is $0.48 \to 0.44 \to 0.78 \to
0.93$ (Table~\ref{tab:domain2}): both gate contrasts replicate (Holm
$p{\le}6.2{\times}10^{-4}$) while the text checklist does not separate from the raw
transcript ($p{=}0.60$; it also fails to separate at Qwen3.6-35B above). The
enforcement gate refused $61$ bookings; its remaining errors are $11$ omissions, one refused
redo, and one premature execution.
\begin{table}[H]
\centering\footnotesize
\setlength{\tabcolsep}{4pt}
\caption{Second generator domain (release engineering) at Qwen3-235B no-think under the amended brief, $S{=}15$ dense, $128$ episodes per cell (seeds $100$--$227$). The release pool holds $18$ step templates with three paraphrases each, slugs disjoint from procurement's, and a per-phase template histogram identical to procurement's, so the dependency sampler induces the same DAG-shape distribution; trap budget, schedule, and scoring are unchanged (\texttt{killtest/gen\_dp.py}). Strict and decline-aware (DA) episode success and, below, exact McNemar for adjacent rungs on shared seeds, Holm-corrected over the three contrasts. The enforcement gate refused $61$ bookings; its remaining errors are $11$ omissions, one refused redo, and one premature execution.}
\label{tab:domain2}
\begin{tabular}{l cccc}
\toprule
& Raw transcript & Text checklist & Directive gate & Enforcement gate \\
\midrule
Strict & 0.48 & 0.44 & 0.78 & 0.93 \\
Decline-aware & 0.63 & 0.61 & 0.82 & 0.93 \\
\midrule
\multicolumn{5}{l}{\emph{Adjacent rungs, primary scoring: Holm-adjusted exact McNemar $p$}} \\
\quad raw$\to$checklist & \multicolumn{4}{c}{0.60 (n.s.)} \\
\quad checklist$\to$directive & \multicolumn{4}{c}{$1.2{\times}10^{-9}$} \\
\quad directive$\to$enforcement & \multicolumn{4}{c}{$6.2{\times}10^{-4}$} \\
\bottomrule
\end{tabular}
\end{table}

\section{Retrieval-Based Memory Baselines and the Self-Ledger}
\label{app:memory}
Table~\ref{tab:memory} runs the two retrieval paradigms and the self-ledger arm of
Section~\ref{sec:systems:ladder} on the amended-brief Qwen3-235B no-think cells of
Table~\ref{tab:e1} (seeds $100$--$227$, identical episodes and scoring;
\texttt{killtest/dp\_memory.py}). The top-$k$ arm embeds each past exchange with a CPU
sentence-transformer \citep{reimers2019sentencebert} (\texttt{all-MiniLM-L6-v2};
\citealp{wang2020minilm}) and recalls the five most similar into
the current turn; the Mem0 arm (OSS $2.0.11$) runs its own extraction and update
LLM (the cell's agent model at temperature $0$) with per-turn add and search; both arms
run in both modes. In replace mode the agent keeps the brief, the last two exchanges, and
the recall block; in augment mode the full transcript stays.

Used as the record, retrieval fails the lifecycle almost completely: $0/128$ and
$2/128$ for top-$k$ retrieval, $0/128$ twice for Mem0 (Wilson $95\%$ upper bound
$0.029$ at $0/128$), against $0.38$ and $0.50$ for the raw transcript on the same
episodes. The failures are omission-heavy ($157$ and $203$ omissions per $128$
episodes for top-$k$ retrieval, $299$ and $274$ for Mem0, against the raw transcript's
$35$ and $24$), with re-execution the other large channel: a recall block of similar
past exchanges neither surfaces every obligation the current request owes nor marks
which recalled work is finished. The recall itself is not broken: the top-$k$ index
returns hits on every queried turn after the first record is stored, and Mem0's store
is populated in all $256$ episodes, though its search returns an empty hit list on
$231$ and $209$ of its $2{,}087$ and $2{,}747$ queried turns.

In augment mode the same recall block on top of the full transcript helps: $0.67$ and
$0.71$, above the text checklist ($0.55$ and $0.59$) and cutting raw-transcript
re-executions from $70$ to $34$ and from $64$ to $37$, because the recalled exchange
often re-surfaces the original booking next to the fresh request. It stays below the
directive gate ($0.84$ and $0.87$): on shared seeds the augment arm fails $37$ episodes
the gate passes at $S{=}10$ and passes $16$ the gate fails (exact McNemar $p{=}0.005$;
$32$ vs.\ $12$, $p{=}0.004$ at $S{=}15$ dense). Retrieved content, like the checklist,
is text the agent may ignore; the ladder's ordering is unchanged.

Mem0 in augment mode does not help: $0.43$ and $0.45$ ($55$ and $57$ of $128$), at the
raw transcript's $0.38$ and $0.50$ (discordant $(26,20)$ and $(27,34)$; Holm $p{=}0.89$
for both) and below top-$k$ augment on shared seeds ($46$ vs.\ $15$ and $44$ vs.\ $10$
episodes; exact McNemar $p{=}8.8{\times}10^{-5}$ and $3.4{\times}10^{-6}$). Its omissions
return to the transcript's level ($31$ and $38$), so the recall block no longer displaces
the record, but its re-executions do not move ($67$ and $71$ against the transcript's
$70$ and $64$; top-$k$ augment $34$ and $37$). What differs is the form of the recall.
Top-$k$ returns past exchanges verbatim, the agent's own receipt line included, and at
none of the $33$ and $34$ turns on which it re-executed did the block contain the redone
step's receipt: with the original booking in view in the agent's own words, the agent
does not book again. Mem0 returns the extractor's prose (``Task s1 \ldots\ has been
completed with reference RC-9142/\#W8733''; about four items per queried turn, empty on
$259$ of $2{,}068$ and $215$ of $2{,}704$ queried turns), and at $28$ and $57$ of the $66$
and $71$ turns on which it re-executed, that prose named the redone step with its receipt
number. A restated completion does not carry the weight of the agent's own receipt, the
asymmetry the trap-turn probe finds when the agent's execution replies are rewritten as
prose (Appendix~\ref{app:validation:probe}).

The self-ledger arm (Section~\ref{sec:systems:ladder}) shares the table. The agent
followed the protocol in all $256$ episodes (median $17$ and $22$ \texttt{LEDGER} marks
per episode; no episode wrote none), and its unverified, self-written record beats the
always-accurate checklist ($0.70$ and $0.79$ vs.\ $0.55$ and $0.59$; Holm $p{=}0.011$
and $0.0016$ over the four self-ledger contrasts) while sitting below the directive
gate in point estimate in both cells, significantly at $S{=}10$ only (disagreeing pairs
$(29,12)$, Holm $p{=}0.023$ over the two gate contrasts; $(24,14)$, $p{=}0.14$ at
$S{=}15$ dense). Its remaining errors are mostly redos ($27$ and $21$ re-executions per $128$
episodes), and re-reading its own record does not stop it: $24$ of $26$ and $20$ of
$20$ on-turn re-executions fire with the step already marked \textsc{done} in the
agent's own rendered ledger. Ownership moves the agent up the text family;
verification and coupling move it out. Its reversal on PM-Bench, where the shown record
beats the self-written ledger (Appendix~\ref{app:pmbench}), fits the same reading: on a
discrete graph with receipts, writing the status line is a commitment the next turn
re-reads, while over an $80$-step week with hidden channels the agent's own ledger
misses updates that a system-kept record carries.
\begin{table}[H]
\centering\footnotesize
\setlength{\tabcolsep}{4pt}
\caption{Retrieval-based memory baselines and the self-ledger arm at Qwen3-235B no-think under the amended brief, $128$ episodes per cell (seeds $100$--$227$), against the four ladder arms of the same configuration (Table~\ref{tab:e1}). \emph{Replace} is deployed memory-system usage: the agent sees the brief, the last two exchanges, and the current turn with a recall block; \emph{augment} appends the same recall block to the full raw transcript; the \emph{self-ledger} arm renders back, in the checklist's slot, a status ledger the agent writes itself and no one verifies (Section~\ref{sec:systems:ladder}; \texttt{killtest/dp\_memory.py}, \texttt{killtest/run\_dp.py}). Strict and decline-aware (DA) episode success, omissions per $128$ episodes (primary scoring), and exact McNemar vs the raw transcript on shared seeds, Holm-corrected over the eight memory-vs-raw contrasts (primary scoring; under decline-aware scoring the replace arms stay below $10^{-18}$, top-$k$ augment is $2{\times}10^{-5}$ and $0.12$, and Mem0 augment is not separated from the raw transcript); the self-ledger $p$ vs raw is Holm-corrected over its own four contrasts (vs raw and vs checklist, both cells). The replace arms sit below the raw transcript, Mem0 augment at it, and top-$k$ augment above the text checklist and below the directive gate (exact McNemar $p{=}0.005$ and $p{=}0.004$); the self-ledger arm sits above the checklist (Holm $p{=}0.011$ and $0.0016$) and below the directive gate in point estimate in both cells, significantly at $S{=}10$ only (discordant $(29,12)$, Holm $p{=}0.023$ over the two gate contrasts; $(24,14)$, $p{=}0.14$ at $S{=}15$ dense).}
\label{tab:memory}
\begin{tabular}{l cccc cccc}
\toprule
& \multicolumn{4}{c}{$S{=}10$} & \multicolumn{4}{c}{$S{=}15$ dense} \\
\cmidrule(lr){2-5}\cmidrule(lr){6-9}
Arm & strict & DA & omis. & $p$ vs.\ raw & strict & DA & omis. & $p$ vs.\ raw \\
\midrule
Raw transcript & 0.38 & 0.50 & 35 & -- & 0.50 & 0.66 & 24 & -- \\
\midrule
Top-$k$ retrieval, replace & 0.00 & 0.00 & 157 & $2.1{\times}10^{-14}$ & 0.02 & 0.02 & 203 & $4.2{\times}10^{-16}$ \\
Mem0, replace & 0.00 & 0.00 & 299 & $2.1{\times}10^{-14}$ & 0.00 & 0.00 & 274 & $8.7{\times}10^{-19}$ \\
Top-$k$ retrieval, augment & 0.67 & 0.73 & 4 & $4.8{\times}10^{-7}$ & 0.71 & 0.77 & 9 & $4.3{\times}10^{-4}$ \\
Mem0, augment & 0.43 & 0.48 & 31 & 0.885 & 0.45 & 0.54 & 38 & 0.885 \\
\midrule
Self-ledger (agent-written) & 0.70 & 0.81 & 4 & $3.8{\times}10^{-7}$ & 0.79 & 0.80 & 12 & $1.3{\times}10^{-6}$ \\
\midrule
Text checklist & 0.55 & 0.79 & 0 & -- & 0.59 & 0.77 & 1 & -- \\
Directive gate & 0.84 & 0.90 & 2 & -- & 0.87 & 0.92 & 3 & -- \\
Enforcement gate & 0.98 & 0.98 & 0 & -- & 0.98 & 0.98 & 2 & -- \\
\bottomrule
\end{tabular}
\end{table}

\section{Real-Tool Harness Details}
\label{app:rw}

\begin{table}[t]
\centering\small
\caption{\textbf{Tool-dispatch harness} ($S{=}10$, $50$ episodes per arm). Strict success (Wilson $95\%$ CI), adjacent-rung McNemar $p$, refusals, and error totals (turn counts summed over the $50$ episodes).}
\label{tab:rw}
\setlength{\tabcolsep}{3pt}
\resizebox{0.84\textwidth}{!}{%
\begin{tabular}{l c l c r l}
\toprule
Arm & Strict & 95\% CI & $p$ vs.\ prev & Refused & Errors (prem.\,/\,unreq.\,/\,redo\,/\,omit.) \\
\midrule
\multicolumn{6}{l}{\emph{Qwen3.6-35B}} \\
Raw transcript   & 0.52 & [0.39,0.65] & --   & --  & 49 / 13 / 7 / 5 \\
Text checklist & 0.58 & [0.44,0.71] & 0.61 & --  & 51 / 14 / 14 / 0 (+2 superseded) \\
Directive gate   & 0.88 & [0.76,0.94] & $3{\times}10^{-4}$ & -- & 18 / 14 / 53 / 3 \\
Enforcement gate & 0.88 & [0.76,0.94] & 1.0  & 63  & 15 / 21 / 0 / 3 \\
Request-bound enforcement & \textbf{0.96} & [0.87,0.99] & 0.22 & 34 & 0 / 0 / 0 / 2 \\
\midrule
\multicolumn{6}{l}{\emph{Qwen3-235B}} \\
Raw transcript   & 0.34 & [0.22,0.48] & --   & --  & 32 / 7 / 39 / 7 \\
Text checklist & 0.62 & [0.48,0.74] & 0.0066 & -- & 20 / 12 / 2 / 1 (+3 refused redos) \\
Directive gate   & 0.96 & [0.87,0.99] & $7.6{\times}10^{-5}$ & -- & 2 / 0 / 0 / 1 \\
Enforcement gate & \textbf{0.98} & [0.90,1.0] & 1.0 & 0 & 0 / 0 / 0 / 1 \\
Request-bound enforcement & \textbf{0.98} & [0.90,1.0] & 1.0 & 0 & 0 / 0 / 0 / 1 \\
\bottomrule
\end{tabular}}
\end{table}

The harness (\texttt{killtest/rw\_domain.py}, \texttt{killtest/run\_rw.py}) maps each of
the $18$ procurement step templates to a tool with a filesystem side effect in a
per-episode workspace: \texttt{collect\_requirements}, \texttt{draft\_rfq},
\texttt{send\_rfq}, \texttt{register\_budget\_code}, \texttt{tabulate\_quotes},
\texttt{run\_security\_review}, \texttt{select\_vendor}, \texttt{record\_negotiation},
\texttt{order\_backup\_units}, \texttt{file\_signed\_msa}, \texttt{place\_main\_order},
\texttt{book\_delivery\_window}, \texttt{schedule\_installers},
\texttt{obtain\_insurance\_cert}, \texttt{arrange\_training},
\texttt{run\_acceptance\_test}, \texttt{close\_project}, and \texttt{book\_waste\_pickup}.
Every execution appends a block stamped \texttt{== EXECUTION \#W1234 ==} to the tool's
artifact, so the artifact is the receipt: scoring inspects the workspace after each turn
and a second stamped block is an observed re-execution. No completion codes exist in this
setting. The harness brief differs from the payload brief of Appendix~\ref{app:prompts}
in four ways (\texttt{killtest/rw\_domain.py}): the agent operates ``a shared
workspace''; each step line carries \texttt{[tool: <name>]} instead of a completion
code; the one-shot sentence is always present (``Each step is performed ONCE. If I
mention or ask about a finished step again without explicitly ordering a redo, confirm
that it is done and do not call its tool again.''); and the execution protocol is
replaced by the \textsc{tools} clause of Appendix~\ref{app:prompts}. Every real-tool
number is therefore an amended-brief number.

The agent loop is implemented as a compiled LangGraph \texttt{StateGraph}
\citep{langgraph} (\texttt{killtest/run\_rw.py}): an agent node
emits \texttt{ACTION} lines, a tool node executes them, and a conditional edge loops back
until no action is emitted, at most three hops per turn. Tool results and refusals return
as a user-role message. The dispatcher refuses any call carrying a stale work order in
every arm, and every refusal is logged with its hop and reason. Five arms are run: the raw
transcript; the text checklist computed from the workspace; the directive gate; the
enforcement gate under state-bound enforcement, which refuses a tool call that violates its compiled
state and returns the refusal as the tool result; and request-bound enforcement, which
also refuses any step other than the one the request resolved to. Step resolution
and explicit-redo intent come from the same matcher as the payload testbed, with the same
budgets (Table~\ref{tab:serving}); the module's graph and revision are compiled from the
brief, with compile accuracy, from the per-episode launcher logs
(\texttt{killtest/redteam\_r2/rw\_compile\_acc.py}), of $1.000$ on tool names and
dependencies at Qwen3.6-35B ($50/50$ episodes) but $49/50$ on tool names and $45/50$ on
dependencies at Qwen3-235B: one step's tool name is left empty at seed $307$ and one
step's dependency list is wrong in each of seeds $317$, $321$, $323$, $329$, and $342$
($0.998$ and $0.990$ averaged over episodes; the two gate arms share the compile cache).

Table~\ref{tab:rw} reports $50$ episodes per arm ($S{=}10$, seeds $300$--$349$). The
stale-work-order refusal fires $5$ times at Qwen3.6-35B and $3$ times at Qwen3-235B in
the raw-transcript arm. At Qwen3.6-35B the raw transcript's dominant failure is premature execution
($49$ over $50$ episodes), with a second pattern the payload testbed cannot produce: on
some turns the agent re-issues a batch of tool calls it made earlier under the current
work order, re-executing several finished steps at once (episode $304$ of the smoke run:
$17$ re-executions in one episode). The enforcement gate refused $63$ tool calls ($62$
\textsc{already done}, one \textsc{cancelled}) across $29$ turns, in $22$ of which the
agent retried after the refusal; its remaining errors ($15$ premature, $21$ unrequested) arise
when the agent, asked for a step whose prerequisites are not done, executes the
prerequisites itself within the turn and then the step, each call admitted because the
module's state is correct at every instant. Request-bound enforcement refuses $34$ calls
on $15$ turns, all \textsc{not requested}, with a retry on $14$ of them; the
\textsc{already done} refusals of state-bound enforcement disappear because re-issued
batches are caught as not-requested first. With the Qwen3-235B agent no gate arm
refuses a call; the enforcement gate's single failure is an omission at seed $329$, one
of the five miscompiled episodes, whose spurious dependency
(\texttt{book\_delivery\_window} on the unfinished \texttt{close\_project}) makes
the directive mark the requested step blocked, so the agent defers (an error of the
state, not of enforcement). Request binding changes nothing for this agent: request-bound
enforcement scores the same $49/50$ with the same omission, zero refusals, and no episode
scored differently from state-bound enforcement, because the obedient agent executes no
unrequested prerequisite for the binding to catch. Every episode's first record is stamped
with the harness name and version (\texttt{langgraph} $1.2.11$).

\section{External-Benchmark Case Study: \texorpdfstring{$\tau^2$}{tau2}-bench Airline}
\label{app:tau2}
The enforcement results above run in environments we built, so we also inject the gate
into a benchmark we did not build: the airline domain of $\tau^2$-bench
\citep{barres2026tau2bench}, the dual-control successor of $\tau$-bench
\citep{yao2024taubench}, MIT licensed. The domain's policy document warns twice that the
booking API does not check its own cancellation rules; the benchmark ships the
enforcement gap this paper studies. The injection is a one-method override of the
environment's tool dispatch (\texttt{killtest/scratch\_tau2/gated\_env.py}): every agent
tool call already routes through \texttt{Environment.get\_response}, and the gated
subclass checks each WRITE-typed call against a compiled ruleset and the current
database before executing it, returning a veto as an ordinary tool-error message. READ
tools, the stock agent loop, the user simulator, and the official DB and COMMUNICATE
scoring are untouched. The ruleset ($16$ deny rules over the six WRITE tools) is
compiled in one call by the agent model from the policy document, as in
Appendix~\ref{app:compile}; it scores $32/33$ on a hand-written unit battery whose
expected verdicts encode a hand-audited reference ruleset (the one miss: the compiled
set does not encode that flight changes cannot be paid with a travel certificate), and
it is behaviorally identical to that reference when replayed over the benchmark's
shipped transcripts. The rule vocabulary went through one round of refinement, an audit
against the benchmark's gold trajectories that touched two rules: it removed a
flown-segment refusal on cancellation which the benchmark's own reference solutions
contradict, and taught the trip-endpoint rule that a swap between same-metro airports
(LGA for JFK) is not an endpoint change, which those solutions perform; the audit's
full effect is isolated at the end of this section. Two paired
runs use the same ruleset. The first runs the $50$ airline tasks for $2$ trials with
paired seeds ($100$ episodes per arm) with Qwen3-235B without thinking as both the agent
and the user simulator; the second runs $4$ trials ($200$ episodes per arm) with
Qwen3.6-35B without thinking as the agent and Qwen3-235B as the user simulator. Because
the user simulator is cluster-served rather than the leaderboard's, all numbers are
internally paired and not comparable to published $\tau^2$-bench scores.

Table~\ref{tab:tau2} reports both pairs. A wrongful-write episode executes at least one
WRITE call outside the task's gold write multiset, comparing tool name and normalized
arguments. At Qwen3-235B the ungated agent is wrongful in $54$ of $100$ episodes ($35$
add extra writes only, $9$ call the right tools with wrong arguments, $10$ do both) and
passes $0.39$. The gate refuses $33$ calls in $24$ episodes ($14$ ineligible
cancellations, $8$ compensation-amount mismatches, $6$ basic-economy modifications, $4$
trip-endpoint changes, $2$ certificate limits, $2$ flown-segment modifications, $1$
baggage decrease; four calls matched two rules), and the paired deltas are $+0.15$ on
pass$^1$ (task-level bootstrap $95\%$ CI $[+0.04,+0.26]$) and $-0.19$ on the wrongful
rate (CI $[-0.30,-0.08]$); pass$^2$ moves from $0.26$ to $0.44$. Replaying the $100$
gated transcripts through the ruleset finds zero executed state-violating writes, and
replaying the task set's $49$ gold-trajectory write actions in order overblocks none.
One live refusal matched a gold call issued out of gold order: in task $32$ the
reference first upgrades a basic-economy reservation and then changes its flights,
while the agent asked for the flight change with the cabin still basic economy, and the
gate refused it as the policy requires. Of the $10$ task-trial pairs that pass ungated
and fail gated, $6$ saw no refusal at all and are run-to-run variance.

The two middle rungs of the ladder run on the same seeds at Qwen3-235B, on the
same ruleset and at the same dispatch point (Table~\ref{tab:tau2}, top). The
\emph{ledger} arm shows the state: every successful entity read and write returns
the stock result followed by a block the checker computes from the database (the
reservation's cabin, insurance, booking age, and segment statuses; the user's
membership and the reservations that qualify for compensation; and, per WRITE
tool, which compiled rules already fire on that state, which depend on what the
call would request, and which are clear); writes execute unguarded. The
\emph{advisory} arm directs: a violating write is returned unexecuted once, with
the rule and its reason and a sentence saying that the identical call issued
again will execute, and the second identical call executes even though it still
violates the rule. Ledger reaches pass$^1$ $0.52$ (paired $+0.13$, CI
$[+0.01,+0.25]$) and executes $4$ state-violating writes of the ungated arm's
$26$; before one of them the block had said \textsc{not permitted}, before two it
had said the verdict depends on the request (basic-economy flight changes, where
the rule needs the requested flights), and one was a booking limit the block does
not list. Advisory reaches $0.49$ ($+0.10$, CI $[+0.01,+0.19]$): $30$ notices in
$23$ episodes, $9$ identical re-issues, $8$ of them executed while still violating
($6$ ineligible cancellations), typically under pressure from the simulated user
(task $1$: ``the system has flagged that the reservation does not meet the
standard cancellation policy \ldots\ since a verbal assurance was made by a
colleague, I will proceed with the cancellation as a goodwill gesture''). One
advisory episode ended on the benchmark's ten-error cap, eight of the ten being
the tool's own payment-arithmetic errors; the same task-trial fails in every arm.
On pass$^1$ the three rule arms sit inside each other's intervals
($0.49$--$0.54$); they differ in the leak ($4$, $8$, $0$ executed violations) and
in what they ask of the agent. Four differences from the payload testbed bound
the comparison: the block appears only on reads and writes the agent chooses to
make, not on every turn, and shows only what the database decides; the notice
costs one refused call, so it is a stronger coupling than the directive of
Section~\ref{sec:results:rq2}; the arms are not nested, since the block derives
facts (booking age against the domain clock, segment status, compensation
eligibility) that the agents of the other three arms must look up themselves;
and every arm, including the ungated one, holds the policy document in its
system prompt, so the arms differ only in how the compiled rules reach the agent
at tool dispatch. Of the $16$ compiled rules, one (a flown-segment condition on
cancellation that the gold-trajectory audit removed from the vocabulary) can fire
in no arm.

At Qwen3.6-35B the same gate changes nothing. Ungated, $57$ of $200$ episodes ($0.285$)
are wrongful, $38$ of them with wrong arguments inside eligible calls; gated, the rate is
$0.260$ and pass$^1$ moves from $0.730$ to $0.755$ (paired CI $[-0.045,+0.100]$; wrongful
CI $[-0.110,+0.055]$), and the multi-trial rates are flat or slightly worse (pass$^4$
$0.560$ ungated, $0.520$ gated). The gate vetoed $12$ live calls in $10$ episodes, the
gated transcripts contain zero executed state-violating writes, and the gold replay
overblocks none.\footnote{On the older task revision embedded in the
benchmark's shipped transcripts the gate overblocks $1$ of $55$ gold write actions, and
all four of its firings on DB-passing episodes come from the same task: airline task
$7$, whose reference trajectory there cancels a reservation that qualifies under no
cancellation branch, a misalignment documented and corrected in the
$\tau^2$-bench-verified follow-up \citep{cuadron2025saber},
\texttt{github.com/amazon-agi/tau2-bench-verified}. In the task set of our runs the
reference trajectory upgrades that reservation to business class before cancelling, and
the gate admits every gold write.}
\begin{table}[h]
\centering\small
\caption{\textbf{$\tau^2$-bench airline case study.} Top: the four arms at Qwen3-235B, $50$ tasks $\times$ $2$ trials on paired seeds, the same model as agent and user simulator: ungated; ledger (the policy state the rules compute from the database is appended to entity reads and successful writes, writes execute unguarded); advisory (a violating write is returned unexecuted once with the rule, and the identical call issued again executes; overridden = re-issued and executed while still violating); gated (refused). Second block: paired arms at Qwen3.6-35B without thinking, $50$ tasks $\times$ $4$ trials, user simulator Qwen3-235B. All numbers are internally paired and not leaderboard-comparable. pass$^k$ is the fraction of tasks passed on every one of $k$ trials. Wrongful episodes execute $\geq 1$ WRITE call outside the task's gold write multiset (wrong-arguments\,/\,extra-only\,/\,both). State-viol.\ = executed writes that violate a compiled rule, counted by replaying each arm's transcripts ($100$ per arm at Qwen3-235B, $200$ at Qwen3.6-35B) through the ruleset: the ungated count is the catchability ceiling, the gated count the leak check. Four of the $33$ vetoes at Qwen3-235B and one of the $12$ at Qwen3.6-35B matched two rules. Paired deltas (gated minus ungated) carry task-level bootstrap $95\%$ CIs. Middle: the same ruleset replayed offline over the benchmark's shipped \texttt{gpt-4.1-mini} transcripts ($200$ sims), next to this study's audits. Bottom: the refusal subset (tasks whose gold trajectory has zero writes, so the correct outcome is to refuse and change nothing), the catchable-episode rate per model, and the Qwen3.5-9B ungated arm; its gated arm was not run under a rule fixed before the run (gate only if $\geq 15\%$ of ungated episodes are catchable).}
\label{tab:tau2}
\setlength{\tabcolsep}{3pt}
\resizebox{\textwidth}{!}{%
\begin{tabular}{l c c c c c}
\toprule
Arm & pass$^1$ & Wrongful ep. & (args\,/\,extra\,/\,both) & Live vetoes & State-viol.\ writes \\
\midrule
\multicolumn{6}{l}{\emph{Qwen3-235B, $50 \times 2$ (agent $=$ user simulator)}} \\
\addlinespace[1pt]
Ungated & 0.390 & 0.540 (54/100) & 9\,/\,35\,/\,10 & -- & 26 \\
Ledger (state shown) & 0.520 & 0.380 (38/100) & 14\,/\,14\,/\,10 & -- (476 blocks) & 4 \\
Advisory (notice, retry executes) & 0.490 & 0.440 (44/100) & 11\,/\,23\,/\,10 & 30 notices in 23 ep., 8 overridden & 8 \\
Gated & 0.540 & 0.350 (35/100) & 14\,/\,14\,/\,7 & 33 in 24 ep. & \textbf{0} \\
\addlinespace[2pt]
Paired $\Delta$, ledger & $+0.130$ $[+0.010,+0.250]$ & $-0.160$ $[-0.280,-0.040]$ & & & \\
Paired $\Delta$, advisory & $+0.100$ $[+0.010,+0.190]$ & $-0.100$ $[-0.200,0.000]$ & & & \\
Paired $\Delta$, gated & $+0.150$ $[+0.040,+0.260]$ & $-0.190$ $[-0.300,-0.080]$ & & & \\
pass$^k$, $k{=}1,2$ & \multicolumn{5}{l}{ungated 0.390, 0.260; ledger 0.520, 0.400; advisory 0.490, 0.340; gated 0.540, 0.440; gold overblock 0/49 in all four replays} \\
\midrule
\multicolumn{6}{l}{\emph{Qwen3.6-35B, $50 \times 4$ (user simulator Qwen3-235B)}} \\
\addlinespace[1pt]
Ungated & 0.730 & 0.285 (57/200) & 38\,/\,14\,/\,5 & -- & 7 \\
Gated & 0.755 & 0.260 (52/200) & 34\,/\,12\,/\,6 & 12 in 10 ep. & \textbf{0} \\
\addlinespace[2pt]
Paired $\Delta$ & $+0.025$ $[-0.045,+0.100]$ & $-0.025$ $[-0.110,+0.055]$ & & & \\
pass$^k$, $k{=}1\ldots4$ & \multicolumn{5}{l}{ungated 0.730, 0.647, 0.595, 0.560; gated 0.755, 0.640, 0.570, 0.520} \\
\midrule
\multicolumn{4}{l}{\emph{Offline replay, shipped \texttt{gpt-4.1-mini} transcripts}} & \multicolumn{2}{l}{\emph{This study's audits}} \\
\addlinespace[1pt]
Episode recall (extra-only) & \multicolumn{3}{l}{0.683 (28/41)} & \multicolumn{2}{l}{Catchable ungated ep.\ 7/200} \\
Episode recall (incl.\ mixed) & \multicolumn{3}{l}{0.714 (40/56)} & \multicolumn{2}{l}{Gated executed violations 0/200 ep.} \\
Episode precision & \multicolumn{3}{l}{0.917 (44/48)} & \multicolumn{2}{l}{Gold overblock (this task set) 0/49} \\
Call precision & \multicolumn{3}{l}{0.923 (48/52)} & \multicolumn{2}{l}{Compile battery 32/33} \\
Gold overblock (shipped tasks) & \multicolumn{3}{l}{1/55 (task 7)} & \multicolumn{2}{l}{} \\
\midrule
\multicolumn{6}{l}{\emph{Refusal subset: gold trajectory has zero writes (24/50 tasks here, 96 ep.\ per arm; 20/50 shipped, 80 ep.)}} \\
\addlinespace[1pt]
Qwen3.6-35B ungated & 0.938 & 0.062 (6/96) & \multicolumn{3}{l}{} \\
Qwen3.6-35B gated & 0.906 & 0.083 (8/96) & \multicolumn{3}{l}{churn, not a gate effect} \\
\texttt{gpt-4.1-mini} shipped & 0.625 & 0.362 (29/80) & \multicolumn{3}{l}{gate catches 0.724 (21/29); 0 fires on 51 clean ep.} \\
\addlinespace[2pt]
Catchable ep.\ (all tasks) & \multicolumn{5}{l}{Qwen3.6-35B 0.035 (7/200) \quad Qwen3.5-9B 0.030 (3/100) \quad \texttt{gpt-4.1-mini} 0.240 (48/200)} \\
Qwen3.5-9B ungated arm & \multicolumn{5}{l}{pass$^1$ 0.710, wrongful 0.230 (23/100); catchable below the pre-set $15\%$ bar, so no gated arm} \\
\bottomrule
\end{tabular}}
\end{table}

The refusal reads as a policy event inside the conversation. In task $38$, the simulated
user asks to cancel a reservation that qualifies under none of the policy's cancellation
branches and confirms when prompted; the agent complies and calls
\texttt{cancel\_reservation}. The gate returns the veto as the tool result, naming the
rule and its reason (cancellation requires booking within $24$ hours, an
airline-cancelled segment, business class, or insurance), and the agent then walks the
user through the four branches and why none applies; the episode passes both the DB and
COMMUNICATE checks.

The difference between the two pairs is the agents' failure mix, not the gate.
Replaying each ungated arm through the same ruleset fires in $24$ of $100$ Qwen3-235B
episodes ($24\%$, every one a DB-failing episode) but in only $7$ of $200$ Qwen3.6-35B
episodes ($3.5\%$): at the smaller model the wrongful mass is argument-level, wrong
flights, wrong payment, or wrong passenger data inside an otherwise eligible call, which
no policy-eligibility gate can decide from database state. The same detection value
shows on the benchmark's shipped \texttt{gpt-4.1-mini} transcripts, whose failure mix
is also eligibility-heavy: offline, the ruleset recalls $0.683$ of extra-write-only
wrongful episodes ($28/41$), $0.714$ counting episodes that also miss writes ($40/56$),
at episode precision $0.917$ ($44/48$) and call precision $0.923$ ($48/52$).

Model size does not predict how often an agent attempts a state-decidable violation,
and the four models compared here cannot separate generation, size, and training. A
smaller model of the same family, Qwen3.5-9B, tool-calls
cleanly (a three-task smoke run passes every check) and almost never makes such attempts
either: in an ungated arm of $50$ tasks $\times$ $2$ trials it reaches pass$^1$ $0.71$
with wrongful writes in $23\%$ of episodes, but replaying its $100$ transcripts through
the ruleset fires in only $3$ ($3\%$), so under a rule we fixed before the run (run the
gated arm only if at least $15\%$ of ungated episodes are catchable) we did not run its
gated arm. The benchmark's own \texttt{gpt-4.1-mini} transcripts make such attempts
often: the ruleset fires on $48$ of their $200$ episodes ($24\%$). The gap is widest on
the refusal subset, the tasks whose gold trajectory contains no write at all, so the
correct outcome is to refuse and change nothing ($24$ of our $50$ tasks, $96$ episodes
per arm; $20$ of $50$ in the shipped task list, $80$ episodes). There
\texttt{gpt-4.1-mini} wrongly writes in $29$ of $80$ episodes ($36.2\%$) and passes only
$0.625$, and the compiled gate catches $21$ of those $29$ offline ($72.4\%$: $16$
ineligible cancellations, $3$ basic-economy modifications, $2$ trip-endpoint changes)
while firing on none of the $51$ clean subset episodes, so it blocks no correct refusal;
our 35B agent is already near the ceiling on the same subset (ungated $6$ of $96$
wrongful, pass$^1$ $0.938$; gated $8$ of $96$, pass$^1$ $0.906$, run-to-run churn rather
than a gate effect). Table~\ref{tab:tau2}, bottom, collects these numbers. The gate's
measured effect tracks the attempt rate across the four agents ($3\%$, $3.5\%$, $24\%$,
and $24\%$ of episodes catchable for Qwen3.5-9B, Qwen3.6-35B, \texttt{gpt-4.1-mini},
and Qwen3-235B): the effect size is set by how often the agent attempts state-decidable
violations, and the class is eliminated regardless. Compiled from an external benchmark's
policy document and wrapped around its tool executor, the gate eliminates the
state-decidable violation class at zero gold overblocks in every arm, and raises strict
success where that class is common; what it cannot catch, argument-level and
intent-level errors, is the same boundary Section~\ref{sec:results:rq3} establishes for
the payload testbed. The other two text domains of $\tau^2$-bench do not carry the class
at all: in the benchmark's shipped retail transcripts every model's extra writes are
wrong arguments or writes on orders the task never touches, and no model repeats a
once-only operation on the same order, because the retail tools refuse such calls
themselves; in telecom, Qwen3-235B writes outside the gold set in $54\%$ of $114$
episodes, but every such write is state-legal and unrequested (enabling roaming in $49$
of them), so a state-bound gate has nothing to refuse there and we did not run one.

The one refinement audit above is itself isolable. Replaying the shipped ruleset under
the pre-audit rule semantics overblocks $4/55$ gold write actions on the
shipped-transcript task revision (the standing task-$7$ incident plus tasks $9$/$37$,
flown-segment cancellation, and task $29$, trip endpoints) and $0/49$ on the task set
of our paired runs under either semantics: the audit's net effect is the removal of
three gold overblocks on one task set and of none on the task set of the paired arms
in Table~\ref{tab:tau2}. A fresh
recompile from the same policy document under the pre-audit vocabulary (one LLM call)
overblocks $14/55$ and $10/49$; the excess is compile variance (a stricter
basic-economy rule family), not the audit: on the unit battery the hand-written
pre-audit reference scores $31/33$, failing exactly the two audited cases (a
flown-segment business cancellation and a same-metro endpoint swap), while the fresh
compile scores $28/33$, failing the same two, two allow cases of its stricter
basic-economy family, and an already-cancelled deny. The
post-audit baselines are $1/55$ (task $7$) and $0/49$; compile-to-compile variance
exceeds the audit's effect, consistent with the compile-accuracy bounds of
Section~\ref{sec:results:rq3}
(\texttt{killtest/scratch\_tau2/artifacts/t2\_isolate\_replay\_*},
\texttt{t2\_preaudit\_*}). Deployed without gold trajectories, a compiled gate therefore
needs its own safeguards: the unit battery and gold replay used here, a compile validator
of the kind in Appendix~\ref{app:residuals:tables}, and, for rules that fail them, the
advisory rung, which surfaces the rule without refusing.

\section{External-Benchmark Case Study: PM-Bench}
\label{app:pmbench}
The $\tau^2$-bench study injects the gate where a benchmark already ships a state-bound
failure class. PM-Bench \citep{pmbench} is the opposite case, and we run the whole ladder
on it to see what owned state does where the failures are mostly not state violations. The
benchmark places one agent in a seven-day simulated week ($80$ steps, $83$ intentions,
$81$ scored after two cancellations): each step is a short vignette with three
ongoing-activity options and a menu of action handles (the day's open tasks plus three
distractors), and the agent must select the handle of every intention that is due now,
where due means that its cue appears in the vignette ($42$ intentions), on a hidden
state channel the agent has to query ($15$), or on the clock, which is also hidden
($26$, exact step time or up to $60$ minutes late). Intentions are assigned in the
morning plan, seven of them as a note on an earlier day for a later one, and $11$
mid-day messages reschedule ($6$), replace the cue of ($3$) or cancel ($2$) an
intention. The
menu hides a task once it is completed, so a re-execution cannot happen and the
benchmark's commission and dependency-violation counters are zero for every released
single-agent run; the state-bound failures it does expose are the $11$ updates, and the
scored failures are misses (the cue passed unnoticed), false alarms
(a selection when nothing or something else was due), and update violations (acting on a
cancelled or rescheduled intention at its old condition). We use the released
deterministic v9 week and the benchmark's own scorer unchanged.

\paragraph{The record.} The owned state here is an intention record: for each
intention, its text, its trigger (a cue phrase, a clock time, or a state channel), the
day it belongs to, and a status in \{pending, done, cancelled\}. Nothing in it is read
from the benchmark's task files. It is compiled from the same text the agent sees by a
separate call to the agent model on the daily header, on each morning plan, and on the
$32$ of $80$ vignettes whose wording can change the record (a note for later, a
correction, a cancellation, a reschedule; a hand-written keyword list, written with this
week's update vocabulary in view, selects them and misses none of the $11$ updates or the
seven later-day notes). The compile call returns add,
cancel, retime and recue operations; a status becomes done only on a receipt, which in
this environment is a handle the agent executed that the next menu no longer lists. As
in Appendix~\ref{app:compile}, compile errors flow into the record uncorrected.

\paragraph{The rungs.} The raw rung is the benchmark's \texttt{single\_baseline}, and
our loop reproduces its prompts to the byte (checked against the benchmark's runner over
all $80$ steps with a stub model). The text checklist renders the record under every
step prompt: pending intentions with their triggers, cancelled ones marked as such,
later-day notes. The directive gate adds the matcher: before the agent acts, a call to
the same model receives the pending intentions, the vignette, and the menu, and names
the handles whose trigger is visible now; the module writes them as \textsc{due now}
lines, keeps an intention announced at the previous step but not executed as
\textsc{still due}, and, when time- or channel-triggered intentions are pending with no
clock reading or channel response this step, says so. After every state query the
matcher runs again on the response and any newly triggered handle is announced. The
enforcement gate binds the gate to the matcher's judgement: it keeps the directives and
refuses, at selection time, any handle that
the matcher has not resolved as triggered from the step text and the responses the
agent obtained; the refusal is reported at the next step. This gate demands
evidence: a time intention selected without a clock reading is refused even when the
guess was right, and the agent can read the clock and select it again inside the late
window. A further arm restores the state-bound policy of
Section~\ref{sec:systems:ladder}: it makes no matcher call for the gate decision and
refuses a selection only when every record intention it resolves to is already
terminal (done or cancelled); a selection that resolves to a pending intention, or to
none, passes. The directive layer is unchanged, and the only matcher influence left in
the decision is the inherited handle-to-intention resolution, so a mis-resolved handle
can still misdirect a state refusal. The benchmark's own \texttt{todo\_ledger} scaffold, in which the agent rewrites
a five-item ledger inside its reply, is the self-written ledger of Section~\ref{sec:results:ladder} and runs unchanged.

\begin{table}[t]
\centering\small
\caption{PM-Bench (synthetic week v9, $80$ steps, $81$ scored intentions) under the state-coupling ladder. Set-F1 is the benchmark's primary metric over per-step chosen-versus-due task sets; a late selection counts as a false positive in Set-F1 but as late rather than missed at the intention level. Hit and miss are shares of the $81$ intentions; FA counts false alarms (selections when nothing or something else was due) and UV selections that violate a cancellation or reschedule; queries are clock and channel reads per run. The self-written ledger is the benchmark's own \texttt{todo\_ledger} scaffold. $T{=}0$: one run at temperature $0$; the remaining columns are mean $\pm$ SD over four runs at temperature $0.7$ of the same released week (repeats of one scenario, not independent tasks). Matcher P/R: precision and recall of the matcher's announcements against the benchmark's per-step ground truth; refused/due: gate refusals per run and how many of them refused a due selection.}
\label{tab:pmbench}
\resizebox{\textwidth}{!}{\begin{tabular}{llrrrrrrrrr}
\toprule
Agent & Rung & Set-F1 $T{=}0$ & Set-F1 $T{=}0.7$ & Hit\% & Miss\% & FA & UV & Queries & Matcher P/R & Refused/due \\
\midrule
Qwen3-235B & raw transcript & 0.67 & 0.67$\pm$0.03 & 56$\pm$2 & 42$\pm$2 & 8$\pm$3 & 4$\pm$2 & 28$\pm$2 & -- & -- \\
 & self-written ledger & 0.73 & 0.71$\pm$0.04 & 60$\pm$5 & 37$\pm$4 & 4$\pm$1 & 2$\pm$1 & 21$\pm$9 & -- & -- \\
 & text checklist & 0.78 & 0.77$\pm$0.01 & 64$\pm$2 & 34$\pm$3 & 1$\pm$1 & 0$\pm$0 & 29$\pm$8 & -- & -- \\
 & directive gate & 0.70 & 0.71$\pm$0.03 & 70$\pm$4 & 25$\pm$4 & 16$\pm$3 & 5$\pm$1 & 62$\pm$23 & 0.72/0.68 & -- \\
 & enforcement gate (judgement) & 0.74 & 0.70$\pm$0.02 & 67$\pm$4 & 28$\pm$7 & 14$\pm$4 & 4$\pm$2 & 46$\pm$20 & 0.73/0.65 & 1.8/0.8 \\
 & enforcement gate (state) & 0.74 & 0.72$\pm$0.01 & 65$\pm$1 & 30$\pm$1 & 9$\pm$4 & 3$\pm$1 & 65$\pm$25 & 0.76/0.68 & 7.8/3.8 \\
\midrule
Qwen3.6-35B & raw transcript & 0.65 & 0.63$\pm$0.06 & 54$\pm$6 & 43$\pm$7 & 11$\pm$11 & 4$\pm$3 & 21$\pm$12 & -- & -- \\
 & self-written ledger & 0.67 & 0.66$\pm$0.02 & 54$\pm$3 & 44$\pm$3 & 7$\pm$2 & 2$\pm$2 & 17$\pm$3 & -- & -- \\
 & text checklist & 0.69 & 0.72$\pm$0.04 & 60$\pm$5 & 38$\pm$4 & 4$\pm$2 & 3$\pm$2 & 25$\pm$13 & -- & -- \\
 & directive gate & 0.71 & 0.72$\pm$0.02 & 62$\pm$5 & 36$\pm$5 & 8$\pm$5 & 4$\pm$2 & 35$\pm$17 & 0.84/0.39 & -- \\
 & enforcement gate (judgement) & 0.52 & 0.55$\pm$0.01 & 42$\pm$1 & 56$\pm$2 & 8$\pm$2 & 3$\pm$2 & 35$\pm$13 & 0.79/0.37 & 21.5/16.8 \\
 & enforcement gate (state) & 0.69 & 0.66$\pm$0.05 & 54$\pm$6 & 44$\pm$6 & 8$\pm$5 & 4$\pm$3 & 30$\pm$19 & 0.78/0.36 & 6.5/4.0 \\
\bottomrule
\end{tabular}}
\end{table}

\paragraph{Results.} Table~\ref{tab:pmbench} reports one run at temperature $0$ and
four at $0.7$ per cell; the prose below quotes the four $0.7$ runs (mean, with SD in
the table), and every run repeats the one released week, so the spread measures
sampling variance, not variation across scenarios. The benchmark's loop caps neither
the state queries in a step nor the prompt, so the servers ran with $64$k (Qwen3-235B)
and $128$k (Qwen3.6-35B) context windows, the latter because in two runs the
Qwen3.6-35B agent queried a channel several hundred times in one step. For scale, the
benchmark's released single runs span Set-F1 $0.42$--$0.79$ over eight models, with
GPT-5.4 at $0.73$ alone and $0.79$ with its best scaffold. Three findings follow, each
different from the payload testbed.

Showing the record is the best rung here. At Qwen3-235B the text checklist lifts
Set-F1 from $0.67$ to $0.77$ (SD $0.01$; $0.67 \to 0.78$ at temperature $0$), removes
the false alarms ($7.8 \to 1.0$ per run) and the update violations ($4.0 \to 0.2$),
and cuts the event-cued misses from $8.2$ to $2.5$ per run, while the self-written
ledger reaches $0.71$; at Qwen3.6-35B the checklist reaches $0.72$ against $0.63$ raw
and $0.66$ for the ledger, and ties the directive gate ($0.72$). The record carries
what the transcript buries, the condition each intention waits for, and an agent that
sees the conditions stops acting when none holds. What the checklist does not do is
make the agent look: time-cued misses ($11.8$ per run) and channel-cued misses ($15$ of
$15$) are unchanged, because the clock and the channels are hidden and only a query
reveals them.

Telling trades misses for false alarms. The directive gate's nudges double the
queries at Qwen3-235B ($28 \to 62$ per run, most of the increase on the clock) and cut
the misses to $25\%$, the lowest of any rung, but the agent executes the matcher's
announcements as given: of $20.8$ false selections per run, $17.5$ were announced, and
the matcher's precision is $0.72$ (recall $0.89$ on event cues, $0.67$ on times, $0.10$
on channel cues, which it sees only after the agent queries). Set-F1 ends at $0.71$,
below the checklist. This is the obedience of Section~\ref{sec:results:rq3} with the
sign reversed: a directive that is right $72\%$ of the time is followed every time.

Enforcing the matcher's judgement hurts; enforcing state does not. At Qwen3-235B the
judgement-bound gate refuses
$1.8$ selections per run ($0.8$ of them due), because under directives the agent
selects almost nothing the matcher did not announce; of its $18.5$ false selections
per run, $16.0$ were announced, and a gate bound to the matcher's evidence admits them
by construction, so Set-F1 stays at $0.70$. At Qwen3.6-35B, whose matcher recalls
$0.37$ of the due intentions, the same gate refuses $21.5$ selections per run, $16.8$ of
them due, and Set-F1 falls to $0.55$, below the raw transcript. The state-bound gate,
with the due judgement out of the decision, refuses $7.8$ and $6.5$ selections per run
and returns to the self-written ledger's level at both agents, $0.72$ (SD $0.01$;
$0.74$ at temperature $0$) against the ledger's $0.71$, and $0.66$ (SD $0.05$; $0.69$)
against $0.66$, no longer below the raw transcript; its residual refusals of due
selections ($3.8$ and $4.0$ per run) come from the handle-to-intention resolution, a
handle mis-resolved to a done or cancelled intention, not from any due judgement. No
form of enforcement beats showing the record here: the judgement-bound rung is
bounded by its matcher exactly as in Appendix~\ref{app:a3}; on a benchmark where
due-ness is a judgement about free text rather than a fact in the state, that bound is
the whole result of enforcing the judgement, a state-bound gate has almost nothing
left to decide, and the rung that wins is the one that asks the least of the model's
obedience and nothing of a matcher.

\section{The One-Node Case}
\label{app:commitment}
A one-node precursor of this study measured a single standing commitment (``use Willow
for my next booking, just this once'') compiled into a guarded trigger that fires on a
later, differently phrased request and then sanitizes the transcript trace of the
consumed one-shot. On a testbed of $100$ concurrent commitments with a Qwen3.6-35B agent,
re-execution of consumed one-shots fell from $0.87$ under a prompt manifest to $0$ at
$0.88$ trigger recall, and removing transcript sanitization alone returned it to $0.64$:
the fresh-order hazard that the graph testbed generalizes from one node to a graph.

\end{document}